\documentclass{article}

\usepackage{genai4health_2026} 

\usepackage[utf8]{inputenc}
\usepackage[T1]{fontenc}

\usepackage{amsfonts}
\usepackage{amsmath}
\AddToHook{cmd/normalsize/after}{\setlength{\abovedisplayskip}{2pt plus 1pt minus 1pt}\setlength{\belowdisplayskip}{2pt plus 1pt minus 1pt}\setlength{\abovedisplayshortskip}{0pt plus 1pt}\setlength{\belowdisplayshortskip}{2pt plus 1pt minus 1pt}}
\usepackage{amssymb}
\usepackage{booktabs}
\usepackage{longtable}
\usepackage{tabularx}
\usepackage{colortbl}
\usepackage{float}
\floatstyle{ruled}
\newfloat{algorithm}{tbp}{loa}[section]
\floatname{algorithm}{Algorithm}
\usepackage{graphicx}
\usepackage{tikz}
\usetikzlibrary{arrows.meta}
\usepackage{microtype}
\usepackage{nicefrac}
\usepackage{url}
\usepackage{xcolor}
\usepackage[colorlinks=true, linkcolor=violet, citecolor=blue!50!white]{hyperref}

\title{Synthesis of Ovarian Cancer Multi-Omics Datasets: A Biology-Informed Evaluation Framework}

\author{Déborah Boyenval\textsuperscript{1}\thanks{Corresponding author: \texttt{deborah.boyenval@helsinki.fi}},
  Alina Malyutina\textsuperscript{2},\\
  \textbf{Jaana Oikkonen\textsuperscript{1}},
  \textbf{Sammeli Liikkanen\textsuperscript{3}}, 
  \textbf{Sampsa Hautaniemi\textsuperscript{1}}\\
  \textsuperscript{1}Research Program in Systems Oncology, Research Programs Unit,\\
  Faculty of Medicine, University of Helsinki, Helsinki, Finland\\
  \textsuperscript{2}Orion Pharma, Espoo, Finland\\
  \textsuperscript{3}PalsaIQ, Finland\\
}

\begin{document}

\maketitle

\begin{abstract}
Paired multi-omic cancer cohorts are costly to assemble and difficult to share, motivating synthetic cohorts as a complementary resource.
As generative model development accelerates, we shift the focus from proposing yet another architecture to defining evaluation criteria that determine whether synthetic data preserve biologically relevant structure.
We introduce a biology-informed and patient-disjoint framework for evaluating the generation of paired RNA expression and copy-number variation (CNV) profiles, and apply it to the longitudinal DECIDER high-grade serous ovarian cancer cohort.
The framework assesses modality-specific fidelity and coherence between RNA and CNV pairs using agreement between real training and held-out data as a reference, 
alongside complementary training-data proximity diagnostics.
We adapt three multi-omic integration architectures based on variational autoencoders to generate paired profiles.
The model leading several fidelity and coherence measures also shows the strongest training-data proximity.
The protocol and case study offer the community a reference for comparing future multi-omic generators. 

\end{abstract}

\section{Introduction}
\label{sec:introduction}
Analyzing molecular data from heterogeneous diseases, such as ovarian cancer, requires cohorts that capture variation across patients and biological contexts \citep{tcga2011ovarian}.
Because these data are sensitive and difficult to pool across institutions, synthetic cohorts could provide a useful resource for model development and evaluation, but only if they preserve biologically relevant structure without simply reproducing training samples \citep{qian2023synthcity,chen2024biologically}.
In sequencing studies, a multimodal molecular cohort contains complementary omic measurements obtained from the same biological samples.
Such cohorts are costly to assemble and often small relative to the number of molecular variables they contain.

Studies of the DECIDER high-grade serous carcinoma (HGSC) cohort illustrate why cross-modal relationships matter for questions of treatment response and resistance \citep{deciderNCT04846933}.
Copy-number variations (CNV) are gains or losses of DNA segments that can alter gene expression in cancer cells.
One study combined CNV profiles with tumor-cell expression estimated from bulk RNA sequencing to quantify these transcriptional effects and identify pathways associated with poor treatment response \citep{jamalzadeh2024cnaimpact}.
Another integrated whole-genome sequencing with bulk RNA sequencing to distinguish functionally credible driver alterations from non-functional events and identify biomarkers that could guide targeted therapy selection \citep{muranen2026multimodal}.
A third combined somatic mutations, CNV, paired bulk RNA sequencing, and histopathology to link tumor evolutionary states to biological pathways, morphology, and treatment response \citep{lahtinen2023evolutionary}.
In this paper, we simplify the multimodal generation problem by considering only two modalities: CNV and RNA.

From a machine-learning perspective, the central problem is to synthesize new paired molecular profiles from such cohorts.
This task is difficult because each modality has a distinct distribution and biological structure, while a valid synthetic sample must also retain the association between modalities.
For example, a generator may reproduce the separate distributions of RNA and CNV yet fail to associate the two measurements within a synthetic sample.
Developing multimodal generators is insufficient without an evaluation framework tailored to multimodal omics.
Statistical fidelity, biological coherence, and training-data proximity must be assessed as separate dimensions \citep{qian2023synthcity,chen2024biologically}.
The framework must also distinguish properties within each modality from coherence across modalities.
We address this need by benchmarking de novo generation of paired RNA and CNV profiles.
Our evaluation examines modality-specific structure, within-sample RNA-CNV associations, and how closely synthetic profiles resemble the training data.

\textbf{Related work.}
Synthcity provides a common framework for generating and benchmarking synthetic tabular data across use cases \citep{qian2023synthcity}.
Its plugins include conditional generative adversarial networks, variational autoencoders, normalizing flows, and diffusion models.
Its fidelity metrics include Jensen-Shannon and Wasserstein distances and real-versus-synthetic detection.
It combines fidelity measures with downstream and empirical privacy evaluations.
This breadth supports consistent comparisons, but each dataset is represented through a single tabular interface.
Synthcity does not test assay-specific biological properties or relationships between separately structured molecular modalities such as the transcriptome, genome, or methylome.

A related line of work addresses multimodal generation by reconstructing a missing modality for an observed sample.
Marchesi et al. use latent-space diffusion models to reconstruct a CNV, RNA, protein, or histopathology representation from the modalities observed for the same tumor in The Cancer Genome Atlas (TCGA) \citep{marchesi2026coherent}.
At single-cell resolution, scDiffusion-X uses a diffusion model to generate paired RNA and chromatin-accessibility profiles or infer either modality from the other \citep{luo2026scdiffusion}.
Multi-Omic Synthetic Augmentation (MOSA) uses a variational autoencoder to complete missing genomic, transcriptomic, proteomic, and drug-response measurements for cancer cell lines \citep{cai2024mosa}.
These settings differ, however, from de novo generation of complete paired bulk RNA and CNV profiles.

Evaluation studies have also moved beyond generic statistical fidelity.
The Critical Assessment of Massive Data Analysis (CAMDA) Health Privacy Challenge evaluates synthetic bulk RNA-sequencing data through distributional fidelity, downstream utility, biological plausibility, and empirical privacy risk \citep{ozturk2026syntheticomics}.
Its biologically motivated tasks test the recovery of condition-specific differential expression and gene-gene co-expression networks.
Its utility analysis asks whether a classifier trained on synthetic TCGA breast-cancer RNA profiles can predict molecular subtypes in held-out real samples.
However, these evaluations remain unimodal.
Marchesi et al. evaluate each reconstructed modality separately by testing whether it retains signals for tumor-type and stage prediction \citep{marchesi2026coherent}.
This is again a modality-specific evaluation and does not assess whether newly generated RNA and CNV profiles preserve their within-sample association.

\textbf{Contributions.}
We study how to evaluate de novo generators of paired RNA and CNV profiles when realistic marginal distributions do not necessarily imply biologically coherent pairs (Figure~\ref{fig:intro_gist}).

\begingroup
\setlength{\leftmargini}{1.2em}
\begin{itemize}
    \item \textbf{Evaluation protocol.}
    We propose an evaluation protocol that uses different patients for training and testing.
    It examines how well synthetic profiles reproduce statistical and biological properties within each modality, preserve relationships across modalities, and avoid excessive similarity to the training data.
    Its RNA-CNV instantiation provides a concrete starting point for broader evaluation frameworks for synthetic multi-omic cohorts.

    \item \textbf{Generative benchmark.}
    We adapt ISOMERIC \citep{selega2025isomeric}, TMO-Net \citep{wang2024tmonet}, and CustOmics \citep{benkirane2023customics}, three multi-omic architectures based on variational autoencoders with distinct integration mechanisms, to a common de novo generation task. 
    Their latent-variable formulations support generative decoding, we define explicit prior-sampling paths that produce complete paired RNA-CNV profiles without encoding an observed sample.
    \item \textbf{Empirical comparison.}
    We evaluate the adapted models on patient-disjoint DECIDER cohorts, highlighting differences in modality-specific fidelity, preservation of RNA-CNV relationships, and proximity to training data.

\end{itemize}
\endgroup

\begingroup
\setlength{\intextsep}{4pt}
\begin{figure}[!htbp]
    \centering
    \setlength{\abovecaptionskip}{3pt}
    \includegraphics[width=\linewidth]{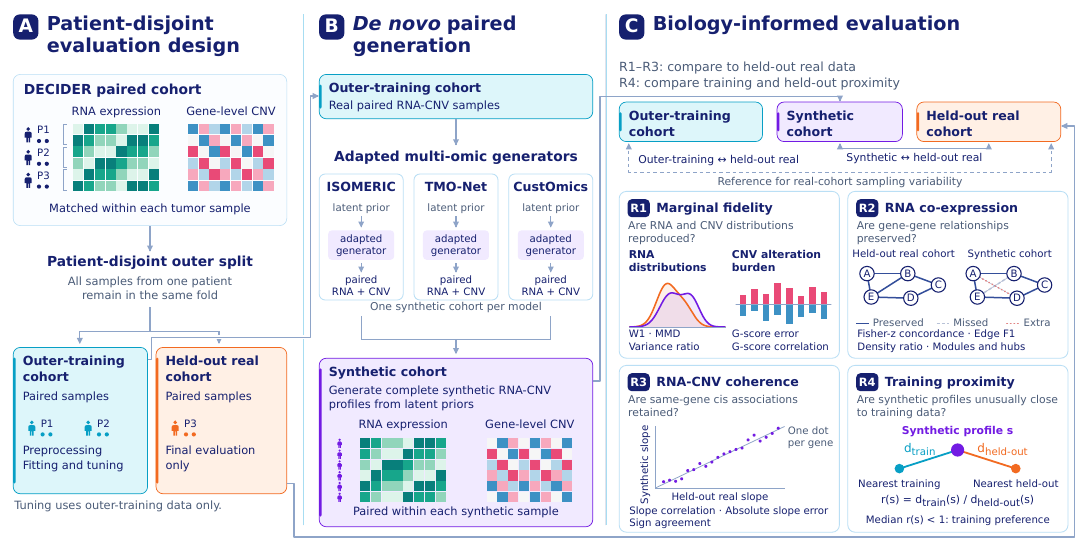}
    \caption{\textbf{Patient-disjoint generation and evaluation of paired RNA--CNV profiles.}
    \textbf{(A)} All samples from a patient remain in one outer fold, preprocessing,
    fitting, and tuning use training data only. Treatment cohorts are evaluated separately.
    \textbf{(B)} Each adapted model generates its own cohort of complete pairs from latent priors.
    \textbf{(C)} R1--R3 compare synthetic and held-out real data, with real
    training--held-out agreement as a reference. R1 assesses marginal fidelity. R2 compares RNA
    co-expression networks, links denote preserved, missed, or extra edges.
    R3 compares RNA-CNV slopes.
    R4 compares nearest-neighbor distances to equally sized training and held-out
    reference sets. A median distance ratio below one indicates training preference,
    without establishing copying or privacy. All illustrations are schematic.}
    \label{fig:intro_gist}
\end{figure}
\endgroup

\section{Background on multi-omic generative models}
\label{sec:models}

Generative approaches for multi-omic data span several model families.
We focus on variational autoencoder (VAE) architectures because they define
distributions over latent variables and specify how these variables are
decoded into each modality. This provides a common framework for comparing
how RNA and CNV are coupled during generation.
We first recall some preliminary concepts \citep{kingma2013autoencoding}, then
describe the three adaptations evaluated for paired RNA-CNV generation.

\subsection{Variational autoencoders}

\textbf{Model.}
A VAE models the marginal distribution of observations $p_\theta(x)$ using an
unobserved latent representation, or latent code, $z$:
\[
p_\theta(x)=\int p_\theta(x\mid z)p(z)\,\mathrm{d}z,
\qquad p(z)=\mathcal{N}(0,I).
\]
The decoder maps $z$ to the parameters of $p_\theta(x\mid z)$, the conditional
distribution of observations given the latent representation.
Learning from an observation $x$ requires inferring the latent values that
could have generated it. Since the exact posterior $p_\theta(z\mid x)$ is
generally intractable, an encoder maps $x$ to the parameters of a tractable
approximation $q_\phi(z\mid x)$.
A common choice, used in our adaptations, is a Gaussian with diagonal covariance:
\[
q_\phi(z\mid x)
=\mathcal{N}\!\left(\mu_\phi(x),
\operatorname{diag}\!\left(\sigma_\phi^2(x)\right)\right).
\]
\textbf{Training.}
Latent vectors are sampled from the encoder's approximate posteriors and
decoded to reconstruct the inputs. Both networks are updated jointly to
minimize the negative evidence lower bound (ELBO), combining reconstruction
loss with Kullback--Leibler (KL) regularization of the posteriors toward the
prior. Reparameterization allows gradients through sampling.
The objective is given in Appendix~\ref{app:model-training}.

\textbf{Sampling.}
Once the model has been trained, a new observation can be generated by drawing
$z\sim p(z)$, followed by $x\sim p_\theta(x\mid z)$.
This procedure does not require encoding an observed profile.
Figure~\ref{fig:generation-dags}(a) illustrates this sampling path.

\subsection{VAE-based models for paired RNA-CNV generation}

We consider paired observations $x=(r,c)$, where
$r\in\mathbb{R}^{d_R}$ and $c\in\mathbb{R}^{d_C}$ denote RNA and CNV profiles
from the same sample, after applying the scaling transformation used by each model.
A hat denotes a reconstruction and a tilde denotes a generated profile.
The RNA and CNV decoders, denoted by $D$, output conditional means.
For all three adaptations, we generate profiles by sampling latent variables
and using these decoder means as synthetic profiles, without adding
observation noise. The generated profiles are then mapped back to the
input scales.

All models use standard Gaussian priors during training. For sampling, we write
\[
S_T(\mu,\sigma)=\mu+T\,\sigma\odot\epsilon,
\qquad \epsilon\sim\mathcal{N}(0,I),
\]
where $T>0$ scales the latent standard deviation and $T=1$ recovers the
specified Gaussian. Thus $S_T(0,1)$ draws from $\mathcal{N}(0,T^2I)$.

\textbf{ISOMERIC \citep{selega2025isomeric}.}
The main idea of the model is the separation of RNA variation into two latent components: $w$, which
contributes to both RNA and CNV reconstruction, and $z$, which contributes to
RNA reconstruction only. 
A single RNA encoder predicts the means and variances of the Gaussian approximate posteriors for $z$ and $w$.  
We retain this factorization:
\begin{equation}
\begin{aligned}
p_\theta(r,c,z,w)
&=p(z)\,p(w)\,p_{\theta_C}(c\mid w)\,p_{\theta_R}(r\mid z,w), \\
q_\phi(z,w\mid r)
&=q_{\phi_z}(z\mid r)\,q_{\phi_w}(w\mid r), \\
\widehat r&=D_R(z,w),\qquad \widehat c=D_C(w).
\end{aligned}
\label{eq:isomeric-factorization}
\end{equation}
The original model was developed for single-cell expression and regional
copy-number profiles. We adapt it to tumor samples, using expression inferred
for tumor cells from bulk RNA sequencing and continuous CNV values for each gene.
Training combines RNA and CNV reconstruction, two KL penalties, and a linear
adversarial CNV predictor $A_C$. The adversarial objective discourages CNV
predictability from $z$ but does not guarantee independence
(Appendix~\ref{app:model-training}). The encoder takes only RNA as input to
parameterize both approximate posteriors. During training, paired CNV profiles
provide targets for the CNV decoder and the adversarial predictor.
For prior-based generation,
\[
z=S_{T_z}(0,1),\qquad w=S_{T_w}(0,1),\qquad
\widetilde r=D_R(z,w),\qquad \widetilde c=D_C(w).
\]

\textbf{TMO-Net \citep{wang2024tmonet}.}
TMO-Net learns to predict missing modalities from observed ones.
We retain RNA and CNV, with one VAE reconstructing each modality from its
own input. A third encoder takes CNV as input and predicts the parameters
of a Gaussian latent distribution. Latent samples are decoded into RNA
profiles using the same decoder as the RNA VAE.
During training, we compare the RNA predicted from CNV with the RNA
observed in the same sample, encouraging the model to learn associations
between copy number and expression.
We retain only the CNV to RNA direction, motivated by the transcriptional
effects of copy number \citep{jamalzadeh2024cnaimpact}.
With CNV as the sole conditioning modality, fusion across source modalities
is unnecessary. The original model also includes clinical prediction heads,
which are outside the scope of our generation task.
The original contrastive objective separates cancer types and is omitted
for this HGSC cohort. We omit the adversarial objective as a training
simplification whose impact is not evaluated separately.
This gives three paths:
\begin{equation}
\begin{aligned}
z_C\sim q_{\phi_C}(z_C\mid c),
&\qquad \widehat c=D_C(z_C), \\
z_R\sim q_{\phi_R}(z_R\mid r),
&\qquad \widehat r_{\mathrm{self}}=D_R(z_R), \\
z_{C\rightarrow R}\sim q_{\phi_{C\rightarrow R}}(z_{C\rightarrow R}\mid c),
&\qquad \widehat r_{C\rightarrow R}=D_R(z_{C\rightarrow R}).
\end{aligned}
\label{eq:tmonet-paths}
\end{equation}
Training combines reconstruction along these paths with KL regularization
toward standard Gaussian priors. An additional KL term encourages agreement
between the RNA latent distributions inferred from CNV and RNA for each
paired sample (Appendix~\ref{app:model-training}).
For generation, we first sample a CNV profile and then generate RNA
conditionally on it, allowing dependencies within each synthetic pair:
\[
\begin{aligned}
z_C&=S_{T_C}(0,1), &
\widetilde c&=D_C(z_C), \\
z_{C\rightarrow R}
&=S_{T_{C\rightarrow R}}\!\left(
\mu_{C\rightarrow R}(\widetilde c),
\sigma_{C\rightarrow R}(\widetilde c)
\right), &
\widetilde r&=D_R(z_{C\rightarrow R}).
\end{aligned}
\]
We sample the CNV latent from its prior, then the RNA latent given the generated CNV.

\textbf{CustOmics \citep{benkirane2023customics}.}
CustOmics first learns a compact representation of each modality, then
integrates them in a central VAE.
The authors propose this strategy to accommodate different learning dynamics
and limit domination by sources with stronger signals.
We adapt this hierarchy to RNA and CNV generation, removing classification
and survival heads.
RNA and CNV autoencoders learn compact vectors $h_R$ and $h_C$ by
reconstructing their respective inputs, without latent sampling.
We concatenate these vectors for each paired sample into $h$.
The central encoder uses $h$ to compute the means and variances of a
Gaussian distribution over $z$, decoder $G$ reconstructs $h$ from a
sampled latent vector:
\begin{equation}
\setlength{\jot}{1pt}
\begin{aligned}
h_R&=E_R(r),\quad \widehat r_{\mathrm{self}}=D_R(h_R),\quad
h_C=E_C(c),\quad \widehat c_{\mathrm{self}}=D_C(h_C), \\
h&=[h_R,h_C],\quad z\sim q_\phi(z\mid h),\quad \widehat h=G(z).
\end{aligned}
\label{eq:customics-architecture}
\end{equation}
Training first fits the modality autoencoders, then jointly adjusts all
modules with modality reconstruction, reconstruction of $h$, and maximum
mean discrepancy (MMD) regularization for prior matching.
There is no direct reconstruction loss on the RNA and CNV profiles obtained
through $G$ and the modality decoders, the full objective is given in
Appendix~\ref{app:model-training}.
For generation, $G$ maps one central latent sample to two representations,
which the modality decoders turn into paired RNA and CNV profiles:
\begingroup
\setlength{\abovedisplayskip}{0pt}
\setlength{\belowdisplayskip}{0pt}
\setlength{\abovedisplayshortskip}{0pt}
\setlength{\belowdisplayshortskip}{0pt}
\[
z=S_T(0,1),\qquad
(\widetilde h_R,\widetilde h_C)=G(z),\qquad
\widetilde r=D_R(\widetilde h_R),\qquad
\widetilde c=D_C(\widetilde h_C).
\]
\endgroup

\textbf{Comparison scope.}
ISOMERIC uses one latent
shared by RNA and CNV together with an RNA-specific latent. TMO-Net generates
CNV first and then RNA conditionally on that CNV. CustOmics branches one
central latent into modality-specific representations. All three produce
complete RNA-CNV pairs without encoding an observed profile.
Appendix~\ref{app:generation-dags} shows the corresponding sampling graphs.
Appendix Figure~\ref{fig:training-architectures} shows the computations and
objectives used to train each adaptation.

\section{Real-data cohort and evaluation framework}
\label{sec:data-evaluation}

\subsection{Cohort and evaluation protocol}
\label{sec:cohort-protocol}
\textbf{DECIDER cohort.}
We use data from DECIDER, a prospective longitudinal observational study of high-grade serous carcinoma (HGSC, NCT04846933) \citep{deciderNCT04846933}. 
Most patients received either primary debulking surgery (PDS) or neoadjuvant chemotherapy (NACT) as their initial treatment strategy. 
Tumor samples were collected from adnexal or tubo-ovarian tissue, solid metastatic sites, or fluids such as ascites or pleural fluid. 
The cohort contains 632 samples with paired RNA and CNV profiles from 255 patients: 493 before-treatment samples from 232 patients and 139 after-treatment samples from 89 patients. 

The RNA modality consists of gene-expression profiles inferred from bulk RNA sequencing and apply a $\log_2(x+1)$ transformation.
The gene expression matrix has been decovoluted to extract cancer-cell specific RNA profiles \citep{hakkinen2021prism}.  
The CNV modality consists of continuous gene-level copy-number log-ratios derived from whole-genome sequencing and adjusted for tumor purity and ploidy. 
Each gene is assigned the minimum adjusted value across overlapping segments, with zero denoting the reference state and positive and negative values indicating relative gains and losses \citep{lavikka2024genomespy}. 
CNV features therefore represent segmental alterations at gene resolution,
and multiple genes may reflect the same event.
The RNA and CNV matrices are aligned by sample, so each paired row represents measurements from the same tumor sample.

\textbf{Cross-validation and feature selection.}
We evaluated the before and after-treatment cohorts separately with three patient-disjoint outer folds, balanced across sampling sites, so all samples from a patient remained in the same fold. 
For each split, one fold was held out for evaluation. 
Feature selection and preprocessing used only the outer training data. 
Patient-grouped inner cross-validation selected training hyperparameters and sampling temperatures for all three models, followed by refitting on the full outer training set.
The resulting synthetic cohort was compared separately with the corresponding real training and held-out sets: held-out comparisons assess fidelity to unseen data, whereas training comparisons support the training-proximity diagnostics.

From the 58037 genes for RNA and 36860 genes for CNV, we independently selected the 3000 highest-variance genes (HVG) per modality within each outer training set to reduce the data dimension. 
However, we retained 14 prespecified HGSC biomarkers to keep copy-number events of biological interest \citep{lavikka2024genomespy}, \citep{muranen2026multimodal}, including \textit{CCNE1}, \textit{AKT2}, \textit{PTK2}, \textit{KRAS}, \textit{NF1} and \textit{PTEN}.
Amplification of these oncogenes or inactivation of these tumor suppressors can alter transcriptional programs and suggest potential therapeutic vulnerabilities in HGSC \citep{muranen2026multimodal}.

\subsection{Evaluation metrics}

\paragraph{Marginal CNV fidelity.}
We assess whether synthetic cohorts reproduce how often each gene is amplified or deleted in the real cohort and the magnitude of these alterations.
For each CNV gene $g$, amplification and deletion G-scores summarize these events across samples:
\vspace{-7pt}
\begin{equation}
G^{\mathrm{amp}}_g =
\frac{1}{n}\sum_{i=1}^{n}
\mathbf{1}(x_{ig}>\tau)\min(x_{ig},\kappa),
\qquad
G^{\mathrm{del}}_g =
\frac{1}{n}\sum_{i=1}^{n}
\mathbf{1}(x_{ig}<-\tau)\min(-x_{ig},\kappa).
\label{eq:cnv-gscore}
\end{equation}
Here $x_{ig}$ is the base-2 logarithm of the copy-number ratio relative to the reference level for gene $g$ in sample $i$, after the purity and ploidy adjustment described in Appendix~\ref{app:cnv-fidelity}.
Values of $0$, $+1$, and $-1$ correspond to the reference level, twice that level, and half that level, respectively.
The G-scores average over $n$ samples, use $\tau=0.7$ to select gains and losses, and cap each contribution at $\kappa=2.5$.
Across the 3,000 CNV genes per fold, we compare synthetic and held-out G-scores using Pearson correlations, with real train--held-out correlations as a reference.
Mean absolute G-score differences and overlap among the 50 highest-scoring genes are additionally reported in Appendix Tables~\ref{tab:before-fidelity-full} and~\ref{tab:after-fidelity-full}.

\paragraph{Marginal RNA fidelity.}
We assess whether each gene shows similar expression levels and variability across samples in the synthetic and real cohorts.
We compare gene-level distributions using mean 1-Wasserstein ($W_1$) and Jensen--Shannon (JS) distances, the correlation of gene means, and the median synthetic-to-real variance ratio (target: $1$) \citep{qian2023synthcity,ozturk2026syntheticomics}. Multivariate fidelity is assessed using radial basis function maximum mean discrepancy (RBF MMD$^2$) in PCA space \citep{gretton2012kernel}. An out-of-fold logistic classifier provides a complementary real-versus-synthetic separability score, $D_{\mathrm{disc}}=2|\mathrm{AUC}-0.5|$, ranging from zero (no separation by this classifier) to one (perfect separation) \citep{lopezpaz2016revisiting}.
Details are provided in Appendix~\ref{app:rna-fidelity}.

\paragraph{RNA co-expression preservation.}
We assess whether synthetic RNA preserves how gene expression levels vary together across samples in the real cohort.
Within each outer fold, we select the same number of samples from the real training, held-out real, and synthetic cohorts.
We then use hCoCena v1.1.2 \citep{oestreich2022hcocena} to build one co-expression network per cohort.
For held-out data $H$ and a candidate cohort $C$ (synthetic or real
training), we compare all evaluated Pearson gene-pair correlations:
\vspace{-7pt}
\begin{equation}
C_z =
\operatorname{cor}_{\mathrm{Pearson}}
\left(\{z(r^H_{gh})\}_{g<h},\{z(r^C_{gh})\}_{g<h}\right),
\qquad z(r)=\operatorname{arctanh}(r).
\label{eq:coexpression-concordance}
\end{equation}
Edges require $r\geq0.70$ and a Benjamini--Hochberg-adjusted $p<0.05$,
their recovery is summarized by
$\mathrm{Edge\ F1}=2|E_H\cap E_C|/(|E_H|+|E_C|)$.
We also compare Leiden modules using the adjusted Rand index on genes
assigned to modules in both networks, and the 50 highest-degree hubs
using Jaccard overlap.
To interpret synthetic-held-out agreement, we also apply these metrics to real training and held-out networks, measuring the reproducibility of co-expression structure between real cohorts.
Appendix~\ref{app:coexpression} provides the detailed metric definitions.

\textbf{Cis RNA-CNV association.}
Marginal fidelity does not establish that RNA and CNV remain associated within paired profiles.
Because copy-number alterations can affect transcription at the same locus
\citep{muranen2026multimodal,selega2025isomeric}, we fit the following model
for each gene in the intersection of the independently selected RNA and CNV panels:
\vspace{-7pt}
\begin{equation}
y_{ig}=\alpha_g+\beta_g x_{ig}+\varepsilon_{ig}.
\label{eq:cis-rna-cnv}
\end{equation}
where $y_{ig}$ is the $\log_2(\mathrm{expression}+1)$ RNA value and $x_{ig}$ is the continuous purified CNV log ratio.
We fit this model separately in the real training, held-out, and synthetic cohorts.
The analysis describes preservation within this selected gene subset and
does not estimate genome-wide cis preservation.
Across genes, we compare the slopes $\beta_g$ using mean and median absolute differences, Pearson and Spearman correlations, and an agreement in slope sign.
The real train-test comparison provides a reference for sampling variation between real cohorts. 
Figure~\ref{fig:best-visual-diagnostics}\textbf{(c)} illustrates the slope comparison.

\paragraph{Memorization diagnostics.}
We assess training preference in RNA, CNV, and paired feature spaces~$m$.
For synthetic profiles $\mathcal S$, we compare nearest-neighbor
distances to equally sized training ($\mathcal T$) and held-out
($\mathcal H$) reference sets:
\vspace{-9pt}
\begin{equation}
R_m =
\underset{s\in\mathcal S}{\operatorname{median}}
\frac{d_{\mathcal T}^{m}(s)}
{\max\{d_{\mathcal H}^{m}(s),\epsilon\}}.
\label{eq:training-proximity}
\end{equation}
Distances are Euclidean and $\epsilon$ is machine epsilon, paired
modality scaling and reference subsampling are detailed in
Appendix~\ref{app:memorization}.
We also report membership AUC using negative distance to the nearest
synthetic profile as the score, with training membership as the
positive class \citep{shokri2017membership,vanbreugel2023membership}.
Values $R_m<1$ or AUC $>0.5$ indicate training preference,
AUC~$<0.5$ indicates reversed separation, not stronger privacy.
Neither diagnostic establishes copying or guarantees privacy.

\section{DECIDER case study: evaluating de novo RNA-CNV generation}
\label{sec:generation-experiments}

We use the before-treatment DECIDER cohort described in Section~\ref{sec:cohort-protocol} as the main case study.
It contains 493 paired RNA-CNV samples from 232 patients. 
Across three outer folds, training sets contain 332, 341, and 313 paired samples, with 161, 152, and 180 corresponding held-out samples, patients are disjoint between training and evaluation within each fold.
Within each fold, we select 3000 RNA and 3000 CNV features from the outer training data and fit each generator from Section~\ref{sec:models} with three random initializations.
Training details are provided in Appendix~\ref{app:model-training}.
Each fitted model generates a cohort with the same number of paired profiles as its outer training set.
Unless otherwise specified, results are means $\pm$ sample standard deviations across three outer folds after averaging metrics over three initializations within each fold. These standard deviations summarize between-fold variability. Descriptive model rankings are complemented by paired fold differences in Appendix~\ref{app:paired-comparisons}.

The after-treatment DECIDER cohort is reported in Appendix Figure~\ref{fig:after-treatment-benchmark},
Appendix Figure~\ref{fig:best-visual-diagnostics-after}, and Appendix~\ref{app:synthetic-data-evaluation}. 
This cohort contains 139 paired samples from 89 patients, with 86-96 training samples per outer fold. 

Figure~\ref{fig:before-treatment-benchmark} summarizes the held-out
evaluation. Complete results are provided in
Appendix~\ref{app:synthetic-data-evaluation}.
Figure~\ref{fig:best-visual-diagnostics} illustrates RNA and CNV fidelity, RNA-CNV association, and RNA co-expression, using the model with the best score on the selected metric for each property.
Appendix~\ref{app:model-visual-diagnostics} provides the same diagnostics for each model individually, with after-treatment counterparts in
Appendix~\ref{app:model-visual-diagnostics-after}.
We examine these results through four questions: marginal fidelity (R1), co-expression (R2), cis RNA-CNV association (R3), and training proximity (R4).

\begin{figure}[t]
  \centering
  \includegraphics[width=\linewidth]
  {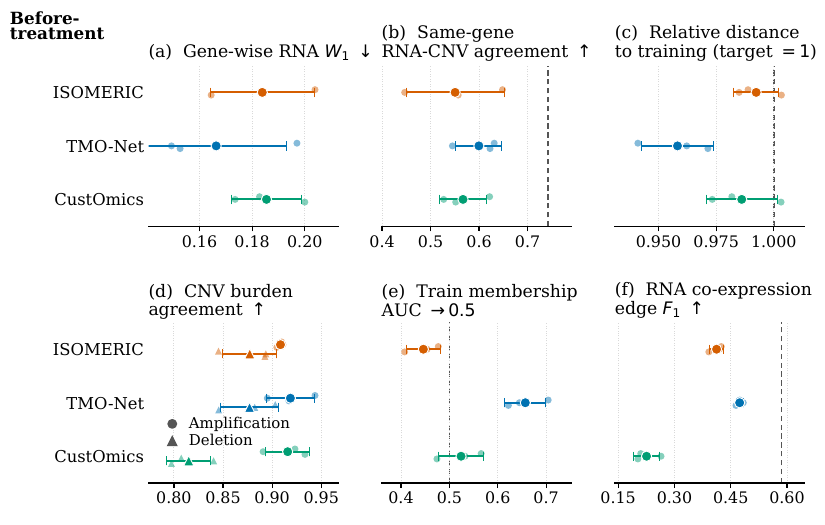}

  \caption{
  \textbf{Held-out evaluation in the before-treatment DECIDER cohort
  (493 paired samples from 232 patients).}
  Panels \textbf{(a--f)} summarize RNA fidelity, same-gene RNA-CNV
  association, paired training proximity, CNV fidelity, paired membership,
  and RNA co-expression. Small points show outer folds, large markers and
  bars show mean $\pm$ sample SD. Dashed lines mark real train--held-out
  references or neutral values.
  Appendix~\ref{app:paired-comparisons} complements these mean comparisons
  with paired model differences within each outer fold, showing whether the
  observed model orderings persist or reverse across folds.
  }
  \label{fig:before-treatment-benchmark}
\end{figure}

\begin{figure}[t]
  \centering
  \includegraphics[width=\linewidth]
  {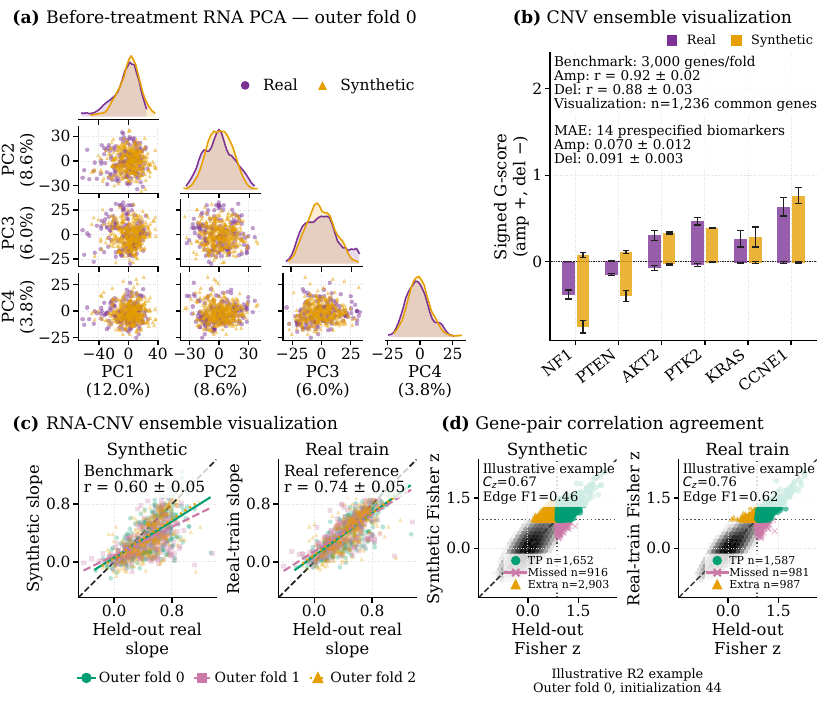}

  \caption{
{  \textbf{Before-treatment DECIDER visual diagnostics.}
  \textbf{(a) ISOMERIC:} RNA PCA, \textbf{(b) TMO-Net:} CNV G-scores,
  \textbf{(c) TMO-Net:} same-gene RNA-CNV slopes,
  \textbf{(d) TMO-Net:} RNA co-expression.
  Model selection and benchmark annotations use metrics computed per
  initialization, averaged within fold, then summarized as mean $\pm$ sample
  SD across three folds. Panels \textbf{(b,c)} display synthetic gene-level
  values averaged across initializations within fold, G-score error bars
  show fold SD. The 1,236 genes are the intersection of the three fold-specific
  3,000-gene CNV panels, used for visualization, benchmark correlations use
  all 3,000 genes per fold. MAE denotes mean absolute error between synthetic
  and held-out G-scores across the 14 prespecified biomarkers, separately for
  amplification and deletion. The MAEs in \textbf{(b)} follow the same metric
  aggregation. Panel \textbf{(d)} shows one illustrative fold and
  initialization, its Fisher-$z$ concordance ($C_z$) and edge F1 describe that example.}
  }
  \label{fig:best-visual-diagnostics}
\end{figure}

\textbf{R1. Are marginal RNA and CNV distributions reproduced?}
We assess RNA fidelity with gene-wise $W_1$, RBF MMD$^2$, and variance
ratios, and CNV fidelity with G-score correlations.
In Figure~\ref{fig:before-treatment-benchmark}\textbf{(a)}, TMO-Net has the
lowest observed mean $W_1$ ($0.166\pm0.027$), compared with $0.184\pm0.020$ for ISOMERIC
and $0.185\pm0.013$ for CustOmics. The corresponding distance between real
training and held-out data is $0.105\pm0.007$.
Mean RBF MMD$^2$ is lowest for ISOMERIC ($0.0125$), followed by TMO-Net
($0.0157$) and CustOmics ($0.0200$). All models underestimate RNA variation:
their median synthetic-to-held-out variance ratios range from $0.57$ to
$0.66$, below the target of $1$
(Appendix Table~\ref{tab:before-fidelity-full}).
The complementary classifier-based diagnostic also gives higher separation
scores for synthetic versus held-out RNA than for real training versus
held-out RNA (Appendix Table~\ref{tab:before-fidelity-full}, methodological
details in Appendix~\ref{app:rna-fidelity}).

For CNV, amplification G-score correlations range from $0.908$ to $0.919$,
compared with $0.952$ between real training and held-out data.
Deletion correlations are $0.877$ for both TMO-Net and ISOMERIC and $0.815$
for CustOmics, compared with $0.969$ between real cohorts
(Figure~\ref{fig:before-treatment-benchmark}\textbf{(d)}).
To assess CNV fidelity at the 14 prespecified biomarkers, we also measure
absolute G-score errors. Across models, mean absolute errors range from
$0.066$ to $0.070$ for amplification and from $0.090$ to $0.126$ for deletion,
compared with $0.041$ and $0.033$, respectively, between real training and
held-out profiles (Appendix Table~\ref{tab:biomarker-fidelity}).
Appendix Tables~\ref{tab:before-biomarker-amp}--\ref{tab:after-biomarker-del}
report G-scores, signed biases, and absolute errors for each biomarker in both cohorts.
Figure~\ref{fig:best-visual-diagnostics}\textbf{(a,b)} illustrates RNA profiles in PCA space and CNV G-scores for six biomarker genes. Overall, high CNV G-score correlations coexist with systematically underestimated RNA variability.

\textbf{R2. Is RNA co-expression preserved?}
TMO-Net has the highest observed mean edge F1, $0.474\pm0.010$,
compared with $0.585\pm0.035$ between real cohorts
(Figure~\ref{fig:before-treatment-benchmark}\textbf{(f)}). This is
$0.81\times$ the reference, versus $0.70\times$ for ISOMERIC and
$0.39\times$ for CustOmics.

All models retain the broad pattern of gene-gene correlations:
Fisher-$z$ concordance ($C_z$) is close to the real-cohort reference
($0.91$--$0.93\times$), with ISOMERIC highest. Yet their networks are too dense, with
density ratios at $1.86\times$ the real-cohort reference ($0.956$) for
TMO-Net, $2.69\times$ for ISOMERIC, and $7.02\times$ for CustOmics.
CustOmics recovers many held-out links, but most of its inferred links
are absent from the held-out network, resulting in low precision and F1.
Figure~\ref{fig:best-visual-diagnostics}\textbf{(d)} shows the agreement
in correlations and the mismatch in recovered links. Further network
comparisons are provided in Appendix Tables~\ref{tab:before-coexpression-full}
and~\ref{tab:after-coexpression-full}.
In the before-treatment cohort, all three models retain mean network density
ratios above one across correlation thresholds from $0.60$ to $0.80$.
The observed ordering of mean edge F1 also remains unchanged
(Appendix~\ref{app:coexpression-threshold-sensitivity}).

Overall, broad correlation agreement coexists with overly dense co-expression networks.
Generating profiles from decoder means omits residual variation and may
contribute to low RNA variance and overly dense co-expression networks.
The experiment in Appendix~\ref{app:stochastic-decoder} tests this explanation
by adding independent Gaussian noise calibrated from training residuals to
the same generated RNA profiles. Mean RNA variance ratios move closer to one,
and mean network density decreases. However, mean edge F1 decreases for
ISOMERIC and CustOmics. Lower network density therefore does not necessarily
indicate better recovery of co-expression structure.

\textbf{R3. Are the cis RNA-CNV associations retained?}
The RNA-CNV panel intersections contain 320, 313, and 313 genes in the three outer folds, respectively, all with estimable slopes $\beta_g$ from Equation~\ref{eq:cis-rna-cnv} in every real and synthetic cohort. We compare these slopes using their correlation ($r_\beta$) and median absolute difference ($\operatorname{median}|\Delta\beta|$).
TMO-Net has the highest observed mean $r_\beta$ ($0.600\pm0.047$), compared with
$0.743\pm0.046$ between real cohorts: $0.81\times$ the reference,
versus $0.76\times$ for CustOmics and $0.74\times$ for ISOMERIC
(Figure~\ref{fig:before-treatment-benchmark}\textbf{(b)}).
Figure~\ref{fig:best-visual-diagnostics}\textbf{(c)} illustrates the
agreement of individual gene slopes.

ISOMERIC has the lowest absolute slope error, at $0.118\pm0.011$ versus
$0.093\pm0.006$ between real cohorts, or $1.26\times$ the reference.
TMO-Net and CustOmics reach $1.63\times$ and $2.31\times$, respectively
(Appendix Table~\ref{tab:before-integration-full}). Observed means favor
TMO-Net for slope correlation and ISOMERIC for absolute slope error.

\textbf{R4. Do generated profiles raise memorization concerns?}
The diagnostics raise memorization concerns, most clearly for TMO-Net.
Its paired distance ratio ($R_{\mathrm{paired}}=0.958\pm0.016$) and
membership AUC ($0.657\pm0.043$) indicate greater proximity to training
than held-out samples (Figure~\ref{fig:before-treatment-benchmark}\textbf{(c,e)}).
The signal is stronger in RNA
(AUC $=0.804\pm0.035$) than CNV ($0.554\pm0.058$).

Paired scores are less concerning for ISOMERIC and CustOmics, but their RNA AUCs ($0.647$ and $0.689$, respectively) also indicate training preference (Appendix Table~\ref{tab:before-training-diagnostics-full}). 
TMO-Net is also the largest adaptation, with approximately 4.2 million trainable parameters (Appendix~\ref{app:model-training}). 
This motivates testing whether model size contributes to overfitting, but the present comparison does not isolate its effect from other modeling and training choices.
Overall, TMO-Net's advantages in fidelity and coherence coexist with the strongest preference for training data, particularly in RNA.

We added two simple controls to assess whether the evaluation detects
disrupted RNA-CNV pairing (Appendix~\ref{app:simple-controls}).
The shuffling control reassigns CNV profiles between held-out patients within
the same tissue and treatment phase, while keeping RNA profiles fixed.
This permutation should preserve modality-specific fidelity but reduce
RNA-CNV slope agreement. The observed slope correlations fall close to zero
in both cohorts. The second control independently resamples RNA and CNV
profiles from different training patients within each tissue group.
This baseline has lower mean RNA Wasserstein distance and higher mean edge F1
than the three adapted VAEs in both cohorts, but its slope correlations
remain near zero.

\section{Conclusion}
\label{sec:conclusion}

We introduced a protocol for evaluating generation of paired RNA
and CNV profiles using statistical and biological metrics.
The protocol keeps training and evaluation patients disjoint and
interprets fidelity and RNA-CNV coherence relative to real train-test
agreement. This empirical reference contextualizes discrepancies between
synthetic and held-out data against variation between real cohorts.
We focused on VAEs because their latent-variable formulations provide a
common framework for adapting different RNA-CNV coupling mechanisms to the
same de novo generation task. ISOMERIC uses shared and modality-specific
latent variables. TMO-Net generates RNA conditionally on CNV. CustOmics uses
a central joint latent representation. 
We illustrate the protocol through a practical case study of ovarian HGSC
in the DECIDER cohort. The contrasting model results identify specific
targets for development that marginal fidelity alone would miss.
For ISOMERIC, a priority is to reduce excess co-expression links while
retaining its comparatively accurate RNA-CNV association magnitudes.
For TMO-Net, greater proximity to training samples,
particularly in RNA, motivates testing regularization focused on RNA generation.
For CustOmics, low RNA variance and excessive co-expression motivate learning
and sampling variation around decoder means. 

\textbf{Limitations and perspectives.}
The focus on VAEs allows us to compare
different coupling strategies while keeping the generative family fixed.
However, it does not assume that VAEs are universally optimal for synthetic multi-omics
generation. Comparisons with generative adversarial networks, normalizing
flows, and diffusion models would extend the benchmark to other generative
families.
Beyond the variational autoencoders studied here, our evaluation framework
could also be applied, for example, to multiomic foundation models that generate paired
molecular profiles. This perspective aligns with recent calls
for benchmarks that better capture the biological relevance of biomedical
foundation models \citep{saezrodriguez2026benchmarking}.
In this paper, the distance-based membership AUC provides an initial assessment of membership
disclosure risk. Further evaluation could compare several membership inference
attacks under explicit assumptions about the attacker's access, following the
CAMDA Health Privacy Challenge benchmark \citep{ozturk2026syntheticomics}.
These attacks test whether an individual's participation in training can be
inferred from synthetic data. An absence of detected training preference
does not guarantee privacy.
Model size and training objectives differ across the evaluated adaptations.
Future comparisons with matched parameter counts and targeted
training-objective ablations could help clarify the contribution of
RNA-CNV coupling.
Also, the generators do not condition on tissue origin,
treatment phase (e.g., before versus after treatment), or other clinical
covariates. Future work could explore conditional and longitudinal generative
approaches that share information across these contexts while accounting for
dependence between repeated measurements from the same patient.
Such extensions would also enable evaluation of differential expression
between groups defined by tissue origin or treatment phase.
Finally, adding molecular modalities such as somatic mutations or DNA methylation
could further broaden biological evaluations and use cases.

\clearpage
\bibliographystyle{plainnat}
\bibliography{references}

\clearpage
\appendix
\textbf{Organization of the appendices.}
\begin{itemize}
  \item Appendix~\ref{app:generation-dags}: prior-sampling graphs.
  \item Appendix~\ref{app:model-training}: training objectives and settings.
  \item Appendix~\ref{app:evaluation-metrics}: metric definitions.
  \item Appendix~\ref{app:synthetic-data-evaluation}: detailed evaluation
  tables for both cohorts.
  \item Appendix~\ref{app:paired-comparisons}: paired model differences across outer folds.
  \item Appendix~\ref{app:after-treatment-diagnostics}: after-treatment
  benchmark and visual diagnostics.
  \item Appendices~\ref{app:model-visual-diagnostics}
  and~\ref{app:model-visual-diagnostics-after}: visual diagnostics for each
  model in the before- and after-treatment cohorts, respectively.
  \item Appendix~\ref{app:stochastic-decoder}: stochastic RNA decoding, variance, and correlations.
  \item Appendix~\ref{app:simple-controls}: shuffled-pair controls and an independent-modality baseline.
  \item Appendix~\ref{app:coexpression-threshold-sensitivity}: sensitivity to the co-expression edge threshold.
\end{itemize}

\section{Prior sampling graphs}
\label{app:generation-dags}

\begin{figure}[H]
  \centering
  \resizebox{\linewidth}{!}{\begin{tikzpicture}[
  x=1cm,
  y=1cm,
  latent/.style={
    circle,
    draw=black,
    line width=0.6pt,
    fill=white,
    minimum size=7.5mm,
    inner sep=0pt,
    font=\small
  },
  profile/.style={
    rounded corners=2pt,
    draw=black,
    line width=0.6pt,
    fill=violet!8,
    minimum width=18mm,
    minimum height=8mm,
    align=center,
    font=\scriptsize
  },
  representation/.style={
    rounded corners=2pt,
    draw=black,
    line width=0.6pt,
    fill=black!5,
    minimum width=14mm,
    minimum height=7mm,
    align=center,
    font=\scriptsize
  },
  flow/.style={-{Latex[length=2.4mm,width=1.6mm]}, line width=0.85pt},
  edge label/.style={font=\scriptsize, fill=white, inner sep=1pt},
  panel title/.style={font=\small\bfseries}
]
  \draw[rounded corners=3pt, draw=black!45] (0,0) rectangle (2.30,5.10);
  \node[panel title] at (1.15,5.42) {(a) Standard VAE};
  \node[latent] (vae-z) at (1.15,4.30) {$z$};
  \node[latent, fill=violet!8] (vae-x) at (1.15,0.55) {$x$};
  \draw[flow] (vae-z) -- node[midway, edge label]
    {$p_\theta(x\mid z)$} (vae-x);

  \draw[rounded corners=3pt, draw=black!45] (2.55,0) rectangle (6.65,5.10);
  \node[panel title] at (4.60,5.42) {(b) ISOMERIC};
  \node[latent] (iso-z) at (3.50,4.30) {$z$};
  \node[latent] (iso-w) at (5.40,4.30) {$w$};
  \node[profile] (iso-r) at (3.60,0.55)
    {RNA\\$\widetilde r=D_R(z,w)$};
  \node[profile] (iso-c) at (5.60,0.55)
    {CNV\\$\widetilde c=D_C(w)$};
  \draw[flow] (iso-z) -- (iso-r);
  \draw[flow] (iso-w) -- (iso-r);
  \draw[flow] (iso-w) -- (iso-c);

  \draw[rounded corners=3pt, draw=black!45] (6.90,0) rectangle (9.45,5.10);
  \node[panel title] at (8.175,5.42) {(c) TMO-Net};
  \node[latent] (tmo-zc) at (8.175,4.55) {$z_C$};
  \node[profile] (tmo-c) at (8.175,3.35) {CNV\\$\widetilde c=D_C(z_C)$};
  \node[latent] (tmo-zx) at (8.175,1.75) {$z_X$};
  \node[profile] (tmo-r) at (8.175,0.55) {RNA\\$\widetilde r=D_R(z_X)$};
  \draw[flow] (tmo-zc) -- (tmo-c);
  \draw[flow] (tmo-c) -- node[midway, right=1mm, edge label]
    {$q_{\phi_{C\rightarrow R}}$} (tmo-zx);
  \draw[flow] (tmo-zx) -- (tmo-r);

  \draw[rounded corners=3pt, draw=black!45] (9.70,0) rectangle (13.80,5.10);
  \node[panel title] at (11.75,5.42) {(d) CustOmics};
  \node[latent] (cust-z) at (11.75,4.35) {$z$};
  \node[representation] (cust-hr) at (10.70,2.65) {$\widetilde h_R$};
  \node[representation] (cust-hc) at (12.80,2.65) {$\widetilde h_C$};
  \node[profile] (cust-r) at (10.70,0.55)
    {RNA\\$\widetilde r=D_R(\widetilde h_R)$};
  \node[profile] (cust-c) at (12.80,0.55)
    {CNV\\$\widetilde c=D_C(\widetilde h_C)$};
  \draw[flow] (cust-z) -- node[midway, left=1mm, edge label] {$G$} (cust-hr);
  \draw[flow] (cust-z) -- node[midway, right=1mm, edge label] {$G$} (cust-hc);
  \draw[flow] (cust-hr) -- (cust-r);
  \draw[flow] (cust-hc) -- (cust-c);
\end{tikzpicture}
}
  \caption{
  \textbf{Generation from the prior in a standard VAE and the three adaptations.}
  \textbf{(a)} The VAE samples $z\sim p(z)$, then $x\sim p_\theta(x\mid z)$.
  \textbf{(b)} ISOMERIC samples independent $z$ and $w$, $w$ contributes to
  both profiles and $z$ only to RNA.
  \textbf{(c)} TMO-Net decodes CNV from $z_C$, then uses the generated CNV to
  parameterize and sample $z_X=z_{C\rightarrow R}$ before decoding RNA.
  \textbf{(d)} CustOmics decodes one central $z$ into two representations, one
  for each modality, and then into RNA and CNV.
  White circles denote sampled Gaussian latents, the shaded circle denotes a
  sampled observation, and boxes denote deterministic decoder outputs.
  Arrows denote computational dependence rather than causality.}
  \label{fig:generation-dags}
\end{figure}
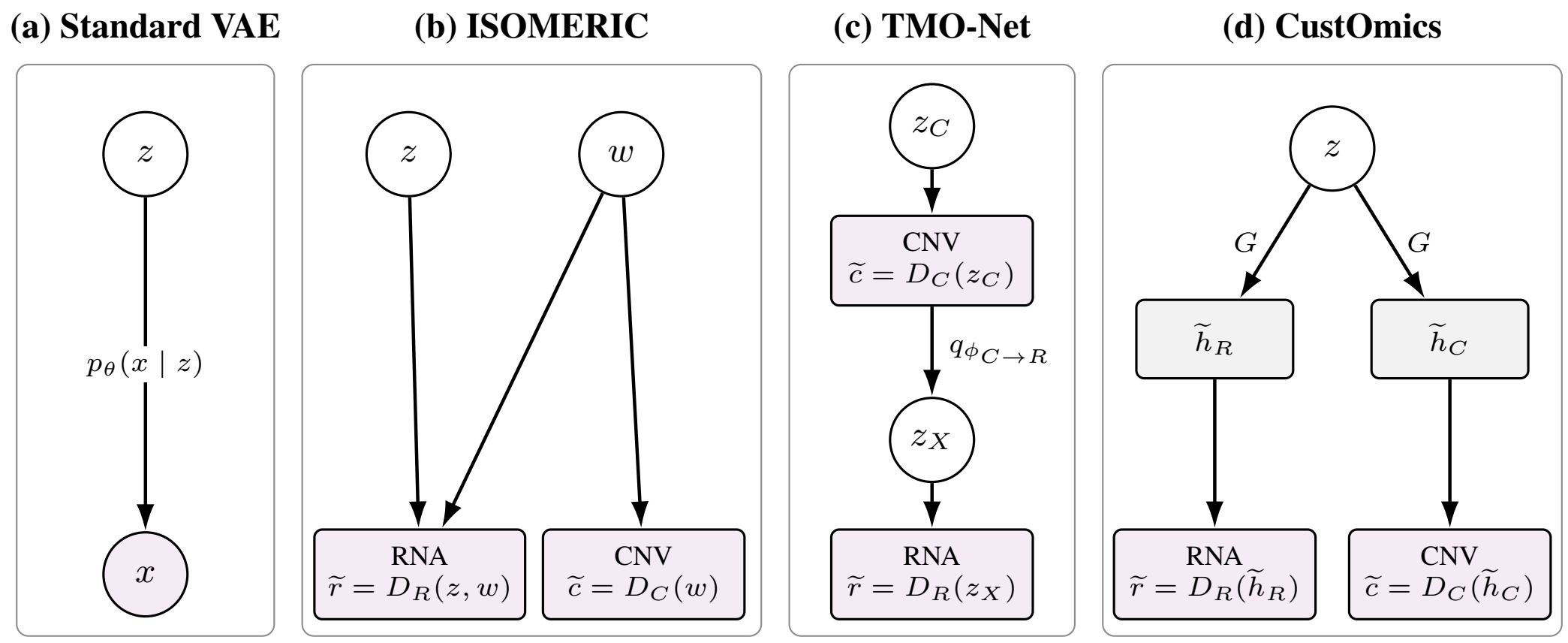

Figure~\ref{fig:generation-dags} translates the sampling equations in
Section~\ref{sec:models} into directed acyclic computational graphs. We omit
paths and objectives used only during training. The encoder from CNV to RNA is
shown because generation uses it.

\section{Model training details}
\label{app:model-training}

\textbf{Common setup.}
All models are trained from scratch on each paired outer training set,
without pan-cancer pretraining or patient, sampling-site, or treatment-phase covariates.
Each fit uses 3,000 RNA and 3,000 CNV variables from 313--341 primary or
86--96 post-treatment samples, depending on the outer split, and is repeated
with three random initializations. For the objectives below, $\ell(a,b)$
denotes mean-squared error in the scaled feature space, and
$\operatorname{KL}$ denotes Kullback--Leibler divergence.
Per-observation objectives are averaged over the training set, reconstruction
expectations are estimated using sampled latents.
The standard negative ELBO has unit KL weight:
\begin{equation}
\begin{aligned}
\mathcal{L}_{\mathrm{VAE}}(x)
&=\mathbb{E}_{q_\phi(z\mid x)}[-\log p_\theta(x\mid z)]\\
&+\operatorname{KL}\!\left(q_\phi(z\mid x)\,\|\,p(z)\right).
\end{aligned}
\label{eq:vae-loss}
\end{equation}
The coefficients below define the weighted objectives used by the adaptations.

\begin{figure}[H]
  \centering
  \resizebox{\linewidth}{!}{\begin{tikzpicture}[
  x=1cm,
  y=1cm,
  every node/.style={font=\scriptsize, inner sep=2pt},
  observed/.style={
    rounded corners=2pt,
    draw=black,
    line width=0.6pt,
    fill=violet!12,
    minimum width=16mm,
    minimum height=7mm,
    align=center
  },
  module/.style={
    rounded corners=2pt,
    draw=black,
    line width=0.6pt,
    fill=blue!7,
    minimum width=15mm,
    minimum height=7mm,
    align=center
  },
  posterior/.style={
    rounded corners=2pt,
    draw=black,
    line width=0.6pt,
    fill=green!8,
    minimum width=25mm,
    minimum height=8mm,
    align=center
  },
  latent/.style={
    circle,
    draw=black,
    line width=0.6pt,
    fill=white,
    minimum size=7mm,
    inner sep=0pt
  },
  output/.style={
    rounded corners=2pt,
    draw=black,
    line width=0.6pt,
    fill=black!4,
    minimum width=18mm,
    minimum height=7mm,
    align=center
  },
  loss/.style={
    rounded corners=2pt,
    draw=orange!70!black,
    line width=0.6pt,
    fill=orange!9,
    minimum width=27mm,
    minimum height=7mm,
    align=center
  },
  flow/.style={-{Latex[length=1.8mm,width=1.2mm]}, line width=0.7pt},
  loss edge/.style={-{Latex[length=1.6mm,width=1.1mm]}, dashed,
    draw=orange!80!black, line width=0.7pt},
  shared/.style={densely dotted, draw=violet!80!black, line width=0.8pt},
  panel/.style={rounded corners=3pt, draw=black!45, line width=0.5pt},
  panel title/.style={font=\small\bfseries, anchor=west},
  group label/.style={font=\scriptsize\itshape, text=black!70},
  note/.style={font=\tiny, align=center, text=black!75}
]
  \draw[panel] (0,10.60) rectangle (15.20,15.05);
  \node[panel title] at (0.30,14.72) {(a) ISOMERIC};

  \node[observed] (iso-r) at (1.00,12.65) {$r$};
  \node[module] (iso-e) at (2.90,12.65) {shared RNA\\encoder $E_R$};
  \node[posterior] (iso-qz) at (5.20,13.40)
    {Gaussian head $q_z(z\mid r)$\\$\mathrm{KL}(q_z\|\mathcal N(0,I))$};
  \node[posterior] (iso-qw) at (5.20,11.90)
    {Gaussian head $q_w(w\mid r)$\\$\mathrm{KL}(q_w\|\mathcal N(0,I))$};
  \node[latent] (iso-z) at (7.20,13.40) {$z$};
  \node[latent] (iso-w) at (7.20,11.90) {$w$};

  \node[module, minimum width=17mm] (iso-a) at (8.65,13.90) {$A_C$\\linear, train only};
  \node[module, minimum width=17mm] (iso-dr) at (8.65,12.65) {RNA decoder\\$D_R(z,w)$};
  \node[module, minimum width=17mm] (iso-dc) at (8.65,11.40) {linear CNV\\decoder $D_C(w)$};
  \node[output] (iso-cz) at (10.70,13.90) {$\widehat c_z$};
  \node[output] (iso-rhat) at (10.70,12.65) {$\widehat r$};
  \node[output] (iso-chat) at (10.70,11.40) {$\widehat c_w$};
  \node[loss] (iso-ladv) at (13.25,13.90)
    {$\mathrm{MSE}(\widehat c_z,c)$\\$A_C$: min., $E_R,q_z$: max.};
  \node[loss] (iso-lr) at (13.25,12.65) {$\mathrm{MSE}(\widehat r,r)$};
  \node[loss] (iso-lc) at (13.25,11.40) {$\mathrm{MSE}(\widehat c_w,c)$};

  \draw[flow] (iso-r) -- (iso-e);
  \draw[flow] (iso-e.east) -- (iso-qz.west);
  \draw[flow] (iso-e.east) -- (iso-qw.west);
  \draw[flow] (iso-qz) -- (iso-z);
  \draw[flow] (iso-qw) -- (iso-w);
  \draw[flow] (iso-z) -- (iso-a);
  \draw[flow] (iso-z) -- (iso-dr);
  \draw[flow] (iso-w) -- (iso-dr);
  \draw[flow] (iso-w) -- (iso-dc);
  \draw[flow] (iso-a) -- (iso-cz);
  \draw[flow] (iso-dr) -- (iso-rhat);
  \draw[flow] (iso-dc) -- (iso-chat);
  \draw[loss edge] (iso-cz) -- (iso-ladv);
  \draw[loss edge] (iso-rhat) -- (iso-lr);
  \draw[loss edge] (iso-chat) -- (iso-lc);
  \node[note, anchor=west] at (0.30,10.92)
    {Paired CNV $c$ is used only as the target of the two CNV losses.};

  \draw[panel] (0,5.40) rectangle (15.20,10.25);
  \node[panel title] at (0.30,9.92) {(b) TMO-Net};

  \node[observed] (tmo-c) at (1.00,8.50) {$c$};
  \node[observed] (tmo-r) at (1.00,6.25) {$r$};
  \node[module] (tmo-ec) at (2.90,9.10) {$E_C$};
  \node[module] (tmo-ex) at (2.90,7.70) {$E_{C\rightarrow R}$};
  \node[module] (tmo-er) at (2.90,6.25) {$E_R$};
  \node[posterior] (tmo-qc) at (5.20,9.10)
    {$q_C(z_C\mid c)$\\$\mathrm{KL}(q_C\|\mathcal N(0,I))$};
  \node[posterior] (tmo-qx) at (5.20,7.70)
    {$q_{C\to R}$ from CNV\\$\mathrm{KL}(q_{C\to R}\|\mathcal N(0,I))$};
  \node[posterior] (tmo-qr) at (5.20,6.25)
    {$q_R(z_R\mid r)$\\$\mathrm{KL}(q_R\|\mathcal N(0,I))$};
  \node[latent] (tmo-zc) at (7.20,9.10) {$z_C$};
  \node[latent, minimum size=8mm] (tmo-zx) at (7.20,7.70) {$z_{C\to R}$};
  \node[latent] (tmo-zr) at (7.20,6.25) {$z_R$};
  \node[module] (tmo-dc) at (8.65,9.10) {$D_C$};
  \node[module] (tmo-drx) at (8.65,7.70) {$D_R$};
  \node[module] (tmo-drr) at (8.65,6.25) {$D_R$};
  \node[output] (tmo-chat) at (10.70,9.10) {$\widehat c$};
  \node[output] (tmo-rx) at (10.70,7.70) {$\widehat r_{C\to R}$};
  \node[output] (tmo-rr) at (10.70,6.25) {$\widehat r_{\rm self}$};
  \node[loss] (tmo-lc) at (13.25,9.10) {$\mathrm{MSE}(\widehat c,c)$};
  \node[loss] (tmo-lx) at (13.25,7.70) {$\mathrm{MSE}(\widehat r_{C\to R},r)$};
  \node[loss] (tmo-lr) at (13.25,6.25) {$\mathrm{MSE}(\widehat r_{\rm self},r)$};

  \draw[flow] (tmo-c.east) -- ++(0.17,0) |- (tmo-ec.west);
  \draw[flow] (tmo-c.east) -- ++(0.17,0) |- (tmo-ex.west);
  \draw[flow] (tmo-r) -- (tmo-er);
  \draw[flow] (tmo-ec) -- (tmo-qc);
  \draw[flow] (tmo-ex) -- (tmo-qx);
  \draw[flow] (tmo-er) -- (tmo-qr);
  \draw[flow] (tmo-qc) -- (tmo-zc);
  \draw[flow] (tmo-qx) -- (tmo-zx);
  \draw[flow] (tmo-qr) -- (tmo-zr);
  \draw[flow] (tmo-zc) -- (tmo-dc);
  \draw[flow] (tmo-zx) -- (tmo-drx);
  \draw[flow] (tmo-zr) -- (tmo-drr);
  \draw[flow] (tmo-dc) -- (tmo-chat);
  \draw[flow] (tmo-drx) -- (tmo-rx);
  \draw[flow] (tmo-drr) -- (tmo-rr);
  \draw[loss edge] (tmo-chat) -- (tmo-lc);
  \draw[loss edge] (tmo-rx) -- (tmo-lx);
  \draw[loss edge] (tmo-rr) -- (tmo-lr);
  \draw[shared] (tmo-drx.south) -- node[right=0.6mm, note]
    {shared\\weights} (tmo-drr.north);
  \draw[loss edge]
    (tmo-qx.south) -- node[left=0.7mm, note]
    {$\mathrm{KL}(q_{C\to R}\|q_R)$} (tmo-qr.north);
  \node[note, anchor=east] at (14.90,5.72)
    {Current adaptation: no discriminator};

  \draw[panel] (0,-0.65) rectangle (15.20,5.05);
  \node[panel title] at (0.30,4.72) {(c) CustOmics};
  \node[group label, anchor=west] at (0.30,4.28)
    {Modality autoencoders: phases 1 and 2};
  \node[group label] at (9.25,1.73)
    {Central MMD-VAE: phase 2 only};

  \node[observed] (cust-r) at (1.00,3.65) {$r$};
  \node[observed] (cust-c) at (1.00,2.50) {$c$};
  \node[module] (cust-er) at (2.90,3.65) {$E_R$};
  \node[module] (cust-ec) at (2.90,2.50) {$E_C$};
  \node[output, minimum width=10mm] (cust-hr) at (4.80,3.65) {$h_R$};
  \node[output, minimum width=10mm] (cust-hc) at (4.80,2.50) {$h_C$};
  \node[module] (cust-dr) at (6.60,3.65) {$D_R$};
  \node[module] (cust-dc) at (6.60,2.50) {$D_C$};
  \node[output] (cust-rr) at (8.65,3.65) {$\widehat r_{\rm self}$};
  \node[output] (cust-cc) at (8.65,2.50) {$\widehat c_{\rm self}$};
  \node[loss] (cust-lr) at (11.20,3.65) {$\mathrm{MSE}(\widehat r_{\rm self},r)$};
  \node[loss] (cust-lc) at (11.20,2.50) {$\mathrm{MSE}(\widehat c_{\rm self},c)$};

  \node[output] (cust-h) at (3.65,1.00)
    {$h=[h_R,h_C]$};
  \node[module] (cust-ez) at (5.60,1.00) {central\\encoder};
  \node[posterior, minimum width=16mm] (cust-q) at (7.60,1.00) {$q_\phi(z\mid h)$};
  \node[latent] (cust-z) at (9.25,1.00) {$z$};
  \node[module] (cust-g) at (10.85,1.00) {central\\decoder $G$};
  \node[output, minimum width=19mm] (cust-hhat) at (12.90,1.00)
    {$\widehat h=[\widehat h_R,\widehat h_C]$};
  \node[loss, minimum width=22mm] (cust-lh) at (12.90,-0.10)
    {$\mathrm{MSE}(\widehat h,h)$};
  \node[loss, minimum width=34mm] (cust-mmd) at (8.40,-0.10)
    {$\mathrm{MMD}^2(q^{\rm agg}_\phi,p)$, $p=\mathcal N(0,I)$};

  \draw[flow] (cust-r) -- (cust-er);
  \draw[flow] (cust-c) -- (cust-ec);
  \draw[flow] (cust-er) -- (cust-hr);
  \draw[flow] (cust-ec) -- (cust-hc);
  \draw[flow] (cust-hr) -- (cust-dr);
  \draw[flow] (cust-hc) -- (cust-dc);
  \draw[flow] (cust-dr) -- (cust-rr);
  \draw[flow] (cust-dc) -- (cust-cc);
  \draw[loss edge] (cust-rr) -- (cust-lr);
  \draw[loss edge] (cust-cc) -- (cust-lc);
  \draw[flow, preaction={draw=white, line width=2.8pt, -}]
    (cust-hr.south) -- (4.80,3.08) -- (3.95,3.08)
    -- (3.95,1.70) -- (cust-h.north);
  \draw[flow] (cust-hc.south) -- (cust-h.north east);
  \draw[flow] (cust-h) -- (cust-ez);
  \draw[flow] (cust-ez) -- (cust-q);
  \draw[flow] (cust-q) -- (cust-z);
  \draw[flow] (cust-z) -- (cust-g);
  \draw[flow] (cust-g) -- (cust-hhat);
  \draw[loss edge] (cust-hhat) -- (cust-lh);
  \draw[loss edge] (cust-z.south) -- (cust-mmd.north);
\end{tikzpicture}
}
  \caption{
  \textbf{Training architectures: (a) ISOMERIC, (b) TMO-Net, and (c) CustOmics.}
  Observed RNA and CNV profiles are denoted by $r$ and $c$, hats denote
  reconstructions. Green boxes denote Gaussian approximate posteriors and
  circles denote reparameterized samples. Solid arrows show forward
  computations, orange dashed arrows show objective terms.
  See Section~\ref{sec:models} for model definitions.}
  \label{fig:training-architectures}
\end{figure}
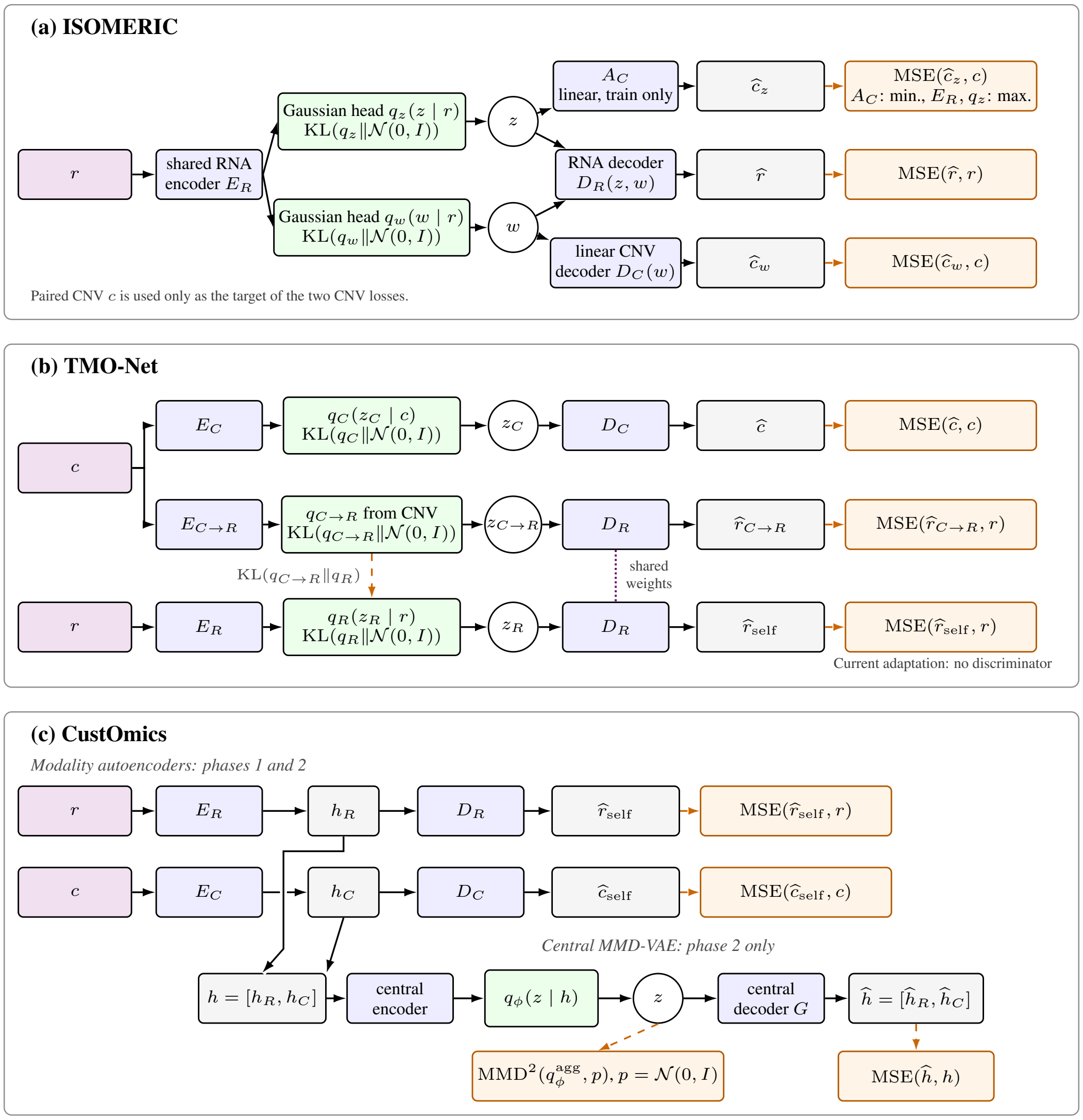

Figure~\ref{fig:training-architectures} summarizes the corresponding training
graphs. It uses the paired observations available during training and is
distinct from the prior-sampling graphs in
Figure~\ref{fig:generation-dags}.

\textbf{ISOMERIC.}
The encoder--decoder parameters minimize
\begin{align}
\mathcal{L}_{\mathrm{ISO}}={}&
\ell(D_R(z,w),r)
\notag\\
&+\beta_z\operatorname{KL}\!\left(q_{\phi_z}(z\mid r)\,\|\,p(z)\right)\notag\\
&+\beta_w\operatorname{KL}\!\left(q_{\phi_w}(w\mid r)\,\|\,p(w)\right) \notag\\
&+\ell(D_C(w),c)\notag\\
&-\beta_{\mathrm{adv}}\ell(A_C(z),c),
\label{eq:isomeric-loss}
\end{align}
while the adversary separately minimizes
$\ell(A_C(\operatorname{stopgrad}(z)),c)$.
The negative term therefore encourages, but does not guarantee, separation of
CNV information from $z$. The selected model uses
$\beta_z=\beta_w=10^{-2}$, $\beta_{\mathrm{adv}}=0.1$, one 128-unit hidden
layer, and 10 dimensions for each of $z$ and $w$. It has 845,488 trainable
parameters and is trained for 300 epochs with Adam, learning rate $10^{-3}$,
and batch size 16.

\textbf{TMO-Net.}
Writing $q_C,q_R,q_X$ for the three approximate posteriors in
Equation~\ref{eq:tmonet-paths} and $p_C,p_R,p_X$ for their standard Gaussian
priors, training minimizes
\begin{equation}
\begin{aligned}
\mathcal{L}_{\mathrm{TMO}}={}&
\mathbb{E}\!\left[
\ell(c,\widehat c)+\ell(r,\widehat r_{\mathrm{self}})\right.\\
&\left.\qquad+\ell(r,\widehat r_{C\rightarrow R})\right] \\
&+\sum_{a\in\{C,R,X\}}\beta_a\operatorname{KL}(q_a\,\|\,p_a)
\\
&+\beta_A\operatorname{KL}(q_X\,\|\,q_R).
\end{aligned}
\label{eq:tmonet-loss}
\end{equation}
The expectation is over the three latent draws. The final term aligns the
CNV-derived RNA posterior with the RNA self-posterior.
All reconstruction weights equal one. Across outer splits, the selected KL weights
are $10^{-3}$ or $3\times10^{-3}$ and the alignment weight is $10^{-3}$ or
$10^{-2}$, these weights increase linearly during the first 100 of 300 training
epochs. Each encoder and decoder has two 256-unit hidden layers. The selected
latent dimension is 32 in the primary cohort and ranges from 16 to 64 after
treatment, giving 4.21--4.31 million trainable parameters, optimization uses
Adam with learning rate $5\times10^{-4}$ and batch size 16.

\textbf{CustOmics.}
Training has two phases. The first fits the two modality autoencoders for 150
epochs with
\[
\begin{aligned}
\mathcal{L}_1
&=\frac{1}{n}\sum_{i=1}^{n}
\left[\ell(D_R(E_R(r_i)),r_i)\right.\\
&\left.\qquad+\ell(D_C(E_C(c_i)),c_i)\right].
\end{aligned}
\]
The second jointly fine-tunes all modules for another 150 epochs with
\[
\begin{aligned}
\mathcal{L}_2
&=\mathcal{L}_1
+\frac{1}{n}\sum_{i=1}^{n}
\mathbb{E}_{q_\phi(z\mid h_i)}[\ell(G(z),h_i)]\\
&+\lambda_{\mathrm{MMD}}\operatorname{MMD}^{2}\!\left(q_\phi^{\mathrm{agg}},p\right).
\end{aligned}
\]
Here $h_i=[E_R(r_i),E_C(c_i)]$, and
\[
q_\phi^{\mathrm{agg}}(z)=n^{-1}\sum_{i=1}^{n}q_\phi(z\mid h_i)
\]
is the aggregated posterior over $n$ training observations.
MMD encourages this mixture to match the Gaussian prior $p(z)$.
Unlike the per-observation KL in a standard VAE, this penalty acts on the
mixture rather than on each posterior separately.
The implementation estimates squared MMD between minibatch posterior samples
and independent standard Gaussian samples, retaining diagonal kernel terms.
It uses the original CustOmics radial basis function kernel
$k(u,v)=\exp(-\|u-v\|^2/d_z^2)$, where $d_z$ is the central latent dimension.
The selected baseline assigns unit weight to these terms and does not add a
direct RNA or CNV reconstruction loss through the central path. Each modality
autoencoder uses widths $3000$--$256$--$128$--$32$ and a mirrored decoder, the
64-dimensional concatenated representation is mapped through central hidden
layers of widths 128 and 64 to a 32-dimensional latent. The model has 3.27
million trainable parameters and uses Adam with learning rate $10^{-3}$ and
batch size 16.

\section{Evaluation metric details}
\label{app:evaluation-metrics}

\subsection{Marginal CNV fidelity}
\label{app:cnv-fidelity}
\label{app:cnv_preprocessing}

Following the purity and ploidy adjustment described in GenomeSpy
\citep{lavikka2024genomespy}, let $R_{is}>0$ denote the observed
linear copy ratio for segment $s$ in sample $i$. The adjusted ratio is
\[
\begin{gathered}
R^{\mathrm{pur}}_{is}
=
\frac{
\rho_i\psi_iR_{is}
+
2(1-\rho_i)(R_{is}-1)
}{
\rho_i\psi_i
},
\\
x_{is}=\log_2 R^{\mathrm{pur}}_{is},
\end{gathered}
\]
where $\rho_i$ is the estimated tumor purity and $\psi_i$ is the
estimated tumor ploidy. We map segments to genes by assigning gene
$g$ the minimum adjusted log ratio among its overlapping segments:
\[
x_{ig}
=
\min_{s:\,s\cap g\neq\varnothing} x_{is}.
\]
Thus, $x_{ig}=0$ denotes the adjusted reference copy-ratio state
rather than zero genomic copies. Positive and negative values indicate
relative copy-number gains and losses, respectively.

Event frequencies alone ignore alteration magnitude. We therefore adapt the
amplitude and frequency summary implemented in GenomeSpy
\citep{lavikka2024genomespy} from genomic segments to gene-level CNV features.
For each cohort, we compute the amplification and deletion G-scores defined
in Equation~\eqref{eq:cnv-gscore}. The alteration threshold is $\tau=0.7$,
and $\kappa=2.5$ caps the contribution of extreme values.
Across all 3,000 CNV genes per fold, we compare synthetic and held-out scores
using gene-wise absolute differences, correlations, and overlap among the
50 highest-scoring genes. Real train--held-out agreement provides a reference.

{For the fixed panel $B$ of 14 prespecified biomarkers, we additionally compute
signed differences and mean absolute errors for each alteration type
$a\in\{\mathrm{amp},\mathrm{del}\}$:
\begin{equation}
\begin{aligned}
\Delta^a_{gfs} &= G^a_{gfs,\mathrm{synth}}-G^a_{gf,\mathrm{test}},\\
\mathrm{MAE}^a_{Bfs} &= \frac{1}{14}\sum_{g\in B}|\Delta^a_{gfs}|,
\end{aligned}
\label{eq:biomarker-gscore-mae}
\end{equation}
where $f$ indexes outer folds and $s$ indexes model initializations.
All 14 biomarkers are present in every fold.
We first compute absolute errors for each initialization, then average
within fold and report the mean $\pm$ sample standard deviation (SD) of the
three fold means. Per-gene absolute errors follow the same aggregation,
signed differences are retained separately to distinguish overestimation
from underestimation. Taking absolute differences before averaging prevents
opposite errors from cancelling. For the real-data reference, we replace
synthetic G-scores with real-training G-scores and summarize the three
train--held-out comparisons. Deletion G-scores are nonnegative, although
the figures plot them below zero. All errors are in G-score units,
which combine event frequency and capped amplitude, rather than percentage
points of deletion frequency. Table~\ref{tab:biomarker-fidelity} pairs panel
MAEs with the correlations over all 3,000 genes, Tables~\ref{tab:before-biomarker-amp}--\ref{tab:after-biomarker-del}
report the individual biomarker scores and errors.}

\subsection{Marginal RNA fidelity}
\label{app:rna-fidelity}

At the gene level, we compute 1-Wasserstein ($W_1$) and Jensen--Shannon (JS)
distances between real and synthetic expression distributions and report
their means across genes. The correlation of gene means assesses whether
genes with high mean expression in real data also have high mean expression
in synthetic data. The median synthetic-to-real variance ratio summarizes
agreement in variability across samples, a ratio of one is the target
\citep{qian2023synthcity,ozturk2026syntheticomics}.

At the sample level, principal component analysis (PCA) provides a common
space for comparing multivariate RNA profiles. We compute radial basis
function maximum mean discrepancy (RBF MMD$^2$) in this space to assess
distributional differences, including location, dispersion, and concentration
\citep{gretton2012kernel}.

As a complementary diagnostic, logistic regression distinguishes real from
synthetic profiles using out-of-fold predictions. We summarize the area under
the receiver operating characteristic curve (AUC) as
\[
D_{\mathrm{disc}}=2\left|\mathrm{AUC}-0.5\right|.
\]
A score of zero indicates no separation by this classifier, while a score
of one indicates perfect separation \citep{lopezpaz2016revisiting}.

{The real-data reference applies the same metrics to real training and
held-out profiles, using the same fold-specific features and training-fitted
preprocessing. We summarize these three comparisons without seed averaging,
the reference variance ratio is training/held-out.
Before treatment, held-out discriminative scores range from $0.183$ to
$0.431$, compared with $0.082$ between real cohorts
(Table~\ref{tab:before-fidelity-full}). After treatment, the corresponding
range is $0.291$--$0.600$, compared with $0.0068$
(Table~\ref{tab:after-fidelity-full}). These differences indicate residual
separability under this classifier despite close agreement in gene means.
The reference describes empirical between-cohort variation, it is not a
formal noise floor. The discriminator uses five sample-stratified internal
folds after training-fitted standardization and PCA. These internal folds
are not grouped by patient, so this score remains a descriptive diagnostic,
a low value does not establish distributional equivalence.}

\subsection{Training-proximity metric definitions}
\label{app:memorization}

Good agreement with real RNA and CNV data does not rule out that generated
samples may be unusually close to the training data
\citep{chen2024biologically,ozturk2026syntheticomics}.
For each feature space $m\in\{R,C,\mathrm{pair}\}$ and synthetic
profile $s\in\mathcal S$, let $d_{\mathcal T}^{m}(s)$ and
$d_{\mathcal H}^{m}(s)$ denote its Euclidean nearest-neighbor
distances to equally sized real training and held-out reference
sets, respectively. We summarize the relative proximity to training
using $R_m$ in Equation~\eqref{eq:training-proximity}.
Here $\epsilon$ is double-precision machine epsilon. The reference sets are
subsampled to the same size, and distances are averaged over five reproducible
subsampling repeats before computing the ratio for each synthetic profile.
A value $R_m=1$ is the equal-proximity reference, while values below or above
one indicate training or held-out preference, respectively.

Each feature is standardized using the real training mean and standard deviation.
For paired profiles, we balance the RNA and CNV blocks using their
median leave-one-out nearest-neighbor distances in the real
training set:
\[
D_{\mathrm{pair}}(u,v)^2
=
\left(\frac{D_R(u,v)}{q_R}\right)^2
+
\left(\frac{D_C(u,v)}{q_C}\right)^2.
\]
We also report the fraction of synthetic profiles whose nearest RNA and CNV
neighbors in the real training set identify the same paired sample.

For the membership diagnostic, each real profile $x$ receives the
score
\[
a_m(x)=-\min_{s\in\mathcal S}D_m(x,s),
\]
and we report the area under the receiver operating characteristic curve (AUC),
with real training members as the positive class and held-out non-members
as the negative class \citep{shokri2017membership,vanbreugel2023membership}.
An AUC of $0.5$ indicates no separation by this score, values above $0.5$
indicate that training samples tend to be closer to synthetic profiles,
whereas values below $0.5$ indicate reversed separation, not stronger privacy.
These diagnostics provide a preliminary assessment of training preference and help identify models that warrant further investigation.
They do not establish copying or guarantee privacy, but can guide dedicated analyses of profile memorization and membership inference under explicit threat models \citep{vanbreugel2023membership}.

\subsection{RNA co-expression metric definitions}
\label{app:coexpression}

Within each fold, the real training, held-out, and synthetic cohorts are
subsampled to the same size before network construction.
For each retained gene pair $g<h$, Fisher-$z$ denotes the Fisher
transformation, not a hypothesis test:
\[
z(r)=\operatorname{arctanh}(r)
=\frac{1}{2}\log\!\left(\frac{1+r}{1-r}\right).
\]
For a candidate cohort $C$, either synthetic or real training, and the
held-out real cohort $H$, Fisher-$z$ concordance $C_z$ is defined in
Equation~\eqref{eq:coexpression-concordance}. This calculation uses all retained
gene pairs with finite correlations and is independent of edge selection.
Real train--held-out agreement provides a descriptive reference for
reproducibility between real cohorts.

An edge is retained separately in each network when $r\geq0.70$ and its
Benjamini--Hochberg-adjusted $p$-value is below $0.05$.
For the thresholded edge sets $E_H$ and $E_C$, we define
\[
\begin{aligned}
\mathrm{TP}&=|E_H\cap E_C|,\\
\mathrm{FN}&=|E_H\setminus E_C|,\\
\mathrm{FP}&=|E_C\setminus E_H|,\\
\mathrm{Edge\ F1}&=\frac{2\mathrm{TP}}
{2\mathrm{TP}+\mathrm{FP}+\mathrm{FN}}.
\end{aligned}
\]
Edge F1 is the harmonic mean of precision and recall. Precision is the
fraction of candidate edges present in the held-out network, while recall
is the fraction of held-out edges recovered.
Thus, true negatives, which are pairs selected as edges in neither network,
do not enter Edge F1. Here TP, FP, and FN treat the held-out network as the
evaluation reference, not as biological ground truth.

Module ARI (adjusted Rand index) compares module assignments after separate
Leiden clustering, using only genes assigned to modules in both networks.
Hub Jaccard is the size of the intersection divided by the size of the union
of the sets of 50 highest-degree genes in the two networks.

Tables~\ref{tab:before-coexpression-full} and~\ref{tab:after-coexpression-full}
report network preservation and edge diagnostics for the before-treatment
and after-treatment cohorts, respectively.

\clearpage

\begin{samepage}
\section{Synthetic data evaluation tables}
\label{app:synthetic-data-evaluation}

\begin{itemize}
  \item Tables~\ref{tab:before-fidelity-full}--\ref{tab:after-fidelity-full}:
  marginal fidelity.
  \item {Table~\ref{tab:biomarker-fidelity} and
  Tables~\ref{tab:before-biomarker-amp}--\ref{tab:after-biomarker-del}:
  biomarker G-score fidelity.}
  \item Tables~\ref{tab:before-coexpression-full}--\ref{tab:after-coexpression-full}:
  RNA co-expression.
  \item Tables~\ref{tab:before-integration-full}--\ref{tab:after-integration-full}:
  cis RNA-CNV association.
  \item Tables~\ref{tab:before-training-diagnostics-full}--\ref{tab:after-training-diagnostics-full}:
  training proximity.
\end{itemize}

{Results are reported separately for the before- and after-treatment cohorts.}
Boldface identifies the most favorable model mean for each metric and
comparison, with ties at the displayed precision retained. This ranking is
descriptive and does not indicate statistical significance. Real-data
references are excluded from the ranking.
\end{samepage}

\begin{table*}[!ht]
\centering
\caption{\textbf{Detailed marginal fidelity for the before-treatment primary cohort.} {Each model cell} reports comparison with real training data on the first line and held-out real test data on the second line. Values are the mean $\pm$ sample SD across three patient-disjoint outer folds after averaging three model initializations within each fold. {The final row in each panel compares real training with held-out data once per fold, its variance ratio is training/held-out.} Bold marks the best model mean separately for train and test, following the directions shown in the headers.}
\label{tab:before-fidelity-full}
\scriptsize
\setlength{\tabcolsep}{1pt}
\textbf{(a) RNA distribution fidelity}\\[2pt]
{\renewcommand{\arraystretch}{1.45}
\begin{tabular}{|>{\raggedright\arraybackslash}p{1.55cm}|*{6}{>{\centering\arraybackslash}p{\dimexpr(\textwidth-1.55cm-14\tabcolsep-8\arrayrulewidth)/6\relax}|}}
\hline
\rowcolor{black!12}
\textbf{Method} & \textbf{Mean gene-wise Wasserstein $\downarrow$} & \textbf{Mean JS distance $\downarrow$} & \textbf{RBF MMD$^2$ $\downarrow$} & \textbf{Median variance ratio $\rightarrow 1$} & \textbf{Gene-mean correlation $\uparrow$} & \textbf{Discriminative score $\downarrow$} \\ \hline
TMO-Net & {\renewcommand{\arraystretch}{1.25}\begin{tabular}{@{}c@{}}$ \textbf{0.1389} \pm \textbf{0.0132} $\\$ \textbf{0.1663} \pm \textbf{0.0266} $\end{tabular}} & {\renewcommand{\arraystretch}{1.25}\begin{tabular}{@{}c@{}}$ \textbf{0.2370} \pm \textbf{0.0057} $\\$ \textbf{0.2807} \pm \textbf{0.0116} $\end{tabular}} & {\renewcommand{\arraystretch}{1.25}\begin{tabular}{@{}c@{}}$ 0.0099 \pm 0.0040 $\\$ 0.0157 \pm 0.0089 $\end{tabular}} & {\renewcommand{\arraystretch}{1.25}\begin{tabular}{@{}c@{}}$ \textbf{0.6287} \pm \textbf{0.0559} $\\$ \textbf{0.6562} \pm \textbf{0.0935} $\end{tabular}} & {\renewcommand{\arraystretch}{1.25}\begin{tabular}{@{}c@{}}$ 0.9978 \pm 0.0005 $\\$ 0.9934 \pm 0.0025 $\end{tabular}} & {\renewcommand{\arraystretch}{1.25}\begin{tabular}{@{}c@{}}$ 0.1006 \pm 0.0535 $\\$ 0.2955 \pm 0.0886 $\end{tabular}} \\ \hline
ISOMERIC & {\renewcommand{\arraystretch}{1.25}\begin{tabular}{@{}c@{}}$ 0.1639 \pm 0.0151 $\\$ 0.1838 \pm 0.0197 $\end{tabular}} & {\renewcommand{\arraystretch}{1.25}\begin{tabular}{@{}c@{}}$ 0.2528 \pm 0.0065 $\\$ 0.2904 \pm 0.0130 $\end{tabular}} & {\renewcommand{\arraystretch}{1.25}\begin{tabular}{@{}c@{}}$ \textbf{0.0089} \pm \textbf{0.0026} $\\$ \textbf{0.0125} \pm \textbf{0.0032} $\end{tabular}} & {\renewcommand{\arraystretch}{1.25}\begin{tabular}{@{}c@{}}$ 0.5419 \pm 0.0714 $\\$ 0.5652 \pm 0.1009 $\end{tabular}} & {\renewcommand{\arraystretch}{1.25}\begin{tabular}{@{}c@{}}$ \textbf{0.9991} \pm \textbf{0.0004} $\\$ \textbf{0.9950} \pm \textbf{0.0012} $\end{tabular}} & {\renewcommand{\arraystretch}{1.25}\begin{tabular}{@{}c@{}}$ \textbf{0.0631} \pm \textbf{0.0108} $\\$ \textbf{0.1832} \pm \textbf{0.0626} $\end{tabular}} \\ \hline
CustOmics & {\renewcommand{\arraystretch}{1.25}\begin{tabular}{@{}c@{}}$ 0.1657 \pm 0.0109 $\\$ 0.1854 \pm 0.0134 $\end{tabular}} & {\renewcommand{\arraystretch}{1.25}\begin{tabular}{@{}c@{}}$ 0.2500 \pm 0.0071 $\\$ 0.2886 \pm 0.0009 $\end{tabular}} & {\renewcommand{\arraystretch}{1.25}\begin{tabular}{@{}c@{}}$ 0.0134 \pm 0.0047 $\\$ 0.0200 \pm 0.0130 $\end{tabular}} & {\renewcommand{\arraystretch}{1.25}\begin{tabular}{@{}c@{}}$ 0.5677 \pm 0.0139 $\\$ 0.5959 \pm 0.0207 $\end{tabular}} & {\renewcommand{\arraystretch}{1.25}\begin{tabular}{@{}c@{}}$ 0.9968 \pm 0.0006 $\\$ 0.9926 \pm 0.0016 $\end{tabular}} & {\renewcommand{\arraystretch}{1.25}\begin{tabular}{@{}c@{}}$ 0.3103 \pm 0.0595 $\\$ 0.4306 \pm 0.1291 $\end{tabular}} \\ \hline
{Real train--held-out} & {$ 0.1055 \pm 0.0071 $} & {$ 0.2392 \pm 0.0046 $} & {$ 0.0092 \pm 0.0024 $} & {$ 1.0330 \pm 0.0598 $} & {$ 0.9960 \pm 0.0009 $} & {$ 0.0818 \pm 0.0280 $} \\ \hline
\end{tabular}
}
\vspace{5pt}
\textbf{(b) CNV G-score fidelity}\\[2pt]
{\renewcommand{\arraystretch}{1.45}
\begin{tabular}{|>{\raggedright\arraybackslash}p{1.55cm}|*{6}{>{\centering\arraybackslash}p{\dimexpr(\textwidth-1.55cm-14\tabcolsep-8\arrayrulewidth)/6\relax}|}}
\hline
\rowcolor{black!12}
\textbf{Method} & \textbf{Mean amplification error $\downarrow$} & \textbf{Mean deletion error $\downarrow$} & \textbf{Amplification Pearson $\uparrow$} & \textbf{Deletion Pearson $\uparrow$} & \textbf{Top-50 amplification overlap $\uparrow$} & \textbf{Top-50 deletion overlap $\uparrow$} \\ \hline
TMO-Net & {\renewcommand{\arraystretch}{1.25}\begin{tabular}{@{}c@{}}$ \textbf{0.0290} \pm \textbf{0.0014} $\\$ \textbf{0.0368} \pm \textbf{0.0047} $\end{tabular}} & {\renewcommand{\arraystretch}{1.25}\begin{tabular}{@{}c@{}}$ 0.0713 \pm 0.0147 $\\$ 0.0856 \pm 0.0192 $\end{tabular}} & {\renewcommand{\arraystretch}{1.25}\begin{tabular}{@{}c@{}}$ \textbf{0.9563} \pm \textbf{0.0112} $\\$ \textbf{0.9186} \pm \textbf{0.0243} $\end{tabular}} & {\renewcommand{\arraystretch}{1.25}\begin{tabular}{@{}c@{}}$ \textbf{0.9169} \pm \textbf{0.0278} $\\$ 0.8769 \pm 0.0292 $\end{tabular}} & {\renewcommand{\arraystretch}{1.25}\begin{tabular}{@{}c@{}}$ \textbf{0.7267} \pm \textbf{0.0200} $\\$ \textbf{0.4644} \pm \textbf{0.2306} $\end{tabular}} & {\renewcommand{\arraystretch}{1.25}\begin{tabular}{@{}c@{}}$ \textbf{0.3467} \pm \textbf{0.1288} $\\$ \textbf{0.2822} \pm \textbf{0.1612} $\end{tabular}} \\ \hline
ISOMERIC & {\renewcommand{\arraystretch}{1.25}\begin{tabular}{@{}c@{}}$ 0.0378 \pm 0.0075 $\\$ 0.0469 \pm 0.0128 $\end{tabular}} & {\renewcommand{\arraystretch}{1.25}\begin{tabular}{@{}c@{}}$ \textbf{0.0697} \pm \textbf{0.0094} $\\$ \textbf{0.0825} \pm \textbf{0.0099} $\end{tabular}} & {\renewcommand{\arraystretch}{1.25}\begin{tabular}{@{}c@{}}$ 0.9441 \pm 0.0115 $\\$ 0.9084 \pm 0.0033 $\end{tabular}} & {\renewcommand{\arraystretch}{1.25}\begin{tabular}{@{}c@{}}$ 0.9143 \pm 0.0236 $\\$ \textbf{0.8772} \pm \textbf{0.0275} $\end{tabular}} & {\renewcommand{\arraystretch}{1.25}\begin{tabular}{@{}c@{}}$ \textbf{0.7267} \pm \textbf{0.1157} $\\$ 0.4289 \pm 0.2586 $\end{tabular}} & {\renewcommand{\arraystretch}{1.25}\begin{tabular}{@{}c@{}}$ 0.3444 \pm 0.1003 $\\$ 0.2444 \pm 0.1452 $\end{tabular}} \\ \hline
CustOmics & {\renewcommand{\arraystretch}{1.25}\begin{tabular}{@{}c@{}}$ 0.0341 \pm 0.0022 $\\$ 0.0395 \pm 0.0005 $\end{tabular}} & {\renewcommand{\arraystretch}{1.25}\begin{tabular}{@{}c@{}}$ 0.0984 \pm 0.0135 $\\$ 0.1175 \pm 0.0183 $\end{tabular}} & {\renewcommand{\arraystretch}{1.25}\begin{tabular}{@{}c@{}}$ 0.9516 \pm 0.0162 $\\$ 0.9158 \pm 0.0222 $\end{tabular}} & {\renewcommand{\arraystretch}{1.25}\begin{tabular}{@{}c@{}}$ 0.8519 \pm 0.0086 $\\$ 0.8151 \pm 0.0224 $\end{tabular}} & {\renewcommand{\arraystretch}{1.25}\begin{tabular}{@{}c@{}}$ 0.4356 \pm 0.0505 $\\$ 0.2556 \pm 0.1718 $\end{tabular}} & {\renewcommand{\arraystretch}{1.25}\begin{tabular}{@{}c@{}}$ 0.1778 \pm 0.0713 $\\$ 0.1689 \pm 0.0880 $\end{tabular}} \\ \hline
{Real train--held-out} & {$ 0.0289 \pm 0.0032 $} & {$ 0.0479 \pm 0.0032 $} & {$ 0.9517 \pm 0.0089 $} & {$ 0.9689 \pm 0.0068 $} & {$ 0.4933 \pm 0.3443 $} & {$ 0.3533 \pm 0.2157 $} \\ \hline
\end{tabular}
}
\end{table*}

\begin{table*}[!ht]
\centering
\caption{\textbf{Detailed marginal fidelity for the after-treatment cohort.} {Each model cell} reports comparison with real training data on the first line and held-out real test data on the second line. Values are the mean $\pm$ sample SD across three patient-disjoint outer folds after averaging three model initializations within each fold. {The final row in each panel compares real training with held-out data once per fold, its variance ratio is training/held-out.} Bold marks the best model mean separately for train and test, following the directions shown in the headers.}
\label{tab:after-fidelity-full}
\scriptsize
\setlength{\tabcolsep}{1pt}
\textbf{(a) RNA distribution fidelity}\\[2pt]
{\renewcommand{\arraystretch}{1.45}
\begin{tabular}{|>{\raggedright\arraybackslash}p{1.55cm}|*{6}{>{\centering\arraybackslash}p{\dimexpr(\textwidth-1.55cm-14\tabcolsep-8\arrayrulewidth)/6\relax}|}}
\hline
\rowcolor{black!12}
\textbf{Method} & \textbf{Mean gene-wise Wasserstein $\downarrow$} & \textbf{Mean JS distance $\downarrow$} & \textbf{RBF MMD$^2$ $\downarrow$} & \textbf{Median variance ratio $\rightarrow 1$} & \textbf{Gene-mean correlation $\uparrow$} & \textbf{Discriminative score $\downarrow$} \\ \hline
TMO-Net & {\renewcommand{\arraystretch}{1.25}\begin{tabular}{@{}c@{}}$ \textbf{0.1551} \pm \textbf{0.0077} $\\$ \textbf{0.2102} \pm \textbf{0.0258} $\end{tabular}} & {\renewcommand{\arraystretch}{1.25}\begin{tabular}{@{}c@{}}$ \textbf{0.3779} \pm \textbf{0.0088} $\\$ \textbf{0.4648} \pm \textbf{0.0148} $\end{tabular}} & {\renewcommand{\arraystretch}{1.25}\begin{tabular}{@{}c@{}}$ \textbf{0.0138} \pm \textbf{0.0039} $\\$ 0.0284 \pm 0.0125 $\end{tabular}} & {\renewcommand{\arraystretch}{1.25}\begin{tabular}{@{}c@{}}$ \textbf{0.6360} \pm \textbf{0.0087} $\\$ \textbf{0.7001} \pm \textbf{0.0371} $\end{tabular}} & {\renewcommand{\arraystretch}{1.25}\begin{tabular}{@{}c@{}}$ 0.9967 \pm 0.0016 $\\$ 0.9864 \pm 0.0059 $\end{tabular}} & {\renewcommand{\arraystretch}{1.25}\begin{tabular}{@{}c@{}}$ \textbf{0.1460} \pm \textbf{0.0235} $\\$ \textbf{0.2912} \pm \textbf{0.1571} $\end{tabular}} \\ \hline
ISOMERIC & {\renewcommand{\arraystretch}{1.25}\begin{tabular}{@{}c@{}}$ 0.1902 \pm 0.0108 $\\$ 0.2314 \pm 0.0062 $\end{tabular}} & {\renewcommand{\arraystretch}{1.25}\begin{tabular}{@{}c@{}}$ 0.3889 \pm 0.0108 $\\$ 0.4714 \pm 0.0118 $\end{tabular}} & {\renewcommand{\arraystretch}{1.25}\begin{tabular}{@{}c@{}}$ 0.0172 \pm 0.0036 $\\$ \textbf{0.0266} \pm \textbf{0.0015} $\end{tabular}} & {\renewcommand{\arraystretch}{1.25}\begin{tabular}{@{}c@{}}$ 0.5037 \pm 0.0379 $\\$ 0.5527 \pm 0.0317 $\end{tabular}} & {\renewcommand{\arraystretch}{1.25}\begin{tabular}{@{}c@{}}$ \textbf{0.9987} \pm \textbf{0.0000} $\\$ \textbf{0.9884} \pm \textbf{0.0009} $\end{tabular}} & {\renewcommand{\arraystretch}{1.25}\begin{tabular}{@{}c@{}}$ 0.2575 \pm 0.0414 $\\$ 0.3463 \pm 0.0649 $\end{tabular}} \\ \hline
CustOmics & {\renewcommand{\arraystretch}{1.25}\begin{tabular}{@{}c@{}}$ 0.2412 \pm 0.0285 $\\$ 0.2694 \pm 0.0411 $\end{tabular}} & {\renewcommand{\arraystretch}{1.25}\begin{tabular}{@{}c@{}}$ 0.3985 \pm 0.0141 $\\$ 0.4782 \pm 0.0185 $\end{tabular}} & {\renewcommand{\arraystretch}{1.25}\begin{tabular}{@{}c@{}}$ 0.0466 \pm 0.0151 $\\$ 0.0597 \pm 0.0257 $\end{tabular}} & {\renewcommand{\arraystretch}{1.25}\begin{tabular}{@{}c@{}}$ 0.4588 \pm 0.1095 $\\$ 0.5106 \pm 0.1393 $\end{tabular}} & {\renewcommand{\arraystretch}{1.25}\begin{tabular}{@{}c@{}}$ 0.9926 \pm 0.0012 $\\$ 0.9827 \pm 0.0040 $\end{tabular}} & {\renewcommand{\arraystretch}{1.25}\begin{tabular}{@{}c@{}}$ 0.3231 \pm 0.1317 $\\$ 0.6001 \pm 0.1544 $\end{tabular}} \\ \hline
{Real train--held-out} & {$ 0.1721 \pm 0.0051 $} & {$ 0.4323 \pm 0.0117 $} & {$ 0.0236 \pm 0.0031 $} & {$ 1.0920 \pm 0.0507 $} & {$ 0.9904 \pm 0.0014 $} & {$ 0.0068 \pm 0.0055 $} \\ \hline
\end{tabular}
}
\vspace{5pt}
\textbf{(b) CNV G-score fidelity}\\[2pt]
{\renewcommand{\arraystretch}{1.45}
\begin{tabular}{|>{\raggedright\arraybackslash}p{1.55cm}|*{6}{>{\centering\arraybackslash}p{\dimexpr(\textwidth-1.55cm-14\tabcolsep-8\arrayrulewidth)/6\relax}|}}
\hline
\rowcolor{black!12}
\textbf{Method} & \textbf{Mean amplification error $\downarrow$} & \textbf{Mean deletion error $\downarrow$} & \textbf{Amplification Pearson $\uparrow$} & \textbf{Deletion Pearson $\uparrow$} & \textbf{Top-50 amplification overlap $\uparrow$} & \textbf{Top-50 deletion overlap $\uparrow$} \\ \hline
TMO-Net & {\renewcommand{\arraystretch}{1.25}\begin{tabular}{@{}c@{}}$ 0.0373 \pm 0.0089 $\\$ \textbf{0.0407} \pm \textbf{0.0067} $\end{tabular}} & {\renewcommand{\arraystretch}{1.25}\begin{tabular}{@{}c@{}}$ 0.1017 \pm 0.0165 $\\$ 0.1354 \pm 0.0174 $\end{tabular}} & {\renewcommand{\arraystretch}{1.25}\begin{tabular}{@{}c@{}}$ 0.8996 \pm 0.0310 $\\$ 0.8368 \pm 0.0735 $\end{tabular}} & {\renewcommand{\arraystretch}{1.25}\begin{tabular}{@{}c@{}}$ 0.8929 \pm 0.0180 $\\$ 0.8110 \pm 0.0388 $\end{tabular}} & {\renewcommand{\arraystretch}{1.25}\begin{tabular}{@{}c@{}}$ 0.3911 \pm 0.1178 $\\$ \textbf{0.2489} \pm \textbf{0.1174} $\end{tabular}} & {\renewcommand{\arraystretch}{1.25}\begin{tabular}{@{}c@{}}$ 0.4244 \pm 0.3431 $\\$ 0.3489 \pm 0.3076 $\end{tabular}} \\ \hline
ISOMERIC & {\renewcommand{\arraystretch}{1.25}\begin{tabular}{@{}c@{}}$ \textbf{0.0335} \pm \textbf{0.0039} $\\$ 0.0470 \pm 0.0065 $\end{tabular}} & {\renewcommand{\arraystretch}{1.25}\begin{tabular}{@{}c@{}}$ \textbf{0.0916} \pm \textbf{0.0210} $\\$ \textbf{0.1319} \pm \textbf{0.0414} $\end{tabular}} & {\renewcommand{\arraystretch}{1.25}\begin{tabular}{@{}c@{}}$ \textbf{0.9211} \pm \textbf{0.0108} $\\$ \textbf{0.8374} \pm \textbf{0.0438} $\end{tabular}} & {\renewcommand{\arraystretch}{1.25}\begin{tabular}{@{}c@{}}$ \textbf{0.9079} \pm \textbf{0.0340} $\\$ \textbf{0.8206} \pm \textbf{0.0367} $\end{tabular}} & {\renewcommand{\arraystretch}{1.25}\begin{tabular}{@{}c@{}}$ 0.4756 \pm 0.1865 $\\$ 0.1556 \pm 0.1367 $\end{tabular}} & {\renewcommand{\arraystretch}{1.25}\begin{tabular}{@{}c@{}}$ \textbf{0.4689} \pm \textbf{0.3529} $\\$ \textbf{0.4222} \pm \textbf{0.2600} $\end{tabular}} \\ \hline
CustOmics & {\renewcommand{\arraystretch}{1.25}\begin{tabular}{@{}c@{}}$ 0.0458 \pm 0.0100 $\\$ 0.0437 \pm 0.0093 $\end{tabular}} & {\renewcommand{\arraystretch}{1.25}\begin{tabular}{@{}c@{}}$ 0.1938 \pm 0.0396 $\\$ 0.2375 \pm 0.0270 $\end{tabular}} & {\renewcommand{\arraystretch}{1.25}\begin{tabular}{@{}c@{}}$ 0.9154 \pm 0.0364 $\\$ 0.8263 \pm 0.1082 $\end{tabular}} & {\renewcommand{\arraystretch}{1.25}\begin{tabular}{@{}c@{}}$ 0.6141 \pm 0.0529 $\\$ 0.5442 \pm 0.0316 $\end{tabular}} & {\renewcommand{\arraystretch}{1.25}\begin{tabular}{@{}c@{}}$ \textbf{0.5333} \pm \textbf{0.1368} $\\$ 0.2156 \pm 0.0391 $\end{tabular}} & {\renewcommand{\arraystretch}{1.25}\begin{tabular}{@{}c@{}}$ 0.0889 \pm 0.1195 $\\$ 0.1778 \pm 0.1413 $\end{tabular}} \\ \hline
{Real train--held-out} & {$ 0.0355 \pm 0.0106 $} & {$ 0.0819 \pm 0.0200 $} & {$ 0.9145 \pm 0.0492 $} & {$ 0.9396 \pm 0.0375 $} & {$ 0.2000 \pm 0.1510 $} & {$ 0.4200 \pm 0.2905 $} \\ \hline
\end{tabular}
}
\end{table*}

\begin{table*}[p]

\centering
\caption{\textbf{CNV correlations and biomarker-level absolute errors.} Pearson correlations use all 3,000 evaluated CNV genes per fold, mean absolute errors (MAEs) use the fixed panel of 14 prespecified biomarkers. All comparisons use held-out real data. Values are means $\pm$ sample SD across three patient-disjoint outer folds after averaging per-initialization metrics within fold. The real-data reference compares real training with held-out scores once per fold. MAEs are in G-score units, lower values indicate closer agreement.}
\label{tab:biomarker-fidelity}
\small
\setlength{\tabcolsep}{5pt}
\renewcommand{\arraystretch}{1.2}
\textbf{(a) Before treatment}\\[3pt]
\begin{tabular*}{\textwidth}{@{\extracolsep{\fill}}lrrrr@{}}
\toprule
 & \multicolumn{2}{c}{Pearson correlation $\uparrow$} & \multicolumn{2}{c}{Biomarker MAE $\downarrow$} \\
\cmidrule(lr){2-3}\cmidrule(lr){4-5}
Method & Amplification & Deletion & Amplification & Deletion \\
\midrule
TMO-Net & $0.919 \pm 0.024$ & $0.877 \pm 0.029$ & $0.070 \pm 0.012$ & $0.091 \pm 0.003$ \\
ISOMERIC & $0.908 \pm 0.003$ & $0.877 \pm 0.028$ & $0.066 \pm 0.011$ & $0.090 \pm 0.028$ \\
CustOmics & $0.916 \pm 0.022$ & $0.815 \pm 0.022$ & $0.066 \pm 0.015$ & $0.126 \pm 0.026$ \\
\midrule
Real train--held-out & $0.952 \pm 0.009$ & $0.969 \pm 0.007$ & $0.041 \pm 0.012$ & $0.033 \pm 0.002$ \\
\bottomrule
\end{tabular*}
\par\vspace{8pt}
\textbf{(b) After treatment}\\[3pt]
\begin{tabular*}{\textwidth}{@{\extracolsep{\fill}}lrrrr@{}}
\toprule
 & \multicolumn{2}{c}{Pearson correlation $\uparrow$} & \multicolumn{2}{c}{Biomarker MAE $\downarrow$} \\
\cmidrule(lr){2-3}\cmidrule(lr){4-5}
Method & Amplification & Deletion & Amplification & Deletion \\
\midrule
TMO-Net & $0.837 \pm 0.074$ & $0.811 \pm 0.039$ & $0.067 \pm 0.024$ & $0.081 \pm 0.016$ \\
ISOMERIC & $0.837 \pm 0.044$ & $0.821 \pm 0.037$ & $0.087 \pm 0.002$ & $0.078 \pm 0.018$ \\
CustOmics & $0.826 \pm 0.108$ & $0.544 \pm 0.032$ & $0.096 \pm 0.008$ & $0.137 \pm 0.022$ \\
\midrule
Real train--held-out & $0.914 \pm 0.049$ & $0.940 \pm 0.038$ & $0.068 \pm 0.018$ & $0.035 \pm 0.008$ \\
\bottomrule
\end{tabular*}
\end{table*}

\begin{table*}[p]

\centering
\caption{\textbf{Biomarker-level amplification G-scores for the before-treatment primary cohort.} All 14 prespecified biomarkers are reported. Values are means $\pm$ sample SD across three patient-disjoint outer folds. Within each fold, synthetic scores, signed biases, and absolute errors are averaged over three initializations, absolute errors are computed before averaging. Bias is synthetic minus held-out real G-score, positive values indicate overestimation. G-scores summarize event frequency and amplitude, so errors are in G-score units, not frequency percentage points. Lower absolute error indicates closer agreement.}
\label{tab:before-biomarker-amp}
\scriptsize
\setlength{\tabcolsep}{5pt}
\renewcommand{\arraystretch}{1.05}
\begin{tabular*}{\textwidth}{@{\extracolsep{\fill}}lllr rr@{}}
\toprule
\textbf{Gene} & \textbf{Real held-out} & \textbf{Model} & \textbf{Synthetic} & \textbf{Bias} & \textbf{Absolute error} \\
\midrule
\textit{FOLR1} & $0.108 \pm 0.041$ & TMO-Net & $0.051 \pm 0.030$ & $-0.057 \pm 0.059$ & $0.059 \pm 0.056$ \\
 &  & ISOMERIC & $0.003 \pm 0.002$ & $-0.105 \pm 0.040$ & $0.105 \pm 0.040$ \\
 &  & CustOmics & $0.015 \pm 0.010$ & $-0.092 \pm 0.045$ & $0.092 \pm 0.045$ \\
\addlinespace[1pt]
\textit{ERBB2} & $0.068 \pm 0.019$ & TMO-Net & $0.030 \pm 0.018$ & $-0.037 \pm 0.035$ & $0.039 \pm 0.033$ \\
 &  & ISOMERIC & $0.001 \pm 0.000$ & $-0.067 \pm 0.019$ & $0.067 \pm 0.019$ \\
 &  & CustOmics & $0.001 \pm 0.001$ & $-0.067 \pm 0.018$ & $0.067 \pm 0.018$ \\
\addlinespace[1pt]
\textit{PTEN} & $0.004 \pm 0.004$ & TMO-Net & $0.112 \pm 0.017$ & $0.108 \pm 0.019$ & $0.108 \pm 0.019$ \\
 &  & ISOMERIC & $0.022 \pm 0.011$ & $0.018 \pm 0.010$ & $0.018 \pm 0.010$ \\
 &  & CustOmics & $0.020 \pm 0.007$ & $0.016 \pm 0.003$ & $0.016 \pm 0.003$ \\
\addlinespace[1pt]
\textit{KRAS} & $0.263 \pm 0.095$ & TMO-Net & $0.284 \pm 0.116$ & $0.021 \pm 0.210$ & $0.168 \pm 0.050$ \\
 &  & ISOMERIC & $0.179 \pm 0.124$ & $-0.084 \pm 0.213$ & $0.136 \pm 0.168$ \\
 &  & CustOmics & $0.196 \pm 0.097$ & $-0.067 \pm 0.192$ & $0.154 \pm 0.110$ \\
\addlinespace[1pt]
\textit{NF1} & $0.001 \pm 0.002$ & TMO-Net & $0.075 \pm 0.030$ & $0.074 \pm 0.032$ & $0.074 \pm 0.032$ \\
 &  & ISOMERIC & $0.026 \pm 0.021$ & $0.025 \pm 0.023$ & $0.025 \pm 0.022$ \\
 &  & CustOmics & $0.009 \pm 0.005$ & $0.008 \pm 0.007$ & $0.010 \pm 0.004$ \\
\addlinespace[1pt]
\textit{AKT2} & $0.303 \pm 0.059$ & TMO-Net & $0.332 \pm 0.011$ & $0.029 \pm 0.069$ & $0.063 \pm 0.011$ \\
 &  & ISOMERIC & $0.348 \pm 0.044$ & $0.045 \pm 0.100$ & $0.092 \pm 0.022$ \\
 &  & CustOmics & $0.323 \pm 0.054$ & $0.020 \pm 0.105$ & $0.089 \pm 0.028$ \\
\addlinespace[1pt]
\textit{PTK2} & $0.467 \pm 0.043$ & TMO-Net & $0.393 \pm 0.006$ & $-0.074 \pm 0.046$ & $0.079 \pm 0.037$ \\
 &  & ISOMERIC & $0.480 \pm 0.040$ & $0.013 \pm 0.040$ & $0.038 \pm 0.004$ \\
 &  & CustOmics & $0.421 \pm 0.027$ & $-0.046 \pm 0.066$ & $0.101 \pm 0.014$ \\
\addlinespace[1pt]
\textit{CCNE1} & $0.637 \pm 0.109$ & TMO-Net & $0.762 \pm 0.091$ & $0.126 \pm 0.199$ & $0.157 \pm 0.165$ \\
 &  & ISOMERIC & $0.753 \pm 0.074$ & $0.117 \pm 0.157$ & $0.156 \pm 0.094$ \\
 &  & CustOmics & $0.700 \pm 0.088$ & $0.064 \pm 0.197$ & $0.140 \pm 0.124$ \\
\addlinespace[1pt]
\textit{MTOR} & $0.062 \pm 0.014$ & TMO-Net & $0.037 \pm 0.023$ & $-0.026 \pm 0.032$ & $0.030 \pm 0.028$ \\
 &  & ISOMERIC & $0.001 \pm 0.003$ & $-0.061 \pm 0.013$ & $0.061 \pm 0.013$ \\
 &  & CustOmics & $0.003 \pm 0.001$ & $-0.060 \pm 0.013$ & $0.060 \pm 0.013$ \\
\addlinespace[1pt]
\textit{PIK3CA} & $0.454 \pm 0.064$ & TMO-Net & $0.452 \pm 0.068$ & $-0.002 \pm 0.130$ & $0.099 \pm 0.049$ \\
 &  & ISOMERIC & $0.299 \pm 0.096$ & $-0.155 \pm 0.157$ & $0.155 \pm 0.157$ \\
 &  & CustOmics & $0.330 \pm 0.062$ & $-0.124 \pm 0.054$ & $0.124 \pm 0.054$ \\
\addlinespace[1pt]
\textit{BRCA1} & $0.009 \pm 0.012$ & TMO-Net & $0.010 \pm 0.011$ & $0.001 \pm 0.018$ & $0.013 \pm 0.008$ \\
 &  & ISOMERIC & $0.000 \pm 0.000$ & $-0.009 \pm 0.012$ & $0.009 \pm 0.012$ \\
 &  & CustOmics & $0.004 \pm 0.005$ & $-0.005 \pm 0.015$ & $0.011 \pm 0.009$ \\
\addlinespace[1pt]
\textit{BRCA2} & $0.025 \pm 0.029$ & TMO-Net & $0.006 \pm 0.004$ & $-0.019 \pm 0.033$ & $0.026 \pm 0.025$ \\
 &  & ISOMERIC & $0.000 \pm 0.000$ & $-0.025 \pm 0.029$ & $0.025 \pm 0.029$ \\
 &  & CustOmics & $0.000 \pm 0.000$ & $-0.025 \pm 0.029$ & $0.025 \pm 0.029$ \\
\addlinespace[1pt]
\textit{RAD51C} & $0.039 \pm 0.015$ & TMO-Net & $0.086 \pm 0.047$ & $0.047 \pm 0.060$ & $0.061 \pm 0.037$ \\
 &  & ISOMERIC & $0.002 \pm 0.002$ & $-0.038 \pm 0.018$ & $0.038 \pm 0.018$ \\
 &  & CustOmics & $0.011 \pm 0.011$ & $-0.028 \pm 0.020$ & $0.029 \pm 0.019$ \\
\addlinespace[1pt]
\textit{RAD51D} & $0.006 \pm 0.010$ & TMO-Net & $0.001 \pm 0.001$ & $-0.005 \pm 0.009$ & $0.005 \pm 0.009$ \\
 &  & ISOMERIC & $0.000 \pm 0.000$ & $-0.006 \pm 0.010$ & $0.006 \pm 0.010$ \\
 &  & CustOmics & $0.000 \pm 0.000$ & $-0.005 \pm 0.010$ & $0.006 \pm 0.010$ \\
\bottomrule
\end{tabular*}
\end{table*}

\begin{table*}[p]

\centering
\caption{\textbf{Biomarker-level deletion G-scores for the before-treatment primary cohort.} All 14 prespecified biomarkers are reported. Values are means $\pm$ sample SD across three patient-disjoint outer folds. Within each fold, synthetic scores, signed biases, and absolute errors are averaged over three initializations, absolute errors are computed before averaging. Bias is synthetic minus held-out real G-score, positive values indicate overestimation. G-scores summarize event frequency and amplitude, so errors are in G-score units, not frequency percentage points. Lower absolute error indicates closer agreement.}
\label{tab:before-biomarker-del}
\scriptsize
\setlength{\tabcolsep}{5pt}
\renewcommand{\arraystretch}{1.05}
\begin{tabular*}{\textwidth}{@{\extracolsep{\fill}}lllr rr@{}}
\toprule
\textbf{Gene} & \textbf{Real held-out} & \textbf{Model} & \textbf{Synthetic} & \textbf{Bias} & \textbf{Absolute error} \\
\midrule
\textit{FOLR1} & $0.009 \pm 0.015$ & TMO-Net & $0.001 \pm 0.001$ & $-0.008 \pm 0.015$ & $0.009 \pm 0.014$ \\
 &  & ISOMERIC & $0.000 \pm 0.000$ & $-0.009 \pm 0.015$ & $0.009 \pm 0.015$ \\
 &  & CustOmics & $0.000 \pm 0.000$ & $-0.009 \pm 0.015$ & $0.009 \pm 0.015$ \\
\addlinespace[1pt]
\textit{ERBB2} & $0.205 \pm 0.033$ & TMO-Net & $0.192 \pm 0.065$ & $-0.013 \pm 0.099$ & $0.070 \pm 0.051$ \\
 &  & ISOMERIC & $0.042 \pm 0.033$ & $-0.164 \pm 0.059$ & $0.164 \pm 0.059$ \\
 &  & CustOmics & $0.105 \pm 0.036$ & $-0.100 \pm 0.046$ & $0.102 \pm 0.043$ \\
\addlinespace[1pt]
\textit{PTEN} & $0.150 \pm 0.010$ & TMO-Net & $0.399 \pm 0.065$ & $0.249 \pm 0.075$ & $0.249 \pm 0.075$ \\
 &  & ISOMERIC & $0.179 \pm 0.021$ & $0.029 \pm 0.023$ & $0.040 \pm 0.011$ \\
 &  & CustOmics & $0.350 \pm 0.089$ & $0.200 \pm 0.097$ & $0.207 \pm 0.085$ \\
\addlinespace[1pt]
\textit{KRAS} & $0.015 \pm 0.007$ & TMO-Net & $0.009 \pm 0.009$ & $-0.006 \pm 0.004$ & $0.006 \pm 0.004$ \\
 &  & ISOMERIC & $0.005 \pm 0.006$ & $-0.010 \pm 0.010$ & $0.014 \pm 0.005$ \\
 &  & CustOmics & $0.001 \pm 0.002$ & $-0.014 \pm 0.008$ & $0.014 \pm 0.008$ \\
\addlinespace[1pt]
\textit{NF1} & $0.380 \pm 0.052$ & TMO-Net & $0.751 \pm 0.074$ & $0.371 \pm 0.125$ & $0.371 \pm 0.125$ \\
 &  & ISOMERIC & $0.719 \pm 0.103$ & $0.338 \pm 0.134$ & $0.338 \pm 0.134$ \\
 &  & CustOmics & $1.049 \pm 0.127$ & $0.668 \pm 0.173$ & $0.668 \pm 0.173$ \\
\addlinespace[1pt]
\textit{AKT2} & $0.073 \pm 0.027$ & TMO-Net & $0.029 \pm 0.010$ & $-0.044 \pm 0.037$ & $0.047 \pm 0.031$ \\
 &  & ISOMERIC & $0.093 \pm 0.043$ & $0.019 \pm 0.069$ & $0.046 \pm 0.046$ \\
 &  & CustOmics & $0.044 \pm 0.025$ & $-0.029 \pm 0.036$ & $0.038 \pm 0.025$ \\
\addlinespace[1pt]
\textit{PTK2} & $0.039 \pm 0.015$ & TMO-Net & $0.006 \pm 0.001$ & $-0.033 \pm 0.016$ & $0.033 \pm 0.016$ \\
 &  & ISOMERIC & $0.038 \pm 0.034$ & $-0.001 \pm 0.047$ & $0.036 \pm 0.014$ \\
 &  & CustOmics & $0.015 \pm 0.003$ & $-0.024 \pm 0.012$ & $0.024 \pm 0.012$ \\
\addlinespace[1pt]
\textit{CCNE1} & $0.012 \pm 0.009$ & TMO-Net & $0.013 \pm 0.008$ & $0.001 \pm 0.015$ & $0.011 \pm 0.008$ \\
 &  & ISOMERIC & $0.024 \pm 0.022$ & $0.012 \pm 0.031$ & $0.025 \pm 0.018$ \\
 &  & CustOmics & $0.012 \pm 0.010$ & $0.000 \pm 0.019$ & $0.015 \pm 0.008$ \\
\addlinespace[1pt]
\textit{MTOR} & $0.062 \pm 0.021$ & TMO-Net & $0.016 \pm 0.004$ & $-0.045 \pm 0.023$ & $0.045 \pm 0.023$ \\
 &  & ISOMERIC & $0.000 \pm 0.000$ & $-0.061 \pm 0.020$ & $0.061 \pm 0.020$ \\
 &  & CustOmics & $0.002 \pm 0.003$ & $-0.060 \pm 0.019$ & $0.060 \pm 0.019$ \\
\addlinespace[1pt]
\textit{PIK3CA} & $0.003 \pm 0.006$ & TMO-Net & $0.000 \pm 0.000$ & $-0.003 \pm 0.006$ & $0.003 \pm 0.006$ \\
 &  & ISOMERIC & $0.000 \pm 0.000$ & $-0.003 \pm 0.006$ & $0.003 \pm 0.006$ \\
 &  & CustOmics & $0.000 \pm 0.000$ & $-0.003 \pm 0.006$ & $0.003 \pm 0.006$ \\
\addlinespace[1pt]
\textit{BRCA1} & $0.227 \pm 0.062$ & TMO-Net & $0.241 \pm 0.101$ & $0.014 \pm 0.163$ & $0.128 \pm 0.048$ \\
 &  & ISOMERIC & $0.119 \pm 0.076$ & $-0.107 \pm 0.138$ & $0.123 \pm 0.123$ \\
 &  & CustOmics & $0.415 \pm 0.115$ & $0.188 \pm 0.133$ & $0.199 \pm 0.121$ \\
\addlinespace[1pt]
\textit{BRCA2} & $0.205 \pm 0.058$ & TMO-Net & $0.143 \pm 0.013$ & $-0.062 \pm 0.064$ & $0.069 \pm 0.057$ \\
 &  & ISOMERIC & $0.051 \pm 0.022$ & $-0.154 \pm 0.074$ & $0.154 \pm 0.074$ \\
 &  & CustOmics & $0.074 \pm 0.020$ & $-0.131 \pm 0.078$ & $0.131 \pm 0.078$ \\
\addlinespace[1pt]
\textit{RAD51C} & $0.120 \pm 0.068$ & TMO-Net & $0.161 \pm 0.124$ & $0.041 \pm 0.192$ & $0.140 \pm 0.099$ \\
 &  & ISOMERIC & $0.023 \pm 0.029$ & $-0.097 \pm 0.096$ & $0.106 \pm 0.082$ \\
 &  & CustOmics & $0.191 \pm 0.163$ & $0.071 \pm 0.222$ & $0.194 \pm 0.009$ \\
\addlinespace[1pt]
\textit{RAD51D} & $0.276 \pm 0.042$ & TMO-Net & $0.206 \pm 0.069$ & $-0.070 \pm 0.111$ & $0.084 \pm 0.097$ \\
 &  & ISOMERIC & $0.135 \pm 0.063$ & $-0.141 \pm 0.094$ & $0.141 \pm 0.094$ \\
 &  & CustOmics & $0.198 \pm 0.043$ & $-0.078 \pm 0.063$ & $0.101 \pm 0.025$ \\
\bottomrule
\end{tabular*}
\end{table*}

\begin{table*}[p]

\centering
\caption{\textbf{Biomarker-level amplification G-scores for the after-treatment cohort.} All 14 prespecified biomarkers are reported. Values are means $\pm$ sample SD across three patient-disjoint outer folds. Within each fold, synthetic scores, signed biases, and absolute errors are averaged over three initializations, absolute errors are computed before averaging. Bias is synthetic minus held-out real G-score, positive values indicate overestimation. G-scores summarize event frequency and amplitude, so errors are in G-score units, not frequency percentage points. Lower absolute error indicates closer agreement.}
\label{tab:after-biomarker-amp}
\scriptsize
\setlength{\tabcolsep}{5pt}
\renewcommand{\arraystretch}{1.05}
\begin{tabular*}{\textwidth}{@{\extracolsep{\fill}}lllr rr@{}}
\toprule
\textbf{Gene} & \textbf{Real held-out} & \textbf{Model} & \textbf{Synthetic} & \textbf{Bias} & \textbf{Absolute error} \\
\midrule
\textit{FOLR1} & $0.115 \pm 0.062$ & TMO-Net & $0.066 \pm 0.023$ & $-0.049 \pm 0.084$ & $0.069 \pm 0.062$ \\
 &  & ISOMERIC & $0.019 \pm 0.024$ & $-0.096 \pm 0.083$ & $0.096 \pm 0.083$ \\
 &  & CustOmics & $0.023 \pm 0.029$ & $-0.092 \pm 0.086$ & $0.095 \pm 0.081$ \\
\addlinespace[1pt]
\textit{ERBB2} & $0.042 \pm 0.044$ & TMO-Net & $0.004 \pm 0.004$ & $-0.038 \pm 0.048$ & $0.044 \pm 0.040$ \\
 &  & ISOMERIC & $0.000 \pm 0.000$ & $-0.042 \pm 0.044$ & $0.042 \pm 0.044$ \\
 &  & CustOmics & $0.000 \pm 0.000$ & $-0.042 \pm 0.044$ & $0.042 \pm 0.044$ \\
\addlinespace[1pt]
\textit{PTEN} & $0.000 \pm 0.000$ & TMO-Net & $0.027 \pm 0.023$ & $0.027 \pm 0.023$ & $0.027 \pm 0.023$ \\
 &  & ISOMERIC & $0.007 \pm 0.009$ & $0.007 \pm 0.009$ & $0.007 \pm 0.009$ \\
 &  & CustOmics & $0.010 \pm 0.014$ & $0.010 \pm 0.014$ & $0.010 \pm 0.014$ \\
\addlinespace[1pt]
\textit{KRAS} & $0.251 \pm 0.047$ & TMO-Net & $0.253 \pm 0.014$ & $0.002 \pm 0.033$ & $0.047 \pm 0.019$ \\
 &  & ISOMERIC & $0.346 \pm 0.167$ & $0.095 \pm 0.198$ & $0.155 \pm 0.129$ \\
 &  & CustOmics & $0.195 \pm 0.057$ & $-0.056 \pm 0.048$ & $0.056 \pm 0.048$ \\
\addlinespace[1pt]
\textit{NF1} & $0.000 \pm 0.000$ & TMO-Net & $0.057 \pm 0.056$ & $0.057 \pm 0.056$ & $0.057 \pm 0.056$ \\
 &  & ISOMERIC & $0.026 \pm 0.020$ & $0.026 \pm 0.020$ & $0.026 \pm 0.020$ \\
 &  & CustOmics & $0.004 \pm 0.002$ & $0.004 \pm 0.002$ & $0.004 \pm 0.002$ \\
\addlinespace[1pt]
\textit{AKT2} & $0.266 \pm 0.109$ & TMO-Net & $0.243 \pm 0.043$ & $-0.023 \pm 0.143$ & $0.109 \pm 0.059$ \\
 &  & ISOMERIC & $0.271 \pm 0.074$ & $0.005 \pm 0.171$ & $0.130 \pm 0.064$ \\
 &  & CustOmics & $0.117 \pm 0.013$ & $-0.150 \pm 0.099$ & $0.150 \pm 0.099$ \\
\addlinespace[1pt]
\textit{PTK2} & $0.496 \pm 0.061$ & TMO-Net & $0.430 \pm 0.021$ & $-0.067 \pm 0.063$ & $0.074 \pm 0.060$ \\
 &  & ISOMERIC & $0.471 \pm 0.077$ & $-0.025 \pm 0.138$ & $0.114 \pm 0.023$ \\
 &  & CustOmics & $0.305 \pm 0.084$ & $-0.191 \pm 0.120$ & $0.191 \pm 0.120$ \\
\addlinespace[1pt]
\textit{CCNE1} & $0.592 \pm 0.276$ & TMO-Net & $0.666 \pm 0.146$ & $0.074 \pm 0.422$ & $0.298 \pm 0.230$ \\
 &  & ISOMERIC & $0.584 \pm 0.171$ & $-0.008 \pm 0.447$ & $0.341 \pm 0.158$ \\
 &  & CustOmics & $0.402 \pm 0.254$ & $-0.190 \pm 0.527$ & $0.469 \pm 0.049$ \\
\addlinespace[1pt]
\textit{MTOR} & $0.006 \pm 0.011$ & TMO-Net & $0.005 \pm 0.005$ & $-0.001 \pm 0.009$ & $0.006 \pm 0.006$ \\
 &  & ISOMERIC & $0.000 \pm 0.000$ & $-0.006 \pm 0.011$ & $0.006 \pm 0.011$ \\
 &  & CustOmics & $0.000 \pm 0.000$ & $-0.006 \pm 0.011$ & $0.006 \pm 0.011$ \\
\addlinespace[1pt]
\textit{PIK3CA} & $0.512 \pm 0.184$ & TMO-Net & $0.451 \pm 0.064$ & $-0.061 \pm 0.245$ & $0.173 \pm 0.145$ \\
 &  & ISOMERIC & $0.379 \pm 0.132$ & $-0.132 \pm 0.314$ & $0.261 \pm 0.151$ \\
 &  & CustOmics & $0.274 \pm 0.107$ & $-0.238 \pm 0.291$ & $0.275 \pm 0.240$ \\
\addlinespace[1pt]
\textit{BRCA1} & $0.005 \pm 0.009$ & TMO-Net & $0.000 \pm 0.000$ & $-0.005 \pm 0.009$ & $0.005 \pm 0.009$ \\
 &  & ISOMERIC & $0.001 \pm 0.002$ & $-0.005 \pm 0.010$ & $0.006 \pm 0.009$ \\
 &  & CustOmics & $0.000 \pm 0.000$ & $-0.005 \pm 0.009$ & $0.005 \pm 0.009$ \\
\addlinespace[1pt]
\textit{BRCA2} & $0.019 \pm 0.032$ & TMO-Net & $0.000 \pm 0.000$ & $-0.019 \pm 0.032$ & $0.019 \pm 0.032$ \\
 &  & ISOMERIC & $0.000 \pm 0.000$ & $-0.019 \pm 0.032$ & $0.019 \pm 0.032$ \\
 &  & CustOmics & $0.000 \pm 0.000$ & $-0.019 \pm 0.032$ & $0.019 \pm 0.032$ \\
\addlinespace[1pt]
\textit{RAD51C} & $0.021 \pm 0.005$ & TMO-Net & $0.011 \pm 0.009$ & $-0.010 \pm 0.013$ & $0.016 \pm 0.007$ \\
 &  & ISOMERIC & $0.002 \pm 0.002$ & $-0.019 \pm 0.005$ & $0.019 \pm 0.005$ \\
 &  & CustOmics & $0.000 \pm 0.000$ & $-0.021 \pm 0.005$ & $0.021 \pm 0.005$ \\
\addlinespace[1pt]
\textit{RAD51D} & $0.000 \pm 0.000$ & TMO-Net & $0.000 \pm 0.000$ & $0.000 \pm 0.000$ & $0.000 \pm 0.000$ \\
 &  & ISOMERIC & $0.000 \pm 0.000$ & $0.000 \pm 0.000$ & $0.000 \pm 0.000$ \\
 &  & CustOmics & $0.000 \pm 0.000$ & $0.000 \pm 0.000$ & $0.000 \pm 0.000$ \\
\bottomrule
\end{tabular*}
\end{table*}

\begin{table*}[p]

\centering
\caption{\textbf{Biomarker-level deletion G-scores for the after-treatment cohort.} All 14 prespecified biomarkers are reported. Values are means $\pm$ sample SD across three patient-disjoint outer folds. Within each fold, synthetic scores, signed biases, and absolute errors are averaged over three initializations, absolute errors are computed before averaging. Bias is synthetic minus held-out real G-score, positive values indicate overestimation. G-scores summarize event frequency and amplitude, so errors are in G-score units, not frequency percentage points. Lower absolute error indicates closer agreement.}
\label{tab:after-biomarker-del}
\scriptsize
\setlength{\tabcolsep}{5pt}
\renewcommand{\arraystretch}{1.05}
\begin{tabular*}{\textwidth}{@{\extracolsep{\fill}}lllr rr@{}}
\toprule
\textbf{Gene} & \textbf{Real held-out} & \textbf{Model} & \textbf{Synthetic} & \textbf{Bias} & \textbf{Absolute error} \\
\midrule
\textit{FOLR1} & $0.006 \pm 0.010$ & TMO-Net & $0.004 \pm 0.004$ & $-0.002 \pm 0.013$ & $0.009 \pm 0.007$ \\
 &  & ISOMERIC & $0.001 \pm 0.002$ & $-0.005 \pm 0.011$ & $0.006 \pm 0.009$ \\
 &  & CustOmics & $0.001 \pm 0.002$ & $-0.004 \pm 0.008$ & $0.004 \pm 0.008$ \\
\addlinespace[1pt]
\textit{ERBB2} & $0.153 \pm 0.009$ & TMO-Net & $0.088 \pm 0.018$ & $-0.065 \pm 0.027$ & $0.065 \pm 0.027$ \\
 &  & ISOMERIC & $0.026 \pm 0.020$ & $-0.127 \pm 0.027$ & $0.127 \pm 0.027$ \\
 &  & CustOmics & $0.059 \pm 0.034$ & $-0.094 \pm 0.041$ & $0.094 \pm 0.041$ \\
\addlinespace[1pt]
\textit{PTEN} & $0.092 \pm 0.036$ & TMO-Net & $0.161 \pm 0.133$ & $0.069 \pm 0.156$ & $0.143 \pm 0.030$ \\
 &  & ISOMERIC & $0.031 \pm 0.030$ & $-0.061 \pm 0.052$ & $0.061 \pm 0.052$ \\
 &  & CustOmics & $0.362 \pm 0.401$ & $0.270 \pm 0.436$ & $0.350 \pm 0.341$ \\
\addlinespace[1pt]
\textit{KRAS} & $0.018 \pm 0.016$ & TMO-Net & $0.018 \pm 0.018$ & $0.000 \pm 0.018$ & $0.016 \pm 0.012$ \\
 &  & ISOMERIC & $0.034 \pm 0.045$ & $0.017 \pm 0.042$ & $0.027 \pm 0.032$ \\
 &  & CustOmics & $0.010 \pm 0.004$ & $-0.008 \pm 0.015$ & $0.017 \pm 0.009$ \\
\addlinespace[1pt]
\textit{NF1} & $0.309 \pm 0.137$ & TMO-Net & $0.810 \pm 0.198$ & $0.501 \pm 0.333$ & $0.501 \pm 0.333$ \\
 &  & ISOMERIC & $0.694 \pm 0.241$ & $0.385 \pm 0.377$ & $0.385 \pm 0.377$ \\
 &  & CustOmics & $1.347 \pm 0.149$ & $1.038 \pm 0.286$ & $1.038 \pm 0.286$ \\
\addlinespace[1pt]
\textit{AKT2} & $0.102 \pm 0.007$ & TMO-Net & $0.037 \pm 0.012$ & $-0.065 \pm 0.011$ & $0.065 \pm 0.011$ \\
 &  & ISOMERIC & $0.045 \pm 0.022$ & $-0.057 \pm 0.028$ & $0.057 \pm 0.028$ \\
 &  & CustOmics & $0.039 \pm 0.030$ & $-0.064 \pm 0.035$ & $0.069 \pm 0.027$ \\
\addlinespace[1pt]
\textit{PTK2} & $0.005 \pm 0.009$ & TMO-Net & $0.000 \pm 0.000$ & $-0.005 \pm 0.009$ & $0.005 \pm 0.009$ \\
 &  & ISOMERIC & $0.005 \pm 0.004$ & $0.000 \pm 0.013$ & $0.010 \pm 0.005$ \\
 &  & CustOmics & $0.005 \pm 0.006$ & $0.000 \pm 0.012$ & $0.008 \pm 0.007$ \\
\addlinespace[1pt]
\textit{CCNE1} & $0.006 \pm 0.010$ & TMO-Net & $0.004 \pm 0.004$ & $-0.002 \pm 0.013$ & $0.010 \pm 0.007$ \\
 &  & ISOMERIC & $0.004 \pm 0.004$ & $-0.002 \pm 0.013$ & $0.009 \pm 0.007$ \\
 &  & CustOmics & $0.002 \pm 0.004$ & $-0.003 \pm 0.012$ & $0.008 \pm 0.009$ \\
\addlinespace[1pt]
\textit{MTOR} & $0.044 \pm 0.033$ & TMO-Net & $0.002 \pm 0.002$ & $-0.042 \pm 0.035$ & $0.042 \pm 0.035$ \\
 &  & ISOMERIC & $0.000 \pm 0.000$ & $-0.044 \pm 0.033$ & $0.044 \pm 0.033$ \\
 &  & CustOmics & $0.005 \pm 0.008$ & $-0.040 \pm 0.036$ & $0.040 \pm 0.036$ \\
\addlinespace[1pt]
\textit{PIK3CA} & $0.000 \pm 0.000$ & TMO-Net & $0.000 \pm 0.000$ & $0.000 \pm 0.000$ & $0.000 \pm 0.000$ \\
 &  & ISOMERIC & $0.000 \pm 0.000$ & $0.000 \pm 0.000$ & $0.000 \pm 0.000$ \\
 &  & CustOmics & $0.000 \pm 0.000$ & $0.000 \pm 0.000$ & $0.000 \pm 0.000$ \\
\addlinespace[1pt]
\textit{BRCA1} & $0.139 \pm 0.031$ & TMO-Net & $0.075 \pm 0.041$ & $-0.064 \pm 0.064$ & $0.074 \pm 0.049$ \\
 &  & ISOMERIC & $0.034 \pm 0.017$ & $-0.105 \pm 0.046$ & $0.105 \pm 0.046$ \\
 &  & CustOmics & $0.072 \pm 0.045$ & $-0.067 \pm 0.073$ & $0.083 \pm 0.051$ \\
\addlinespace[1pt]
\textit{BRCA2} & $0.156 \pm 0.056$ & TMO-Net & $0.079 \pm 0.054$ & $-0.077 \pm 0.110$ & $0.090 \pm 0.095$ \\
 &  & ISOMERIC & $0.053 \pm 0.052$ & $-0.104 \pm 0.107$ & $0.108 \pm 0.101$ \\
 &  & CustOmics & $0.045 \pm 0.028$ & $-0.112 \pm 0.080$ & $0.112 \pm 0.080$ \\
\addlinespace[1pt]
\textit{RAD51C} & $0.024 \pm 0.041$ & TMO-Net & $0.014 \pm 0.013$ & $-0.010 \pm 0.053$ & $0.037 \pm 0.029$ \\
 &  & ISOMERIC & $0.003 \pm 0.005$ & $-0.021 \pm 0.044$ & $0.027 \pm 0.039$ \\
 &  & CustOmics & $0.008 \pm 0.012$ & $-0.016 \pm 0.049$ & $0.032 \pm 0.035$ \\
\addlinespace[1pt]
\textit{RAD51D} & $0.206 \pm 0.035$ & TMO-Net & $0.143 \pm 0.040$ & $-0.064 \pm 0.070$ & $0.071 \pm 0.060$ \\
 &  & ISOMERIC & $0.083 \pm 0.028$ & $-0.123 \pm 0.061$ & $0.123 \pm 0.061$ \\
 &  & CustOmics & $0.192 \pm 0.071$ & $-0.014 \pm 0.057$ & $0.062 \pm 0.015$ \\
\bottomrule
\end{tabular*}
\end{table*}

\begin{table*}[!ht]
\centering
\caption{\textbf{RNA co-expression preservation in the before-treatment primary cohort.}
Synthetic rows compare synthetic with held-out real networks, the final row in
each panel compares real training with held-out real networks. Networks use
equal-sized cohorts, the same 3,000-gene panel, positive Pearson correlations
$\geq 0.70$, and Benjamini--Hochberg-adjusted $p<0.05$. Values are the mean $\pm$
sample SD across three patient-disjoint outer folds after averaging three model
initializations within each fold. Bold marks the highest model mean except for the density
ratio, for which it marks the mean closest to one.}
\label{tab:before-coexpression-full}
\scriptsize
\setlength{\tabcolsep}{2pt}

\textbf{(a) Network preservation}\\[2pt]
{\renewcommand{\arraystretch}{1.25}
\begin{tabular}{|>{\raggedright\arraybackslash}p{1.7cm}|*{4}{>{\centering\arraybackslash}p{\dimexpr(\textwidth-1.7cm-10\tabcolsep-6\arrayrulewidth)/4\relax}|}}
\hline
\rowcolor{black!12}
\textbf{Method} & \textbf{Fisher-$z$ concordance $\uparrow$} & \textbf{Edge F1 $\uparrow$} & \textbf{Module ARI $\uparrow$} & \textbf{Hub Jaccard $\uparrow$} \\ \hline
ISOMERIC & $\textbf{0.6897} \pm \textbf{0.0125}$ & $0.4121 \pm 0.0180$ & $0.6859 \pm 0.0836$ & $\textbf{0.2599} \pm \textbf{0.0385}$ \\ \hline
TMO-Net & $0.6837 \pm 0.0278$ & $\textbf{0.4737} \pm \textbf{0.0097}$ & $\textbf{0.7216} \pm \textbf{0.0548}$ & $0.2505 \pm 0.0756$ \\ \hline
CustOmics & $0.6805 \pm 0.0016$ & $0.2256 \pm 0.0337$ & $0.6258 \pm 0.0585$ & $0.1761 \pm 0.0640$ \\ \hline
Real train--test & $0.7438 \pm 0.0117$ & $0.5849 \pm 0.0345$ & $0.8475 \pm 0.0267$ & $0.4244 \pm 0.1403$ \\ \hline
\end{tabular}
}

\vspace{4pt}
\textbf{(b) Edge diagnostics}\\[2pt]
{\renewcommand{\arraystretch}{1.25}
\begin{tabular}{|>{\raggedright\arraybackslash}p{1.7cm}|*{3}{>{\centering\arraybackslash}p{\dimexpr(\textwidth-1.7cm-8\tabcolsep-5\arrayrulewidth)/3\relax}|}}
\hline
\rowcolor{black!12}
\textbf{Method} & \textbf{Edge precision $\uparrow$} & \textbf{Edge recall $\uparrow$} & \textbf{Density ratio $\rightarrow 1$} \\ \hline
ISOMERIC & $0.2896 \pm 0.0209$ & $0.7306 \pm 0.0307$ & $2.5755 \pm 0.2995$ \\ \hline
TMO-Net & $\textbf{0.3889} \pm \textbf{0.0637}$ & $0.6476 \pm 0.1257$ & $\textbf{1.7802} \pm \textbf{0.6328}$ \\ \hline
CustOmics & $0.1316 \pm 0.0249$ & $\textbf{0.8365} \pm \textbf{0.0722}$ & $6.7102 \pm 1.7398$ \\ \hline
Real train--test & $0.6077 \pm 0.0519$ & $0.5739 \pm 0.0884$ & $0.9560 \pm 0.2152$ \\ \hline
\end{tabular}
}
\end{table*}

\begin{table*}[!ht]
\centering
\caption{\textbf{RNA co-expression preservation in the after-treatment cohort.}
Synthetic rows compare synthetic with held-out real networks, the final row in
each panel compares real training with held-out real networks. Networks use
equal-sized cohorts, the same 3,000-gene panel, positive Pearson correlations
$\geq 0.70$, and Benjamini--Hochberg-adjusted $p<0.05$. Values are the mean $\pm$
sample SD across three patient-disjoint outer folds after averaging three model
initializations within each fold. Bold marks the highest model mean except for the density
ratio, for which it marks the mean closest to one.}
\label{tab:after-coexpression-full}
\scriptsize
\setlength{\tabcolsep}{2pt}

\textbf{(a) Network preservation}\\[2pt]
{\renewcommand{\arraystretch}{1.25}
\begin{tabular}{|>{\raggedright\arraybackslash}p{1.7cm}|*{4}{>{\centering\arraybackslash}p{\dimexpr(\textwidth-1.7cm-10\tabcolsep-6\arrayrulewidth)/4\relax}|}}
\hline
\rowcolor{black!12}
\textbf{Method} & \textbf{Fisher-$z$ concordance $\uparrow$} & \textbf{Edge F1 $\uparrow$} & \textbf{Module ARI $\uparrow$} & \textbf{Hub Jaccard $\uparrow$} \\ \hline
ISOMERIC & $0.5389 \pm 0.0437$ & $0.2456 \pm 0.0150$ & $0.3105 \pm 0.0337$ & $0.0994 \pm 0.0342$ \\ \hline
TMO-Net & $\textbf{0.5552} \pm \textbf{0.0167}$ & $\textbf{0.2539} \pm \textbf{0.0164}$ & $\textbf{0.3325} \pm \textbf{0.0442}$ & $\textbf{0.1077} \pm \textbf{0.0456}$ \\ \hline
CustOmics & $0.5335 \pm 0.0156$ & $0.1472 \pm 0.0135$ & $0.2762 \pm 0.0682$ & $0.0561 \pm 0.0249$ \\ \hline
Real train--test & $0.5840 \pm 0.0238$ & $0.3372 \pm 0.0292$ & $0.4323 \pm 0.0376$ & $0.2449 \pm 0.1367$ \\ \hline
\end{tabular}
}

\vspace{4pt}
\textbf{(b) Edge diagnostics}\\[2pt]
{\renewcommand{\arraystretch}{1.25}
\begin{tabular}{|>{\raggedright\arraybackslash}p{1.7cm}|*{3}{>{\centering\arraybackslash}p{\dimexpr(\textwidth-1.7cm-8\tabcolsep-5\arrayrulewidth)/3\relax}|}}
\hline
\rowcolor{black!12}
\textbf{Method} & \textbf{Edge precision $\uparrow$} & \textbf{Edge recall $\uparrow$} & \textbf{Density ratio $\rightarrow 1$} \\ \hline
ISOMERIC & $0.1779 \pm 0.0122$ & $0.4119 \pm 0.0470$ & $2.3747 \pm 0.2633$ \\ \hline
TMO-Net & $\textbf{0.1887} \pm \textbf{0.0220}$ & $0.4058 \pm 0.0487$ & $\textbf{2.2251} \pm \textbf{0.4432}$ \\ \hline
CustOmics & $0.0854 \pm 0.0094$ & $\textbf{0.5561} \pm \textbf{0.0215}$ & $6.7724 \pm 0.7913$ \\ \hline
Real train--test & $0.3292 \pm 0.0236$ & $0.3467 \pm 0.0426$ & $1.0528 \pm 0.0976$ \\ \hline
\end{tabular}
}
\end{table*}

\begin{table*}[!ht]
\centering
\caption{\textbf{Detailed cis RNA-CNV association results for the before-treatment primary cohort.} Values are the mean $\pm$ sample SD across three patient-disjoint outer folds after averaging three model initializations within each fold. The RNA-CNV panel intersections contain 320, 313, and 313 genes in the three folds, respectively, all have estimable slopes for every model and initialization. Cells show train--synthetic, test--synthetic, and the real train--test reference in brackets. {Sign agreement is the fraction of genes with matching slope signs among finite, nonzero pairs, all shared genes satisfy this criterion here. Bold marks the highest correlation or sign agreement, or lowest error, separately for train and test.}}
\label{tab:before-integration-full}
\scriptsize
\setlength{\tabcolsep}{3pt}
\textbf{Same-gene copy-number impact}\\[2pt]
{\renewcommand{\arraystretch}{1.45}
\begin{tabular}{|>{\raggedright\arraybackslash}p{1.55cm}|*{5}{>{\centering\arraybackslash}p{\dimexpr(\textwidth-1.55cm-12\tabcolsep-7\arrayrulewidth)/5\relax}|}}
\hline
\rowcolor{black!12}
\textbf{Method} & \textbf{Pearson of slopes $\uparrow$} & \textbf{Median $|\Delta\beta|$ $\downarrow$} & {\textbf{Spearman of slopes $\uparrow$}} & {\textbf{Mean $|\Delta\beta|$ $\downarrow$}} & {\textbf{Sign agreement $\uparrow$}} \\ \hline
TMO-Net & {\renewcommand{\arraystretch}{1.25}\begin{tabular}{@{}c@{}}$ \textbf{0.8349} \pm \textbf{0.0233} $\\$ \textbf{0.5996} \pm \textbf{0.0475} $\\{}[$ 0.7425 \pm 0.0455 $]\end{tabular}} & {\renewcommand{\arraystretch}{1.25}\begin{tabular}{@{}c@{}}$ 0.0972 \pm 0.0107 $\\$ 0.1518 \pm 0.0084 $\\{}[$ 0.0931 \pm 0.0063 $]\end{tabular}} & {\renewcommand{\arraystretch}{1.25}\begin{tabular}{@{}c@{}}$ \textbf{0.8514} \pm \textbf{0.0218} $\\$ 0.6209 \pm 0.0343 $\\{}[$ 0.7581 \pm 0.0326 $]\end{tabular}} & {\renewcommand{\arraystretch}{1.25}\begin{tabular}{@{}c@{}}$ 0.1267 \pm 0.0143 $\\$ 0.1936 \pm 0.0074 $\\{}[$ 0.1291 \pm 0.0060 $]\end{tabular}} & {\renewcommand{\arraystretch}{1.25}\begin{tabular}{@{}c@{}}$ \textbf{0.9485} \pm \textbf{0.0037} $\\$ \textbf{0.9266} \pm \textbf{0.0110} $\\{}[$ 0.9577 \pm 0.0084 $]\end{tabular}} \\ \hline
ISOMERIC & {\renewcommand{\arraystretch}{1.25}\begin{tabular}{@{}c@{}}$ 0.7365 \pm 0.0874 $\\$ 0.5506 \pm 0.1010 $\\{}[$ 0.7425 \pm 0.0455 $]\end{tabular}} & {\renewcommand{\arraystretch}{1.25}\begin{tabular}{@{}c@{}}$ \textbf{0.0758} \pm \textbf{0.0008} $\\$ \textbf{0.1176} \pm \textbf{0.0108} $\\{}[$ 0.0931 \pm 0.0063 $]\end{tabular}} & {\renewcommand{\arraystretch}{1.25}\begin{tabular}{@{}c@{}}$ 0.8007 \pm 0.0681 $\\$ \textbf{0.6602} \pm \textbf{0.0617} $\\{}[$ 0.7581 \pm 0.0326 $]\end{tabular}} & {\renewcommand{\arraystretch}{1.25}\begin{tabular}{@{}c@{}}$ \textbf{0.1164} \pm \textbf{0.0123} $\\$ \textbf{0.1670} \pm \textbf{0.0105} $\\{}[$ 0.1291 \pm 0.0060 $]\end{tabular}} & {\renewcommand{\arraystretch}{1.25}\begin{tabular}{@{}c@{}}$ 0.9465 \pm 0.0116 $\\$ 0.9190 \pm 0.0029 $\\{}[$ 0.9577 \pm 0.0084 $]\end{tabular}} \\ \hline
CustOmics & {\renewcommand{\arraystretch}{1.25}\begin{tabular}{@{}c@{}}$ 0.7828 \pm 0.0326 $\\$ 0.5669 \pm 0.0491 $\\{}[$ 0.7425 \pm 0.0455 $]\end{tabular}} & {\renewcommand{\arraystretch}{1.25}\begin{tabular}{@{}c@{}}$ 0.1906 \pm 0.0146 $\\$ 0.2154 \pm 0.0141 $\\{}[$ 0.0931 \pm 0.0063 $]\end{tabular}} & {\renewcommand{\arraystretch}{1.25}\begin{tabular}{@{}c@{}}$ 0.8005 \pm 0.0286 $\\$ 0.5991 \pm 0.0440 $\\{}[$ 0.7581 \pm 0.0326 $]\end{tabular}} & {\renewcommand{\arraystretch}{1.25}\begin{tabular}{@{}c@{}}$ 0.2034 \pm 0.0088 $\\$ 0.2470 \pm 0.0081 $\\{}[$ 0.1291 \pm 0.0060 $]\end{tabular}} & {\renewcommand{\arraystretch}{1.25}\begin{tabular}{@{}c@{}}$ 0.8943 \pm 0.0069 $\\$ 0.8866 \pm 0.0083 $\\{}[$ 0.9577 \pm 0.0084 $]\end{tabular}} \\ \hline
\end{tabular}
}
\end{table*}

\begin{table*}[!ht]
\centering
\caption{\textbf{Detailed cis RNA-CNV association results for the after-treatment cohort.} Values are the mean $\pm$ sample SD across three patient-disjoint outer folds after averaging three model initializations within each fold. The RNA-CNV panel intersections contain 349, 291, and 293 genes in the three folds, respectively, all have estimable slopes for every model and initialization. Cells show train--synthetic, test--synthetic, and the real train--test reference in brackets. {Sign agreement is the fraction of genes with matching slope signs among finite, nonzero pairs, all shared genes satisfy this criterion here. Bold marks the highest correlation or sign agreement, or lowest error, separately for train and test.}}
\label{tab:after-integration-full}
\scriptsize
\setlength{\tabcolsep}{3pt}
\textbf{Same-gene copy-number impact}\\[2pt]
{\renewcommand{\arraystretch}{1.45}
\begin{tabular}{|>{\raggedright\arraybackslash}p{1.55cm}|*{5}{>{\centering\arraybackslash}p{\dimexpr(\textwidth-1.55cm-12\tabcolsep-7\arrayrulewidth)/5\relax}|}}
\hline
\rowcolor{black!12}
\textbf{Method} & \textbf{Pearson of slopes $\uparrow$} & \textbf{Median $|\Delta\beta|$ $\downarrow$} & {\textbf{Spearman of slopes $\uparrow$}} & {\textbf{Mean $|\Delta\beta|$ $\downarrow$}} & {\textbf{Sign agreement $\uparrow$}} \\ \hline
TMO-Net & {\renewcommand{\arraystretch}{1.25}\begin{tabular}{@{}c@{}}$ \textbf{0.7686} \pm \textbf{0.0180} $\\$ 0.3049 \pm 0.0419 $\\{}[$ 0.4235 \pm 0.0744 $]\end{tabular}} & {\renewcommand{\arraystretch}{1.25}\begin{tabular}{@{}c@{}}$ 0.1167 \pm 0.0130 $\\$ 0.1987 \pm 0.0093 $\\{}[$ 0.1697 \pm 0.0188 $]\end{tabular}} & {\renewcommand{\arraystretch}{1.25}\begin{tabular}{@{}c@{}}$ 0.7620 \pm 0.0196 $\\$ 0.3270 \pm 0.0388 $\\{}[$ 0.4594 \pm 0.0793 $]\end{tabular}} & {\renewcommand{\arraystretch}{1.25}\begin{tabular}{@{}c@{}}$ 0.1458 \pm 0.0106 $\\$ 0.2534 \pm 0.0057 $\\{}[$ 0.2139 \pm 0.0075 $]\end{tabular}} & {\renewcommand{\arraystretch}{1.25}\begin{tabular}{@{}c@{}}$ \textbf{0.9342} \pm \textbf{0.0041} $\\$ \textbf{0.8921} \pm \textbf{0.0065} $\\{}[$ 0.9297 \pm 0.0133 $]\end{tabular}} \\ \hline
ISOMERIC & {\renewcommand{\arraystretch}{1.25}\begin{tabular}{@{}c@{}}$ 0.7622 \pm 0.0291 $\\$ \textbf{0.3389} \pm \textbf{0.0755} $\\{}[$ 0.4235 \pm 0.0744 $]\end{tabular}} & {\renewcommand{\arraystretch}{1.25}\begin{tabular}{@{}c@{}}$ \textbf{0.1157} \pm \textbf{0.0131} $\\$ \textbf{0.1741} \pm \textbf{0.0014} $\\{}[$ 0.1697 \pm 0.0188 $]\end{tabular}} & {\renewcommand{\arraystretch}{1.25}\begin{tabular}{@{}c@{}}$ \textbf{0.7856} \pm \textbf{0.0282} $\\$ \textbf{0.3614} \pm \textbf{0.0659} $\\{}[$ 0.4594 \pm 0.0793 $]\end{tabular}} & {\renewcommand{\arraystretch}{1.25}\begin{tabular}{@{}c@{}}$ \textbf{0.1430} \pm \textbf{0.0121} $\\$ \textbf{0.2343} \pm \textbf{0.0034} $\\{}[$ 0.2139 \pm 0.0075 $]\end{tabular}} & {\renewcommand{\arraystretch}{1.25}\begin{tabular}{@{}c@{}}$ 0.9260 \pm 0.0104 $\\$ 0.8807 \pm 0.0059 $\\{}[$ 0.9297 \pm 0.0133 $]\end{tabular}} \\ \hline
CustOmics & {\renewcommand{\arraystretch}{1.25}\begin{tabular}{@{}c@{}}$ 0.6333 \pm 0.0481 $\\$ 0.2226 \pm 0.0829 $\\{}[$ 0.4235 \pm 0.0744 $]\end{tabular}} & {\renewcommand{\arraystretch}{1.25}\begin{tabular}{@{}c@{}}$ 0.2334 \pm 0.0563 $\\$ 0.2483 \pm 0.0109 $\\{}[$ 0.1697 \pm 0.0188 $]\end{tabular}} & {\renewcommand{\arraystretch}{1.25}\begin{tabular}{@{}c@{}}$ 0.6590 \pm 0.0586 $\\$ 0.2479 \pm 0.0678 $\\{}[$ 0.4594 \pm 0.0793 $]\end{tabular}} & {\renewcommand{\arraystretch}{1.25}\begin{tabular}{@{}c@{}}$ 0.2539 \pm 0.0403 $\\$ 0.2998 \pm 0.0146 $\\{}[$ 0.2139 \pm 0.0075 $]\end{tabular}} & {\renewcommand{\arraystretch}{1.25}\begin{tabular}{@{}c@{}}$ 0.8667 \pm 0.0136 $\\$ 0.8351 \pm 0.0050 $\\{}[$ 0.9297 \pm 0.0133 $]\end{tabular}} \\ \hline
\end{tabular}
}
\end{table*}

\begin{table*}[!ht]
\centering
\caption{\textbf{Training-proximity diagnostics for the before-treatment primary cohort.} Values are the mean $\pm$ sample SD across three patient-disjoint outer folds after averaging three model initializations within each fold. A nearest-neighbor distance ratio of $1$ indicates equal proximity to training and held-out samples, a membership AUC of $0.5$ indicates no separation between the two groups. For paired nearest-neighbor agreement, the held-out real-pair baseline is shown in brackets. Bold marks the model mean closest to one for the distance ratio, 0.5 for membership AUC, and the real-pair baseline for paired nearest-neighbor agreement.}
\label{tab:before-training-diagnostics-full}
\scriptsize
\setlength{\tabcolsep}{3pt}
\textbf{(a) RNA diagnostics}\\[2pt]
{\renewcommand{\arraystretch}{1.35}
\begin{tabular}{|>{\raggedright\arraybackslash}p{1.55cm}|*{2}{>{\centering\arraybackslash}p{\dimexpr(\textwidth-1.55cm-6\tabcolsep-4\arrayrulewidth)/2\relax}|}}
\hline
\rowcolor{black!12}
\textbf{Method} & \textbf{Median NN ratio $\rightarrow 1$} & \textbf{Membership AUC $\rightarrow 0.5$} \\ \hline
TMO-Net & $ 0.8857 \pm 0.0227 $ & $ 0.8039 \pm 0.0347 $ \\ \hline
ISOMERIC & $ \textbf{0.9501} \pm \textbf{0.0101} $ & $ \textbf{0.6469} \pm \textbf{0.0495} $ \\ \hline
CustOmics & $ 0.9380 \pm 0.0094 $ & $ 0.6887 \pm 0.0450 $ \\ \hline
\end{tabular}
}
\vspace{3pt}
\textbf{(b) CNV diagnostics}\\[2pt]
{\renewcommand{\arraystretch}{1.35}
\begin{tabular}{|>{\raggedright\arraybackslash}p{1.55cm}|*{2}{>{\centering\arraybackslash}p{\dimexpr(\textwidth-1.55cm-6\tabcolsep-4\arrayrulewidth)/2\relax}|}}
\hline
\rowcolor{black!12}
\textbf{Method} & \textbf{Median NN ratio $\rightarrow 1$} & \textbf{Membership AUC $\rightarrow 0.5$} \\ \hline
TMO-Net & $ 0.9825 \pm 0.0200 $ & $ 0.5538 \pm 0.0581 $ \\ \hline
ISOMERIC & $ \textbf{1.0074} \pm \textbf{0.0203} $ & $ 0.3884 \pm 0.0375 $ \\ \hline
CustOmics & $ 0.9853 \pm 0.0306 $ & $ \textbf{0.4846} \pm \textbf{0.0505} $ \\ \hline
\end{tabular}
}
\vspace{3pt}
\textbf{(c) Paired RNA-CNV diagnostics}\\[2pt]
{\renewcommand{\arraystretch}{1.35}
\begin{tabular}{|>{\raggedright\arraybackslash}p{1.55cm}|*{3}{>{\centering\arraybackslash}p{\dimexpr(\textwidth-1.55cm-8\tabcolsep-5\arrayrulewidth)/3\relax}|}}
\hline
\rowcolor{black!12}
\textbf{Method} & \textbf{Median NN ratio $\rightarrow 1$} & \textbf{Membership AUC $\rightarrow 0.5$} & \textbf{Paired NN agreement} \\ \hline
TMO-Net & $ 0.9582 \pm 0.0156 $ & $ 0.6570 \pm 0.0425 $ & {\renewcommand{\arraystretch}{1.25}\begin{tabular}{@{}c@{}}$ 0.0895 \pm 0.0126 $\\{}[$ 0.0163 \pm 0.0037 $]\end{tabular}} \\ \hline
ISOMERIC & $ \textbf{0.9923} \pm \textbf{0.0097} $ & $ 0.4462 \pm 0.0354 $ & {\renewcommand{\arraystretch}{1.25}\begin{tabular}{@{}c@{}}$ 0.0273 \pm 0.0018 $\\{}[$ 0.0163 \pm 0.0037 $]\end{tabular}} \\ \hline
CustOmics & $ 0.9861 \pm 0.0154 $ & $ \textbf{0.5241} \pm \textbf{0.0463} $ & {\renewcommand{\arraystretch}{1.25}\begin{tabular}{@{}c@{}}$ \textbf{0.0156} \pm \textbf{0.0031} $\\{}[$ 0.0163 \pm 0.0037 $]\end{tabular}} \\ \hline
\end{tabular}
}

\par
\caption{\textbf{Training-proximity diagnostics for the after-treatment cohort.} Values are the mean $\pm$ sample SD across three patient-disjoint outer folds after averaging three model initializations within each fold. A nearest-neighbor distance ratio of $1$ indicates equal proximity to training and held-out samples, a membership AUC of $0.5$ indicates no separation between the two groups. For paired nearest-neighbor agreement, the held-out real-pair baseline is shown in brackets. Bold marks the model mean closest to one for the distance ratio, 0.5 for membership AUC, and the real-pair baseline for paired nearest-neighbor agreement.}
\label{tab:after-training-diagnostics-full}
\scriptsize
\setlength{\tabcolsep}{3pt}
\textbf{(a) RNA diagnostics}\\[2pt]
{\renewcommand{\arraystretch}{1.35}
\begin{tabular}{|>{\raggedright\arraybackslash}p{1.55cm}|*{2}{>{\centering\arraybackslash}p{\dimexpr(\textwidth-1.55cm-6\tabcolsep-4\arrayrulewidth)/2\relax}|}}
\hline
\rowcolor{black!12}
\textbf{Method} & \textbf{Median NN ratio $\rightarrow 1$} & \textbf{Membership AUC $\rightarrow 0.5$} \\ \hline
TMO-Net & $ 0.8105 \pm 0.0212 $ & $ 0.8564 \pm 0.0165 $ \\ \hline
ISOMERIC & $ \textbf{0.9154} \pm \textbf{0.0218} $ & $ \textbf{0.6861} \pm \textbf{0.0250} $ \\ \hline
CustOmics & $ 0.9067 \pm 0.0134 $ & $ 0.7199 \pm 0.0243 $ \\ \hline
\end{tabular}
}
\vspace{3pt}
\textbf{(b) CNV diagnostics}\\[2pt]
{\renewcommand{\arraystretch}{1.35}
\begin{tabular}{|>{\raggedright\arraybackslash}p{1.55cm}|*{2}{>{\centering\arraybackslash}p{\dimexpr(\textwidth-1.55cm-6\tabcolsep-4\arrayrulewidth)/2\relax}|}}
\hline
\rowcolor{black!12}
\textbf{Method} & \textbf{Median NN ratio $\rightarrow 1$} & \textbf{Membership AUC $\rightarrow 0.5$} \\ \hline
TMO-Net & $ 0.9738 \pm 0.0561 $ & $ 0.5984 \pm 0.0903 $ \\ \hline
ISOMERIC & $ \textbf{0.9985} \pm \textbf{0.0453} $ & $ \textbf{0.4615} \pm \textbf{0.0820} $ \\ \hline
CustOmics & $ 0.9783 \pm 0.0526 $ & $ 0.5492 \pm 0.0882 $ \\ \hline
\end{tabular}
}
\vspace{3pt}
\textbf{(c) Paired RNA-CNV diagnostics}\\[2pt]
{\renewcommand{\arraystretch}{1.35}
\begin{tabular}{|>{\raggedright\arraybackslash}p{1.55cm}|*{3}{>{\centering\arraybackslash}p{\dimexpr(\textwidth-1.55cm-8\tabcolsep-5\arrayrulewidth)/3\relax}|}}
\hline
\rowcolor{black!12}
\textbf{Method} & \textbf{Median NN ratio $\rightarrow 1$} & \textbf{Membership AUC $\rightarrow 0.5$} & \textbf{Paired NN agreement} \\ \hline
TMO-Net & $ 0.9124 \pm 0.0358 $ & $ 0.7712 \pm 0.0594 $ & {\renewcommand{\arraystretch}{1.25}\begin{tabular}{@{}c@{}}$ 0.1669 \pm 0.0294 $\\{}[$ 0.0295 \pm 0.0361 $]\end{tabular}} \\ \hline
ISOMERIC & $ \textbf{0.9730} \pm \textbf{0.0222} $ & $ \textbf{0.5631} \pm \textbf{0.0697} $ & {\renewcommand{\arraystretch}{1.25}\begin{tabular}{@{}c@{}}$ 0.0640 \pm 0.0146 $\\{}[$ 0.0295 \pm 0.0361 $]\end{tabular}} \\ \hline
CustOmics & $ 0.9699 \pm 0.0231 $ & $ 0.5978 \pm 0.0722 $ & {\renewcommand{\arraystretch}{1.25}\begin{tabular}{@{}c@{}}$ \textbf{0.0439} \pm \textbf{0.0113} $\\{}[$ 0.0295 \pm 0.0361 $]\end{tabular}} \\ \hline
\end{tabular}
}
\end{table*}

\begingroup
\raggedbottom
\clearpage
\section{Paired model comparisons across outer folds}
\label{app:paired-comparisons}

To show whether model orderings persist across the evaluated folds, we report
paired differences for all 33 metrics reported in
Appendix~\ref{app:synthetic-data-evaluation}. Fidelity, co-expression and cis
association use held-out references, training-proximity diagnostics retain
their original definitions. Let $m_{A,f}$ be the metric for model $A$ in fold
$f$, averaged over training seeds 42, 43 and 44.
Tables~\ref{tab:paired-before} and~\ref{tab:paired-after} give
$\Delta_f=m_{A,f}-m_{B,f}$ for each fold and model pair, together with the
mean across the three folds.

Positive differences indicate a higher score for the first model, a sign
change indicates a reversal of the observed ordering. Higher scores are not
uniformly preferable, particularly for ratios and proximity diagnostics.
These comparisons are descriptive: held-out patient groups are disjoint,
but training sets overlap, and selected gene panels can differ between folds.
The three fold differences are therefore not independent replications and
do not establish statistically significant superiority or equivalence.

\begingroup
\scriptsize
\setlength{\tabcolsep}{4pt}
\renewcommand{\arraystretch}{1.04}
\setlength{\LTcapwidth}{\linewidth}
\begin{longtable}{@{}p{0.40\linewidth}lrrrr@{}}
\caption{\textbf{Before-treatment paired model differences.} T: TMO-Net, I: ISOMERIC, C: CustOmics. Each contrast is the first model minus the second after averaging three training seeds within fold. Arrows indicate preferred directions or targets for the original scores, differences retain their original units. NN: nearest neighbor, AUC: area under the receiver operating characteristic curve.}\label{tab:paired-before}\\
\toprule
Metric & $A-B$ & $\Delta_0$ & $\Delta_1$ & $\Delta_2$ & Mean \\
\midrule
\endfirsthead
\multicolumn{6}{l}{\tablename~\thetable: Before-treatment paired differences (continued).}\\
\toprule
Metric & $A-B$ & $\Delta_0$ & $\Delta_1$ & $\Delta_2$ & Mean \\
\midrule
\endhead
\midrule
\multicolumn{6}{r}{Continued on next page.}\\
\endfoot
\bottomrule
\endlastfoot
\addlinespace[3pt]
\multicolumn{6}{l}{\textbf{RNA fidelity}}\\*
Mean gene-wise $W_1$ $\downarrow$ & T--I & $-0.0069$ & $-0.0339$ & $-0.0118$ & $-0.0175$ \\*
 & T--C & $+0.0142$ & $-0.0242$ & $-0.0474$ & $-0.0191$ \\*
 & I--C & $+0.0211$ & $+0.0097$ & $-0.0356$ & $-0.0016$ \\
\addlinespace[1pt]
Mean Jensen--Shannon distance $\downarrow$ & T--I & $-0.0075$ & $-0.0164$ & $-0.0053$ & $-0.0097$ \\*
 & T--C & $+0.0038$ & $-0.0096$ & $-0.0180$ & $-0.0079$ \\*
 & I--C & $+0.0113$ & $+0.0068$ & $-0.0127$ & $+0.0018$ \\
\addlinespace[1pt]
RBF MMD$^2$ $\downarrow$ & T--I & $+0.0107$ & $-0.0012$ & $+0.0002$ & $+0.0033$ \\*
 & T--C & $+0.0140$ & $-0.0060$ & $-0.0206$ & $-0.0042$ \\*
 & I--C & $+0.0033$ & $-0.0049$ & $-0.0209$ & $-0.0075$ \\
\addlinespace[1pt]
Median variance ratio ($\to 1$) & T--I & $+0.0564$ & $+0.1708$ & $+0.0458$ & $+0.0910$ \\*
 & T--C & $-0.0225$ & $+0.0888$ & $+0.1146$ & $+0.0603$ \\*
 & I--C & $-0.0789$ & $-0.0820$ & $+0.0689$ & $-0.0307$ \\
\addlinespace[1pt]
Gene-mean correlation $\uparrow$ & T--I & $-0.0034$ & $-0.0009$ & $-0.0004$ & $-0.0016$ \\*
 & T--C & $-0.0028$ & $+0.0019$ & $+0.0034$ & $+0.0009$ \\*
 & I--C & $+0.0007$ & $+0.0028$ & $+0.0038$ & $+0.0024$ \\
\addlinespace[1pt]
Discriminative score $\downarrow$ & T--I & $+0.1646$ & $+0.0799$ & $+0.0926$ & $+0.1123$ \\*
 & T--C & $+0.0313$ & $-0.1907$ & $-0.2459$ & $-0.1351$ \\*
 & I--C & $-0.1333$ & $-0.2705$ & $-0.3385$ & $-0.2474$ \\
\addlinespace[1pt]
\addlinespace[3pt]
\multicolumn{6}{l}{\textbf{CNV fidelity}}\\*
Mean amplification error $\downarrow$ & T--I & $-0.0119$ & $+0.0027$ & $-0.0213$ & $-0.0101$ \\*
 & T--C & $-0.0076$ & $-0.0009$ & $+0.0002$ & $-0.0028$ \\*
 & I--C & $+0.0044$ & $-0.0036$ & $+0.0214$ & $+0.0074$ \\
\addlinespace[1pt]
Mean deletion error $\downarrow$ & T--I & $-0.0157$ & $+0.0031$ & $+0.0219$ & $+0.0031$ \\*
 & T--C & $-0.0309$ & $-0.0640$ & $-0.0007$ & $-0.0319$ \\*
 & I--C & $-0.0153$ & $-0.0671$ & $-0.0226$ & $-0.0350$ \\
\addlinespace[1pt]
Amplification Pearson $\uparrow$ & T--I & $+0.0340$ & $-0.0159$ & $+0.0123$ & $+0.0101$ \\*
 & T--C & $+0.0204$ & $+0.0043$ & $-0.0163$ & $+0.0028$ \\*
 & I--C & $-0.0136$ & $+0.0202$ & $-0.0286$ & $-0.0073$ \\
\addlinespace[1pt]
Deletion Pearson $\uparrow$ & T--I & $+0.0577$ & $-0.0113$ & $-0.0471$ & $-0.0002$ \\*
 & T--C & $+0.0957$ & $+0.0419$ & $+0.0480$ & $+0.0619$ \\*
 & I--C & $+0.0379$ & $+0.0532$ & $+0.0951$ & $+0.0621$ \\
\addlinespace[1pt]
Top-50 amplification overlap $\uparrow$ & T--I & $+0.0800$ & $-0.1000$ & $+0.1267$ & $+0.0356$ \\*
 & T--C & $+0.1133$ & $+0.2867$ & $+0.2267$ & $+0.2089$ \\*
 & I--C & $+0.0333$ & $+0.3867$ & $+0.1000$ & $+0.1733$ \\
\addlinespace[1pt]
Top-50 deletion overlap $\uparrow$ & T--I & $+0.0467$ & $+0.0600$ & $+0.0067$ & $+0.0378$ \\*
 & T--C & $+0.2067$ & $+0.0733$ & $+0.0600$ & $+0.1133$ \\*
 & I--C & $+0.1600$ & $+0.0133$ & $+0.0533$ & $+0.0756$ \\
\addlinespace[1pt]
\addlinespace[3pt]
\multicolumn{6}{l}{\textbf{Prespecified 14-biomarker panel}}\\*
Amplification mean absolute error $\downarrow$ & T--I & $-0.0016$ & $+0.0072$ & $+0.0053$ & $+0.0036$ \\*
 & T--C & $-0.0074$ & $+0.0077$ & $+0.0116$ & $+0.0040$ \\*
 & I--C & $-0.0057$ & $+0.0005$ & $+0.0063$ & $+0.0004$ \\
\addlinespace[1pt]
Deletion mean absolute error $\downarrow$ & T--I & $-0.0106$ & $-0.0172$ & $+0.0293$ & $+0.0005$ \\*
 & T--C & $-0.0242$ & $-0.0617$ & $-0.0207$ & $-0.0356$ \\*
 & I--C & $-0.0137$ & $-0.0445$ & $-0.0500$ & $-0.0361$ \\
\addlinespace[1pt]
\addlinespace[3pt]
\multicolumn{6}{l}{\textbf{Cis RNA--CNV association}}\\*
Pearson of slopes $\uparrow$ & T--I & $-0.0171$ & $+0.0984$ & $+0.0656$ & $+0.0490$ \\*
 & T--C & $+0.0095$ & $+0.0179$ & $+0.0708$ & $+0.0327$ \\*
 & I--C & $+0.0266$ & $-0.0805$ & $+0.0052$ & $-0.0162$ \\
\addlinespace[1pt]
Median absolute slope error $\downarrow$ & T--I & $+0.0507$ & $+0.0387$ & $+0.0132$ & $+0.0342$ \\*
 & T--C & $-0.0660$ & $-0.0463$ & $-0.0787$ & $-0.0637$ \\*
 & I--C & $-0.1166$ & $-0.0850$ & $-0.0920$ & $-0.0979$ \\
\addlinespace[1pt]
Spearman of slopes $\uparrow$ & T--I & $-0.0680$ & $-0.0092$ & $-0.0407$ & $-0.0393$ \\*
 & T--C & $-0.0004$ & $+0.0286$ & $+0.0373$ & $+0.0218$ \\*
 & I--C & $+0.0676$ & $+0.0378$ & $+0.0780$ & $+0.0611$ \\
\addlinespace[1pt]
Mean absolute slope error $\downarrow$ & T--I & $+0.0382$ & $+0.0272$ & $+0.0146$ & $+0.0267$ \\*
 & T--C & $-0.0629$ & $-0.0391$ & $-0.0582$ & $-0.0534$ \\*
 & I--C & $-0.1011$ & $-0.0663$ & $-0.0729$ & $-0.0801$ \\
\addlinespace[1pt]
Slope sign agreement $\uparrow$ & T--I & $+0.0208$ & $+0.0021$ & $0.0000$ & $+0.0077$ \\*
 & T--C & $+0.0615$ & $+0.0330$ & $+0.0256$ & $+0.0400$ \\*
 & I--C & $+0.0406$ & $+0.0309$ & $+0.0256$ & $+0.0324$ \\
\addlinespace[1pt]
\addlinespace[3pt]
\multicolumn{6}{l}{\textbf{RNA co-expression}}\\*
Fisher-$z$ concordance $\uparrow$ & T--I & $-0.0242$ & $-0.0330$ & $+0.0392$ & $-0.0060$ \\*
 & T--C & $-0.0155$ & $-0.0118$ & $+0.0366$ & $+0.0031$ \\*
 & I--C & $+0.0087$ & $+0.0212$ & $-0.0025$ & $+0.0091$ \\
\addlinespace[1pt]
Edge F1 $\uparrow$ & T--I & $+0.0490$ & $+0.0632$ & $+0.0724$ & $+0.0615$ \\*
 & T--C & $+0.2637$ & $+0.2192$ & $+0.2615$ & $+0.2481$ \\*
 & I--C & $+0.2147$ & $+0.1560$ & $+0.1891$ & $+0.1866$ \\
\addlinespace[1pt]
Module adjusted Rand index $\uparrow$ & T--I & $+0.0148$ & $+0.0752$ & $+0.0173$ & $+0.0358$ \\*
 & T--C & $+0.1080$ & $+0.1150$ & $+0.0645$ & $+0.0959$ \\*
 & I--C & $+0.0932$ & $+0.0398$ & $+0.0473$ & $+0.0601$ \\
\addlinespace[1pt]
Hub Jaccard $\uparrow$ & T--I & $-0.0278$ & $-0.0429$ & $+0.0423$ & $-0.0095$ \\*
 & T--C & $+0.1293$ & $+0.0183$ & $+0.0754$ & $+0.0744$ \\*
 & I--C & $+0.1571$ & $+0.0613$ & $+0.0331$ & $+0.0838$ \\
\addlinespace[1pt]
Edge precision $\uparrow$ & T--I & $+0.0641$ & $+0.1580$ & $+0.0759$ & $+0.0993$ \\*
 & T--C & $+0.2441$ & $+0.3010$ & $+0.2269$ & $+0.2573$ \\*
 & I--C & $+0.1800$ & $+0.1430$ & $+0.1510$ & $+0.1580$ \\
\addlinespace[1pt]
Edge recall $\uparrow$ & T--I & $-0.0584$ & $-0.1879$ & $-0.0027$ & $-0.0830$ \\*
 & T--C & $-0.1947$ & $-0.2438$ & $-0.1280$ & $-0.1888$ \\*
 & I--C & $-0.1363$ & $-0.0559$ & $-0.1253$ & $-0.1058$ \\
\addlinespace[1pt]
Edge density ratio ($\to 1$) & T--I & $-0.6092$ & $-1.2305$ & $-0.5461$ & $-0.7953$ \\*
 & T--C & $-5.7368$ & $-3.5961$ & $-5.4572$ & $-4.9300$ \\*
 & I--C & $-5.1276$ & $-2.3655$ & $-4.9111$ & $-4.1348$ \\
\addlinespace[1pt]
\addlinespace[3pt]
\multicolumn{6}{l}{\textbf{Training proximity}}\\*
RNA median NN ratio & T--I & $-0.0785$ & $-0.0545$ & $-0.0602$ & $-0.0644$ \\*
 & T--C & $-0.0682$ & $-0.0354$ & $-0.0531$ & $-0.0523$ \\*
 & I--C & $+0.0103$ & $+0.0191$ & $+0.0071$ & $+0.0121$ \\
\addlinespace[1pt]
RNA membership AUC & T--I & $+0.1396$ & $+0.1629$ & $+0.1684$ & $+0.1570$ \\*
 & T--C & $+0.1044$ & $+0.1069$ & $+0.1343$ & $+0.1152$ \\*
 & I--C & $-0.0352$ & $-0.0560$ & $-0.0341$ & $-0.0418$ \\
\addlinespace[1pt]
CNV median NN ratio & T--I & $-0.0498$ & $-0.0125$ & $-0.0123$ & $-0.0249$ \\*
 & T--C & $-0.0231$ & $+0.0201$ & $-0.0052$ & $-0.0027$ \\*
 & I--C & $+0.0267$ & $+0.0326$ & $+0.0071$ & $+0.0221$ \\
\addlinespace[1pt]
CNV membership AUC & T--I & $+0.2107$ & $+0.1281$ & $+0.1573$ & $+0.1654$ \\*
 & T--C & $+0.0954$ & $+0.0367$ & $+0.0756$ & $+0.0692$ \\*
 & I--C & $-0.1154$ & $-0.0914$ & $-0.0817$ & $-0.0962$ \\
\addlinespace[1pt]
Paired median NN ratio & T--I & $-0.0478$ & $-0.0227$ & $-0.0319$ & $-0.0341$ \\*
 & T--C & $-0.0408$ & $-0.0111$ & $-0.0318$ & $-0.0279$ \\*
 & I--C & $+0.0070$ & $+0.0116$ & $+0.0001$ & $+0.0063$ \\
\addlinespace[1pt]
Paired membership AUC & T--I & $+0.2272$ & $+0.1907$ & $+0.2144$ & $+0.2108$ \\*
 & T--C & $+0.1386$ & $+0.1123$ & $+0.1477$ & $+0.1329$ \\*
 & I--C & $-0.0886$ & $-0.0784$ & $-0.0667$ & $-0.0779$ \\
\addlinespace[1pt]
Paired NN agreement & T--I & $+0.0703$ & $+0.0479$ & $+0.0682$ & $+0.0621$ \\*
 & T--C & $+0.0853$ & $+0.0616$ & $+0.0745$ & $+0.0738$ \\*
 & I--C & $+0.0151$ & $+0.0137$ & $+0.0064$ & $+0.0117$ \\
\addlinespace[1pt]
\end{longtable}
\endgroup

\clearpage
\begingroup
\scriptsize
\setlength{\tabcolsep}{4pt}
\renewcommand{\arraystretch}{1.04}
\setlength{\LTcapwidth}{\linewidth}
\begin{longtable}{@{}p{0.40\linewidth}lrrrr@{}}
\caption{\textbf{After-treatment paired model differences.} T: TMO-Net, I: ISOMERIC, C: CustOmics. Each contrast is the first model minus the second after averaging three training seeds within fold. Arrows indicate preferred directions or targets for the original scores, differences retain their original units. NN: nearest neighbor, AUC: area under the receiver operating characteristic curve.}\label{tab:paired-after}\\
\toprule
Metric & $A-B$ & $\Delta_0$ & $\Delta_1$ & $\Delta_2$ & Mean \\
\midrule
\endfirsthead
\multicolumn{6}{l}{\tablename~\thetable: After-treatment paired differences (continued).}\\
\toprule
Metric & $A-B$ & $\Delta_0$ & $\Delta_1$ & $\Delta_2$ & Mean \\
\midrule
\endhead
\midrule
\multicolumn{6}{r}{Continued on next page.}\\
\endfoot
\bottomrule
\endlastfoot
\addlinespace[3pt]
\multicolumn{6}{l}{\textbf{RNA fidelity}}\\*
Mean gene-wise $W_1$ $\downarrow$ & T--I & $-0.0354$ & $-0.0295$ & $+0.0012$ & $-0.0212$ \\*
 & T--C & $-0.0826$ & $-0.0263$ & $-0.0687$ & $-0.0592$ \\*
 & I--C & $-0.0473$ & $+0.0032$ & $-0.0699$ & $-0.0380$ \\
\addlinespace[1pt]
Mean Jensen--Shannon distance $\downarrow$ & T--I & $-0.0093$ & $-0.0081$ & $-0.0023$ & $-0.0066$ \\*
 & T--C & $-0.0158$ & $-0.0031$ & $-0.0213$ & $-0.0134$ \\*
 & I--C & $-0.0064$ & $+0.0050$ & $-0.0190$ & $-0.0068$ \\
\addlinespace[1pt]
RBF MMD$^2$ $\downarrow$ & T--I & $-0.0034$ & $-0.0079$ & $+0.0167$ & $+0.0018$ \\*
 & T--C & $-0.0453$ & $-0.0106$ & $-0.0379$ & $-0.0313$ \\*
 & I--C & $-0.0419$ & $-0.0027$ & $-0.0547$ & $-0.0331$ \\
\addlinespace[1pt]
Median variance ratio ($\to 1$) & T--I & $+0.2089$ & $+0.1272$ & $+0.1061$ & $+0.1474$ \\*
 & T--C & $+0.1907$ & $+0.0784$ & $+0.2995$ & $+0.1896$ \\*
 & I--C & $-0.0182$ & $-0.0488$ & $+0.1935$ & $+0.0422$ \\
\addlinespace[1pt]
Gene-mean correlation $\uparrow$ & T--I & $+0.0001$ & $+0.0017$ & $-0.0079$ & $-0.0021$ \\*
 & T--C & $+0.0054$ & $+0.0042$ & $+0.0013$ & $+0.0037$ \\*
 & I--C & $+0.0053$ & $+0.0025$ & $+0.0093$ & $+0.0057$ \\
\addlinespace[1pt]
Discriminative score $\downarrow$ & T--I & $+0.0101$ & $-0.3004$ & $+0.1250$ & $-0.0551$ \\*
 & T--C & $-0.3915$ & $-0.3038$ & $-0.2316$ & $-0.3090$ \\*
 & I--C & $-0.4016$ & $-0.0034$ & $-0.3566$ & $-0.2539$ \\
\addlinespace[1pt]
\addlinespace[3pt]
\multicolumn{6}{l}{\textbf{CNV fidelity}}\\*
Mean amplification error $\downarrow$ & T--I & $-0.0019$ & $-0.0117$ & $-0.0055$ & $-0.0064$ \\*
 & T--C & $-0.0057$ & $+0.0056$ & $-0.0090$ & $-0.0030$ \\*
 & I--C & $-0.0038$ & $+0.0173$ & $-0.0035$ & $+0.0033$ \\
\addlinespace[1pt]
Mean deletion error $\downarrow$ & T--I & $+0.0222$ & $+0.0121$ & $-0.0237$ & $+0.0035$ \\*
 & T--C & $-0.1014$ & $-0.1410$ & $-0.0638$ & $-0.1021$ \\*
 & I--C & $-0.1235$ & $-0.1532$ & $-0.0401$ & $-0.1056$ \\
\addlinespace[1pt]
Amplification Pearson $\uparrow$ & T--I & $-0.0361$ & $+0.0087$ & $+0.0255$ & $-0.0006$ \\*
 & T--C & $+0.0568$ & $-0.0175$ & $-0.0078$ & $+0.0105$ \\*
 & I--C & $+0.0929$ & $-0.0262$ & $-0.0334$ & $+0.0111$ \\
\addlinespace[1pt]
Deletion Pearson $\uparrow$ & T--I & $-0.0543$ & $-0.0093$ & $+0.0348$ & $-0.0096$ \\*
 & T--C & $+0.2175$ & $+0.3369$ & $+0.2461$ & $+0.2668$ \\*
 & I--C & $+0.2717$ & $+0.3462$ & $+0.2112$ & $+0.2764$ \\
\addlinespace[1pt]
Top-50 amplification overlap $\uparrow$ & T--I & $+0.0933$ & $+0.0267$ & $+0.1600$ & $+0.0933$ \\*
 & T--C & $-0.0733$ & $+0.0600$ & $+0.1133$ & $+0.0333$ \\*
 & I--C & $-0.1667$ & $+0.0333$ & $-0.0467$ & $-0.0600$ \\
\addlinespace[1pt]
Top-50 deletion overlap $\uparrow$ & T--I & $+0.0200$ & $-0.2067$ & $-0.0333$ & $-0.0733$ \\*
 & T--C & $+0.3933$ & $+0.0200$ & $+0.1000$ & $+0.1711$ \\*
 & I--C & $+0.3733$ & $+0.2267$ & $+0.1333$ & $+0.2444$ \\
\addlinespace[1pt]
\addlinespace[3pt]
\multicolumn{6}{l}{\textbf{Prespecified 14-biomarker panel}}\\*
Amplification mean absolute error $\downarrow$ & T--I & $+0.0051$ & $-0.0291$ & $-0.0354$ & $-0.0198$ \\*
 & T--C & $-0.0048$ & $-0.0455$ & $-0.0354$ & $-0.0286$ \\*
 & I--C & $-0.0099$ & $-0.0164$ & $0.0000$ & $-0.0088$ \\
\addlinespace[1pt]
Deletion mean absolute error $\downarrow$ & T--I & $+0.0135$ & $-0.0040$ & $-0.0013$ & $+0.0027$ \\*
 & T--C & $-0.0727$ & $-0.0461$ & $-0.0501$ & $-0.0563$ \\*
 & I--C & $-0.0861$ & $-0.0421$ & $-0.0488$ & $-0.0590$ \\
\addlinespace[1pt]
\addlinespace[3pt]
\multicolumn{6}{l}{\textbf{Cis RNA--CNV association}}\\*
Pearson of slopes $\uparrow$ & T--I & $-0.0778$ & $+0.0132$ & $-0.0375$ & $-0.0340$ \\*
 & T--C & $+0.0813$ & $+0.1239$ & $+0.0417$ & $+0.0823$ \\*
 & I--C & $+0.1591$ & $+0.1107$ & $+0.0792$ & $+0.1163$ \\
\addlinespace[1pt]
Median absolute slope error $\downarrow$ & T--I & $+0.0177$ & $+0.0202$ & $+0.0356$ & $+0.0245$ \\*
 & T--C & $-0.0673$ & $-0.0512$ & $-0.0304$ & $-0.0497$ \\*
 & I--C & $-0.0850$ & $-0.0715$ & $-0.0660$ & $-0.0742$ \\
\addlinespace[1pt]
Spearman of slopes $\uparrow$ & T--I & $-0.0386$ & $-0.0045$ & $-0.0600$ & $-0.0343$ \\*
 & T--C & $+0.0908$ & $+0.1061$ & $+0.0404$ & $+0.0791$ \\*
 & I--C & $+0.1293$ & $+0.1105$ & $+0.1004$ & $+0.1134$ \\
\addlinespace[1pt]
Mean absolute slope error $\downarrow$ & T--I & $+0.0248$ & $+0.0113$ & $+0.0213$ & $+0.0191$ \\*
 & T--C & $-0.0608$ & $-0.0406$ & $-0.0377$ & $-0.0464$ \\*
 & I--C & $-0.0856$ & $-0.0519$ & $-0.0590$ & $-0.0655$ \\
\addlinespace[1pt]
Slope sign agreement $\uparrow$ & T--I & $+0.0191$ & $+0.0137$ & $+0.0011$ & $+0.0113$ \\*
 & T--C & $+0.0602$ & $+0.0573$ & $+0.0535$ & $+0.0570$ \\*
 & I--C & $+0.0411$ & $+0.0435$ & $+0.0523$ & $+0.0456$ \\
\addlinespace[1pt]
\addlinespace[3pt]
\multicolumn{6}{l}{\textbf{RNA co-expression}}\\*
Fisher-$z$ concordance $\uparrow$ & T--I & $+0.0116$ & $+0.0469$ & $-0.0096$ & $+0.0163$ \\*
 & T--C & $+0.0450$ & $+0.0093$ & $+0.0108$ & $+0.0217$ \\*
 & I--C & $+0.0334$ & $-0.0376$ & $+0.0204$ & $+0.0054$ \\
\addlinespace[1pt]
Edge F1 $\uparrow$ & T--I & $+0.0079$ & $+0.0071$ & $+0.0100$ & $+0.0083$ \\*
 & T--C & $+0.1148$ & $+0.0989$ & $+0.1064$ & $+0.1067$ \\*
 & I--C & $+0.1069$ & $+0.0917$ & $+0.0964$ & $+0.0983$ \\
\addlinespace[1pt]
Module adjusted Rand index $\uparrow$ & T--I & $+0.0334$ & $+0.0273$ & $+0.0053$ & $+0.0220$ \\*
 & T--C & $+0.0280$ & $+0.0721$ & $+0.0687$ & $+0.0562$ \\*
 & I--C & $-0.0054$ & $+0.0448$ & $+0.0634$ & $+0.0342$ \\
\addlinespace[1pt]
Hub Jaccard $\uparrow$ & T--I & $-0.0027$ & $-0.0100$ & $+0.0373$ & $+0.0082$ \\*
 & T--C & $+0.0663$ & $+0.0270$ & $+0.0614$ & $+0.0516$ \\*
 & I--C & $+0.0689$ & $+0.0370$ & $+0.0241$ & $+0.0434$ \\
\addlinespace[1pt]
Edge precision $\uparrow$ & T--I & $+0.0068$ & $+0.0037$ & $+0.0221$ & $+0.0108$ \\*
 & T--C & $+0.0998$ & $+0.0928$ & $+0.1174$ & $+0.1033$ \\*
 & I--C & $+0.0930$ & $+0.0891$ & $+0.0954$ & $+0.0925$ \\
\addlinespace[1pt]
Edge recall $\uparrow$ & T--I & $+0.0085$ & $+0.0204$ & $-0.0471$ & $-0.0061$ \\*
 & T--C & $-0.1190$ & $-0.1626$ & $-0.1691$ & $-0.1503$ \\*
 & I--C & $-0.1276$ & $-0.1830$ & $-0.1220$ & $-0.1442$ \\
\addlinespace[1pt]
Edge density ratio ($\to 1$) & T--I & $+0.0062$ & $+0.0758$ & $-0.5310$ & $-0.1496$ \\*
 & T--C & $-4.6661$ & $-4.8912$ & $-4.0846$ & $-4.5473$ \\*
 & I--C & $-4.6723$ & $-4.9670$ & $-3.5536$ & $-4.3976$ \\
\addlinespace[1pt]
\addlinespace[3pt]
\multicolumn{6}{l}{\textbf{Training proximity}}\\*
RNA median NN ratio & T--I & $-0.1436$ & $-0.0897$ & $-0.0816$ & $-0.1050$ \\*
 & T--C & $-0.1166$ & $-0.0881$ & $-0.0839$ & $-0.0962$ \\*
 & I--C & $+0.0269$ & $+0.0015$ & $-0.0023$ & $+0.0087$ \\
\addlinespace[1pt]
RNA membership AUC & T--I & $+0.1788$ & $+0.2032$ & $+0.1288$ & $+0.1703$ \\*
 & T--C & $+0.1210$ & $+0.1579$ & $+0.1307$ & $+0.1365$ \\*
 & I--C & $-0.0578$ & $-0.0453$ & $+0.0019$ & $-0.0337$ \\
\addlinespace[1pt]
CNV median NN ratio & T--I & $-0.0333$ & $-0.0096$ & $-0.0310$ & $-0.0246$ \\*
 & T--C & $-0.0114$ & $-0.0047$ & $+0.0025$ & $-0.0045$ \\*
 & I--C & $+0.0219$ & $+0.0049$ & $+0.0335$ & $+0.0201$ \\
\addlinespace[1pt]
CNV membership AUC & T--I & $+0.1531$ & $+0.1475$ & $+0.1101$ & $+0.1369$ \\*
 & T--C & $+0.0543$ & $+0.0569$ & $+0.0363$ & $+0.0492$ \\*
 & I--C & $-0.0988$ & $-0.0906$ & $-0.0738$ & $-0.0877$ \\
\addlinespace[1pt]
Paired median NN ratio & T--I & $-0.0755$ & $-0.0478$ & $-0.0585$ & $-0.0606$ \\*
 & T--C & $-0.0712$ & $-0.0455$ & $-0.0559$ & $-0.0575$ \\*
 & I--C & $+0.0043$ & $+0.0023$ & $+0.0026$ & $+0.0031$ \\
\addlinespace[1pt]
Paired membership AUC & T--I & $+0.2010$ & $+0.2297$ & $+0.1936$ & $+0.2081$ \\*
 & T--C & $+0.1632$ & $+0.1964$ & $+0.1607$ & $+0.1734$ \\*
 & I--C & $-0.0378$ & $-0.0333$ & $-0.0329$ & $-0.0347$ \\
\addlinespace[1pt]
Paired NN agreement & T--I & $+0.0620$ & $+0.1319$ & $+0.1146$ & $+0.1028$ \\*
 & T--C & $+0.1085$ & $+0.1458$ & $+0.1146$ & $+0.1230$ \\*
 & I--C & $+0.0465$ & $+0.0139$ & $0.0000$ & $+0.0201$ \\
\addlinespace[1pt]
\end{longtable}
\endgroup

\clearpage
  \section{After-treatment benchmark and visual diagnostics}
  \label{app:after-treatment-diagnostics}

\begin{figure}[H]
  \centering
  \includegraphics[width=\linewidth]{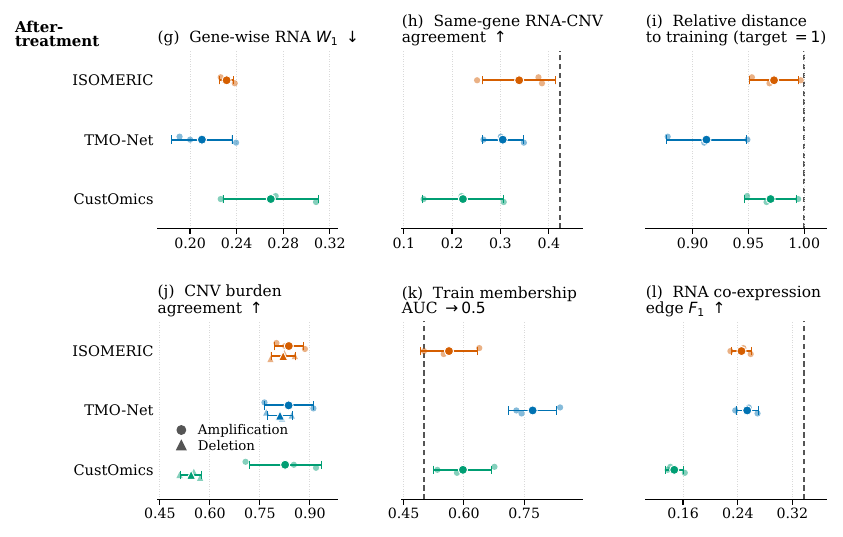}
  \caption{
  \textbf{Held-out evaluation in the after-treatment DECIDER cohort
  (139 paired samples from 89 patients).} \textbf{(g)} Mean gene-wise RNA $W_1$,
  \textbf{(h)} correlation of
  same-gene RNA-CNV slopes, \textbf{(i)} distance to the closest paired
  training sample relative to the closest paired held-out sample,
  \textbf{(j)} amplification and deletion G-score
  correlations, and \textbf{(k)} paired membership AUC. Panel \textbf{(l)}
  reports RNA co-expression edge F1. Arrows
  and dashed lines indicate preferred directions, neutral targets, or real
  train-test references. Small points show individual outer folds, large
  markers and horizontal bars show the mean and sample standard deviation
  across the three folds. TMO-Net has the lowest RNA $W_1$ and the highest
  co-expression edge F1 but also shows the strongest training preference.
  ISOMERIC leads the same-gene and G-score metrics. No architecture dominates
  overall.
  }
  \label{fig:after-treatment-benchmark}
\end{figure}

\clearpage
\begin{figure}[H]
  \centering
  \includegraphics[width=\linewidth,height=0.53\textheight,keepaspectratio]{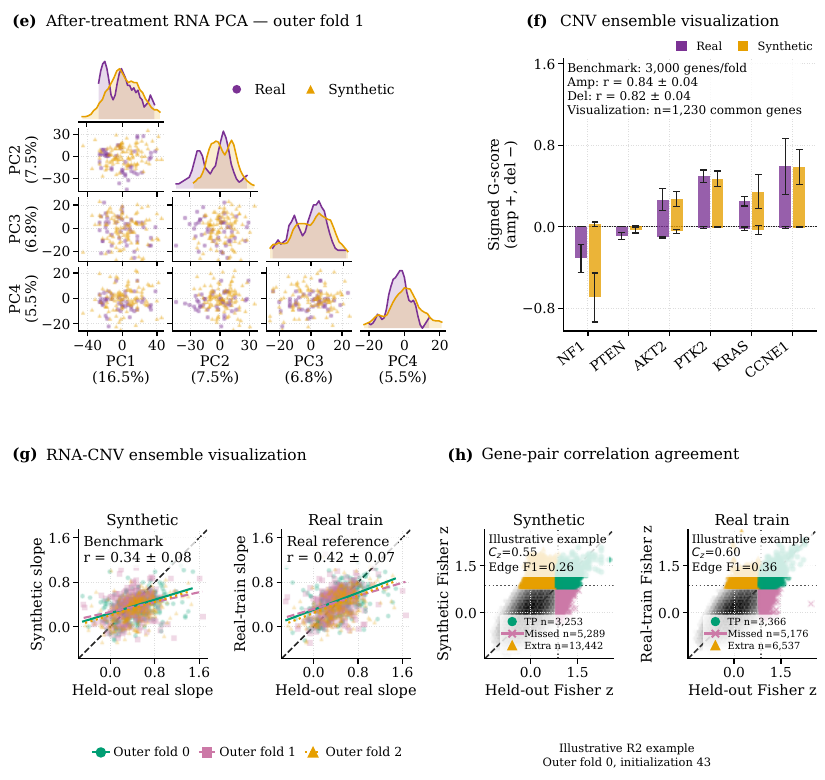}
  \caption{
{  \textbf{After-treatment visual diagnostics.}
  \textbf{(e) RNA fidelity:} Held-out real}
  and ISOMERIC synthetic RNA profiles in a PCA space fitted on real training
{  data.}
  \textbf{(f) CNV fidelity:} Real and ISOMERIC synthetic G-scores for six HGSC
{  genes, deletions are plotted below zero.}
  Bars show mean G-scores across the three folds, error bars indicate sample SD.
{  The CNV visualization uses the intersection of fold-specific evaluated
  genes ($n=1{,}230$), benchmark correlations use all 3,000 CNV genes per fold.}
  \textbf{(g) RNA-CNV association:} For each gene, the linear association
  estimated in the held-out data is compared with the corresponding estimate
  from ISOMERIC synthetic and real-training data. Points near the diagonal
{  indicate agreement. Panels \textbf{(f,g)} are ensemble visualizations:
  synthetic gene-level values are averaged across initializations within each
  fold. Their benchmark annotations use correlations computed separately per
  initialization, averaged within fold, then summarized as mean $\pm$ sample
  SD across three folds. MAE denotes mean absolute error between synthetic
  and held-out G-scores across the 14 prespecified biomarkers. Amplification
  and deletion MAEs are reported in Table~\ref{tab:biomarker-fidelity}.}
  \textbf{(h) RNA co-expression:} The two subpanels compare
  Fisher-transformed gene-pair correlations in held-out real data (x-axis)
  with those in TMO-Net synthetic data and real training data (y-axes), using
{  identical scales. Diagonals mark equality, dotted lines mark $r=0.70$.
  Grayscale hexbins show gene-pair density, colored markers show recovered
  (TP), missed (FN), and extra (FP) edges, with counts in the legends.
  Panel \textbf{(h)} is an illustrative example from one fold and initialization,
  its Fisher-$z$ concordance ($C_z$) and edge F1 are example statistics.}
  Models were selected separately for each property, so the panels are
  illustrative rather than an overall model ranking.
  The CNV criterion is the mean of amplification and deletion G-score correlations.
  Examples in \textbf{(e,h)} use the median fold and the initialization closest
  to its mean score (Appendix~\ref{app:model-visual-diagnostics}).
  }
  \label{fig:best-visual-diagnostics-after}
\end{figure}

\endgroup

\clearpage
  \section{Before-treatment visual diagnostics by model}
  \label{app:model-visual-diagnostics}

\begin{figure}[H]
  \centering
  \includegraphics[width=\linewidth]
  {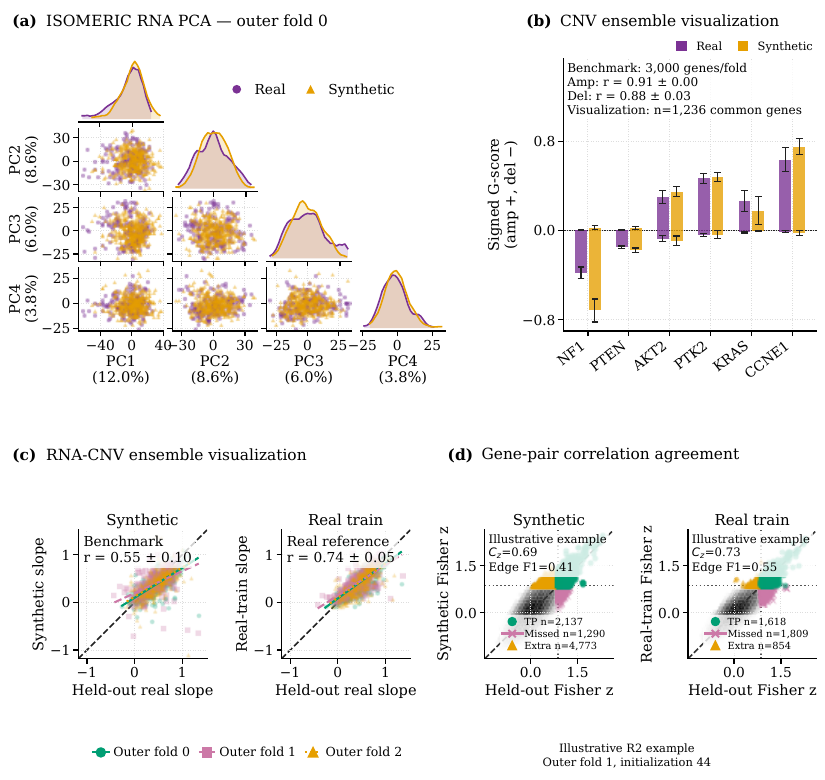}
  \caption{
  \textbf{ISOMERIC: before-treatment visual diagnostics.}
  \textbf{(a)} RNA PCA,
  \textbf{(b)} CNV G-scores, \textbf{(c)} same-gene RNA-CNV slopes, and
{  \textbf{(d)} RNA co-expression. Panels \textbf{(b,c)} are ensemble
  visualizations, averaging synthetic gene-level values across initializations
  within each fold. The CNV plot uses 1,236 genes common to folds,
  benchmark correlations use all 3,000 CNV genes per fold. MAE denotes mean
  absolute error between synthetic and held-out G-scores across the 14
  prespecified biomarkers. Amplification and deletion MAEs are reported in
  Table~\ref{tab:biomarker-fidelity}. Benchmark correlations are computed per
  initialization, averaged within fold, then summarized as mean $\pm$ sample
  SD across three folds.
  Panel \textbf{(d)} shows one
  illustrative fold and initialization, Fisher-$z$ concordance ($C_z$) and edge F1 describe that example.
  Other plotting conventions follow Figure~\ref{fig:best-visual-diagnostics}.}
  }
  \label{fig:visual-diagnostics-isomeric}
\end{figure}

\begin{figure}[H]
  \centering
  \includegraphics[width=\linewidth]
  {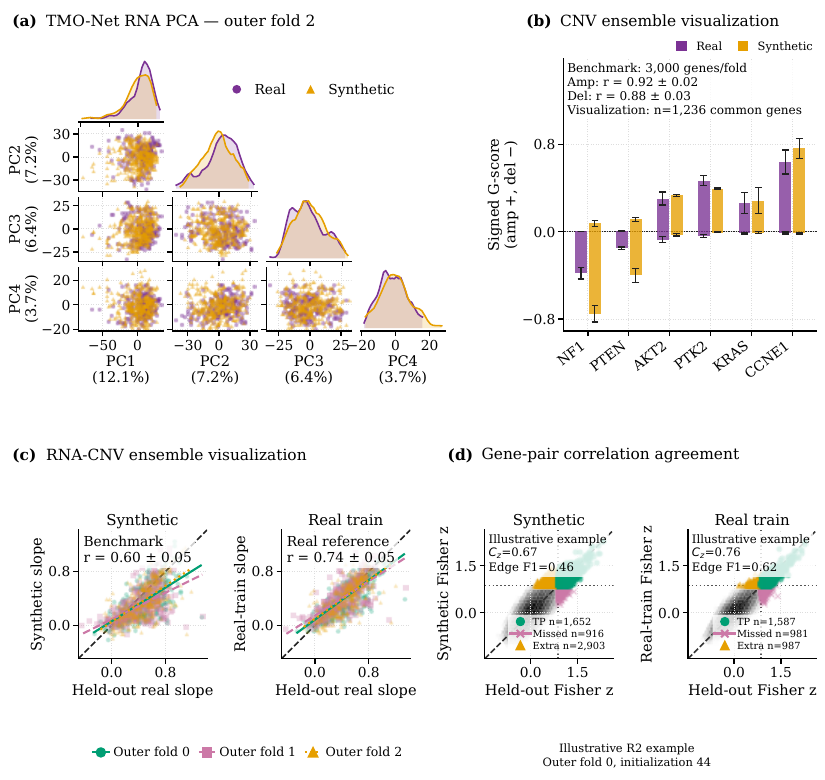}
  \caption{
  \textbf{TMO-Net: before-treatment visual diagnostics.}
  \textbf{(a)} RNA PCA,
  \textbf{(b)} CNV G-scores, \textbf{(c)} same-gene RNA-CNV slopes, and
{  \textbf{(d)} RNA co-expression. Panels \textbf{(b,c)} are ensemble
  visualizations, averaging synthetic gene-level values across initializations
  within each fold. The CNV plot uses 1,236 genes common to folds,
  benchmark correlations use all 3,000 CNV genes per fold. MAE denotes mean
  absolute error between synthetic and held-out G-scores across the 14
  prespecified biomarkers. Amplification and deletion MAEs are reported in
  Table~\ref{tab:biomarker-fidelity}. Benchmark correlations are computed per
  initialization, averaged within fold, then summarized as mean $\pm$ sample
  SD across three folds.
  Panel \textbf{(d)} shows one
  illustrative fold and initialization, Fisher-$z$ concordance ($C_z$) and edge F1 describe that example.
  Other plotting conventions follow Figure~\ref{fig:best-visual-diagnostics}.}
  }
  \label{fig:visual-diagnostics-tmonet}
\end{figure}

\begin{figure}[H]
  \centering
  \includegraphics[width=\linewidth]
  {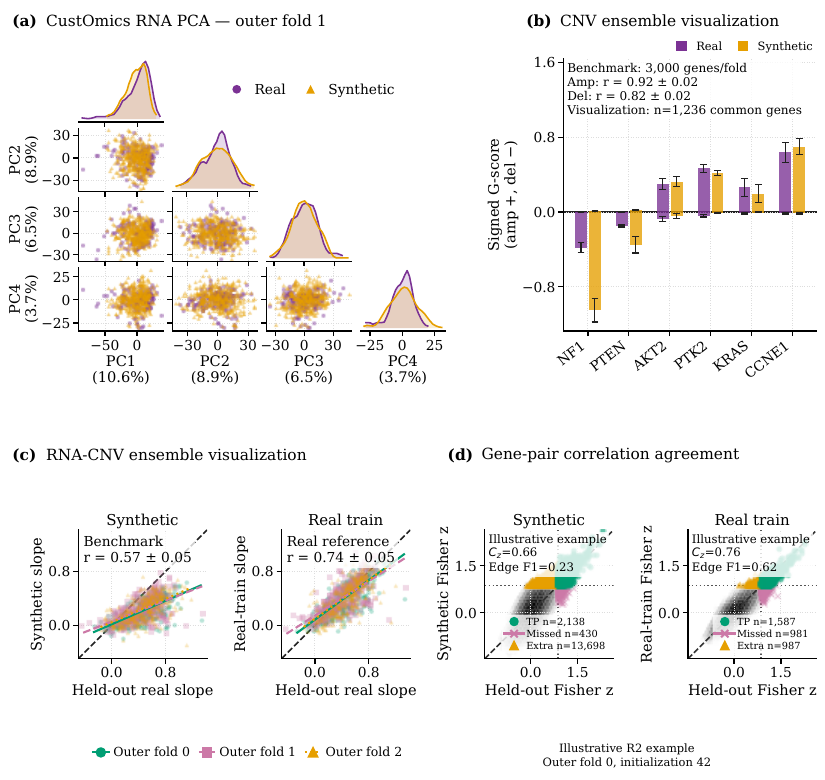}
  \caption{
  \textbf{CustOmics: before-treatment visual diagnostics.}
  \textbf{(a)} RNA PCA,
  \textbf{(b)} CNV G-scores, \textbf{(c)} same-gene RNA-CNV slopes, and
{  \textbf{(d)} RNA co-expression. Panels \textbf{(b,c)} are ensemble
  visualizations, averaging synthetic gene-level values across initializations
  within each fold. The CNV plot uses 1,236 genes common to folds,
  benchmark correlations use all 3,000 CNV genes per fold. MAE denotes mean
  absolute error between synthetic and held-out G-scores across the 14
  prespecified biomarkers. Amplification and deletion MAEs are reported in
  Table~\ref{tab:biomarker-fidelity}. Benchmark correlations are computed per
  initialization, averaged within fold, then summarized as mean $\pm$ sample
  SD across three folds.
  Panel \textbf{(d)} shows one
  illustrative fold and initialization, Fisher-$z$ concordance ($C_z$) and edge F1 describe that example.
  Other plotting conventions follow Figure~\ref{fig:best-visual-diagnostics}.}
  }
  \label{fig:visual-diagnostics-customics}
\end{figure}

  Figures~\ref{fig:visual-diagnostics-isomeric}--\ref{fig:visual-diagnostics-customics}
{  show diagnostics for each model. PCA and co-expression examples select
  the median outer fold by initialization-averaged mean PC Wasserstein distance
  and edge F1, respectively, then the initialization closest to that fold's
  mean score.}

\clearpage

\clearpage

\clearpage
  \section{After-treatment visual diagnostics by model}
  \label{app:model-visual-diagnostics-after}

\begin{figure}[H]
  \centering
  \includegraphics[width=\linewidth]
  {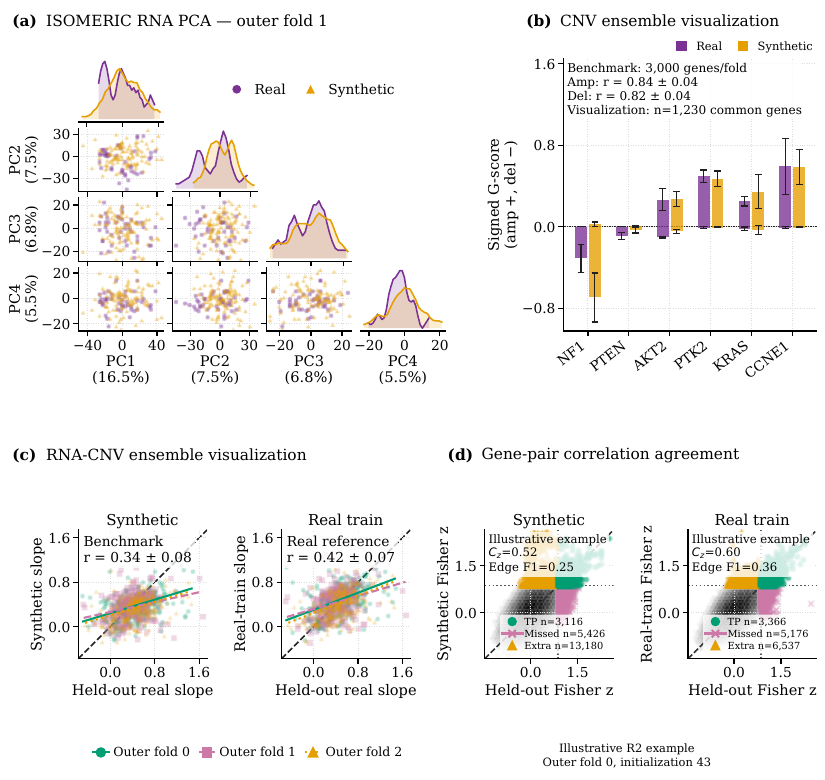}
  \caption{
  \textbf{ISOMERIC: after-treatment visual diagnostics.}
  \textbf{(a)} RNA PCA,
  \textbf{(b)} CNV G-scores, \textbf{(c)} same-gene RNA-CNV slopes, and
{  \textbf{(d)} RNA co-expression. Panels \textbf{(b,c)} are ensemble
  visualizations, averaging synthetic gene-level values across initializations
  within each fold. The CNV plot uses 1,230 genes common to folds,
  benchmark correlations use all 3,000 CNV genes per fold. MAE denotes mean
  absolute error between synthetic and held-out G-scores across the 14
  prespecified biomarkers. Amplification and deletion MAEs are reported in
  Table~\ref{tab:biomarker-fidelity}. Benchmark correlations are computed per
  initialization, averaged within fold, then summarized as mean $\pm$ sample
  SD across three folds.
  Panel \textbf{(d)} shows one
  illustrative fold and initialization, Fisher-$z$ concordance ($C_z$) and edge F1 describe that example.
  Other plotting conventions follow Figure~\ref{fig:best-visual-diagnostics-after}.}
  }
  \label{fig:visual-diagnostics-isomeric-after}
\end{figure}

\begin{figure}[H]
  \centering
  \includegraphics[width=\linewidth]
  {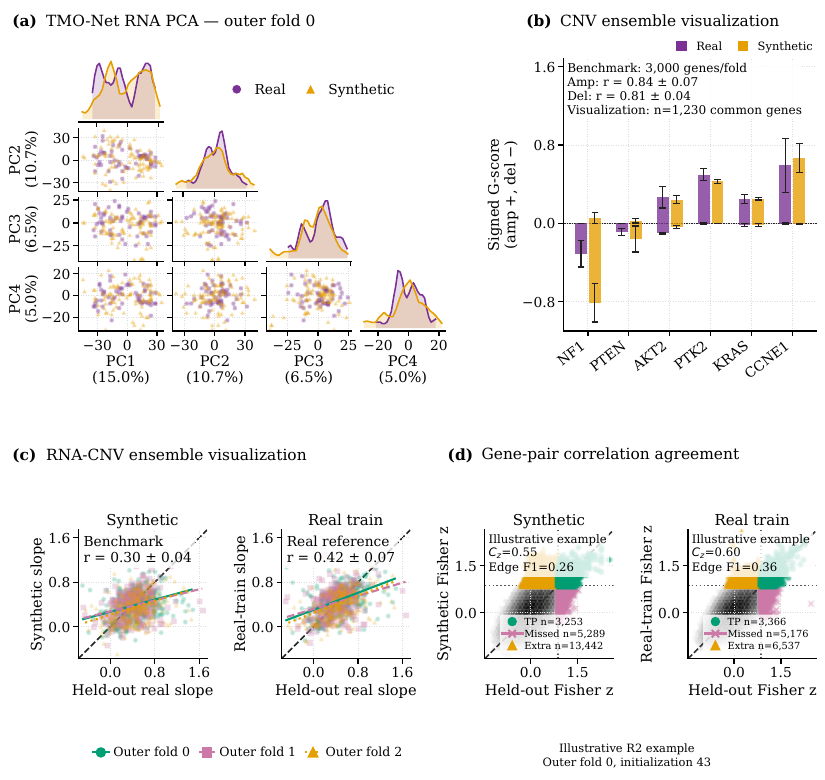}
  \caption{
  \textbf{TMO-Net: after-treatment visual diagnostics.}
  \textbf{(a)} RNA PCA,
  \textbf{(b)} CNV G-scores, \textbf{(c)} same-gene RNA-CNV slopes, and
{  \textbf{(d)} RNA co-expression. Panels \textbf{(b,c)} are ensemble
  visualizations, averaging synthetic gene-level values across initializations
  within each fold. The CNV plot uses 1,230 genes common to folds,
  benchmark correlations use all 3,000 CNV genes per fold. MAE denotes mean
  absolute error between synthetic and held-out G-scores across the 14
  prespecified biomarkers. Amplification and deletion MAEs are reported in
  Table~\ref{tab:biomarker-fidelity}. Benchmark correlations are computed per
  initialization, averaged within fold, then summarized as mean $\pm$ sample
  SD across three folds.
  Panel \textbf{(d)} shows one
  illustrative fold and initialization, Fisher-$z$ concordance ($C_z$) and edge F1 describe that example.
  Other plotting conventions follow Figure~\ref{fig:best-visual-diagnostics-after}.}
  }
  \label{fig:visual-diagnostics-tmonet-after}
\end{figure}

\begin{figure}[H]
  \centering
  \includegraphics[width=\linewidth]
  {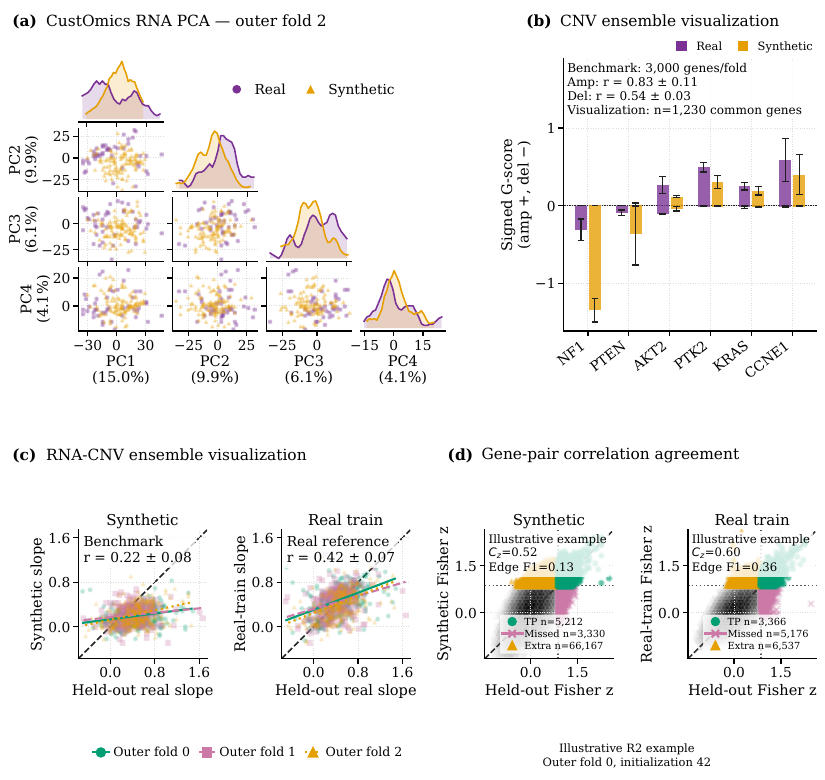}
  \caption{
  \textbf{CustOmics: after-treatment visual diagnostics.}
  \textbf{(a)} RNA PCA,
  \textbf{(b)} CNV G-scores, \textbf{(c)} same-gene RNA-CNV slopes, and
{  \textbf{(d)} RNA co-expression. Panels \textbf{(b,c)} are ensemble
  visualizations, averaging synthetic gene-level values across initializations
  within each fold. The CNV plot uses 1,230 genes common to folds,
  benchmark correlations use all 3,000 CNV genes per fold. MAE denotes mean
  absolute error between synthetic and held-out G-scores across the 14
  prespecified biomarkers. Amplification and deletion MAEs are reported in
  Table~\ref{tab:biomarker-fidelity}. Benchmark correlations are computed per
  initialization, averaged within fold, then summarized as mean $\pm$ sample
  SD across three folds.
  Panel \textbf{(d)} shows one
  illustrative fold and initialization, Fisher-$z$ concordance ($C_z$) and edge F1 describe that example.
  Other plotting conventions follow Figure~\ref{fig:best-visual-diagnostics-after}.}
  }
  \label{fig:visual-diagnostics-customics-after}
\end{figure}

  Figures~\ref{fig:visual-diagnostics-isomeric-after}--\ref{fig:visual-diagnostics-customics-after}
  show the four diagnostics for each benchmark model in the after-treatment
  DECIDER cohort (139 paired samples from 89 patients). Fold and initialization
  selection and within-fold averaging follow
  Appendix~\ref{app:model-visual-diagnostics}.

\clearpage

\clearpage

\clearpage
\begingroup
\raggedbottom
\section{Stochastic decoding: RNA variance and co-expression}
\label{app:stochastic-decoder}

\paragraph{Question.}
Does omitting variation around RNA decoder means contribute to the variance
deficit and excessive co-expression observed in Section~\ref{sec:generation-experiments}?
We examine this question by adding independent Gaussian residual noise to the
same generated RNA profiles and assessing variance, network density, and edge
preservation together.

\paragraph{Matched diagnostic intervention.}
We use the before-treatment cohort, three patient-disjoint outer folds, and
one training initialization (seed 42) for each architecture, giving nine fitted
models. The fitted weights, selected sampling temperatures, latent draws,
synthetic cohort sizes, and generated CNV profiles remain fixed. Only the RNA
decoder output is perturbed, no model is retrained or sampling setting retuned.

For each fitted model, we reconstruct its outer-training RNA in the saved,
model-specific scaled space, using latent means. ISOMERIC infers both latent
means from RNA. TMO-Net maps observed training CNV through its CNV-to-RNA
encoder and decodes the conditional latent mean, neither its RNA self-encoder
nor a preceding CNV reconstruction is used. CustOmics encodes paired RNA and
CNV, passes their concatenated representations through the central latent mean
and central decoder, then applies the RNA modality decoder.
Writing $e_{ig}=r_{ig}-\widehat r_{ig}$ for these scaled training residuals,
we compute
\[
\begin{aligned}
\bar e_g&=\frac{1}{n}\sum_{i=1}^{n}e_{ig},\qquad
s_g^2=\frac{1}{n-1}\sum_{i=1}^{n}(e_{ig}-\bar e_g)^2,\\
\widetilde r^{\mathrm{G}}&=\widetilde r^{\mathrm{D}}+\eta,
\qquad \eta\sim\mathcal N\!\left(0,\operatorname{diag}(s_1^2,\ldots,s_{d_R}^2)\right).
\end{aligned}
\]
Here D denotes decoder means and G the Gaussian condition. Independent noise
vectors are drawn for each synthetic sample, with unit noise amplitude, before
applying the original inverse scaling and export procedure. Calibration uses
outer-training observations only. Centering removes the estimated residual
bias from calibration, the generated noise is zero-mean in expectation.

Both conditions are evaluated with the same R1--R4 procedures, including
co-expression thresholds and reference subsampling. We report arithmetic
means $\pm$ sample standard deviations across the three folds. For each
metric, $\Delta=\mathrm{G}-\mathrm{D}$ is computed within fold before
summarizing, its standard deviation describes paired fold differences.

\paragraph{Results.}
Adding noise increases the mean RNA variance ratio from $0.667$ to $0.714$
for TMO-Net, $0.568$ to $0.844$ for ISOMERIC, and $0.600$ to $1.058$ for
CustOmics (Figure~\ref{fig:stochastic-decoder-ablation},
Table~\ref{tab:stochastic-decoder-primary}). TMO-Net's density ratio approaches
one ($1.681$ to $1.051$), with edge F1 increasing from $0.499$ to $0.519$.
ISOMERIC and CustOmics instead become overly sparse: density ratios fall to
$0.190$ and $0.049$, while edge F1 falls from $0.432$ to $0.193$ and from
$0.226$ to $0.076$, respectively. Gene-wise Wasserstein distances decrease for
all three models, but Fisher-$z$ concordance and cis-slope Pearson correlations
also decrease (Table~\ref{tab:stochastic-decoder-secondary}). RNA membership
AUCs remain above $0.5$ (Table~\ref{tab:stochastic-decoder-proximity}).

\paragraph{Interpretation and scope.}
Independent noise can reduce correlations mechanically. These results support
a contribution of omitted residual variation to variance compression, but
restoring variance alone does not recover co-expression structure. Training
residuals combine model error and unexplained biological variation and do not
identify observation noise. For TMO-Net, some of this variation may already
be represented by its conditional latent, so the perturbation can double-count
it. This is a diagnostic intervention, not a learned observation model or an
established improvement in generation. The single initialization provides no
estimate of between-initialization variability, and proximity changes do not
establish privacy protection.

\begin{figure}[H]
  \centering
  \includegraphics[width=\linewidth]{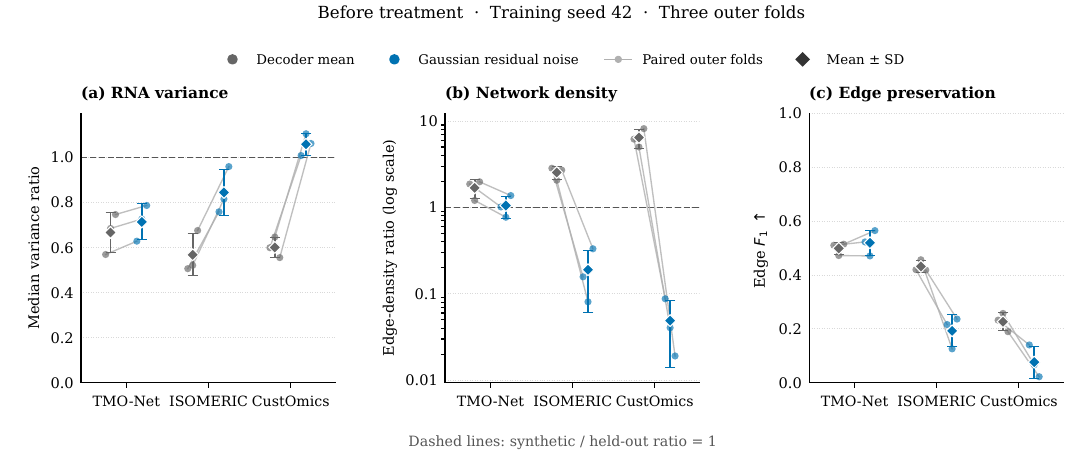}
  \caption{\textbf{RNA residual-noise ablation before treatment.}
  Decoder means (gray) and Gaussian residual noise (blue) share fitted models,
  sampling temperatures, latent draws, and CNV profiles.
  \textbf{(a)} Median gene-wise RNA variance ratio,
  \textbf{(b)} co-expression edge-density ratio (logarithmic axis),
  \textbf{(c)} edge F1. Ratios compare synthetic with held-out data and target
  one (dashed lines), higher F1 indicates greater edge agreement.
  Connected points show paired outer folds, diamonds and bars show arithmetic
  means $\pm$ sample SD across three folds, using training seed 42 only.}
  \label{fig:stochastic-decoder-ablation}
\end{figure}

\begin{table}[H]\centering
\begingroup\footnotesize
\setlength{\tabcolsep}{1.8pt}
\renewcommand{\arraystretch}{1.25}
\begin{tabular}{@{}llccc@{}}
\toprule
Model & Condition & \shortstack{RNA variance\\ratio $\to1$} & \shortstack{Edge-density\\ratio $\to1$} & Edge $F_1\uparrow$ \\
\midrule
TMO-Net & Decoder mean & $0.667 \pm 0.090$ & $1.681 \pm 0.417$ & $0.499 \pm 0.023$ \\
 & Gaussian & $0.714 \pm 0.080$ & $1.051 \pm 0.302$ & $0.519 \pm 0.047$ \\
 & $\Delta$ (G $-$ D) & $+0.047 \pm 0.011$ & $-0.630 \pm 0.209$ & $+0.021 \pm 0.027$ \\
\addlinespace[4pt]
ISOMERIC & Decoder mean & $0.568 \pm 0.094$ & $2.535 \pm 0.413$ & $0.432 \pm 0.022$ \\
 & Gaussian & $0.844 \pm 0.103$ & $0.190 \pm 0.129$ & $0.193 \pm 0.059$ \\
 & $\Delta$ (G $-$ D) & $+0.277 \pm 0.022$ & $-2.345 \pm 0.350$ & $-0.239 \pm 0.080$ \\
\addlinespace[4pt]
CustOmics & Decoder mean & $0.600 \pm 0.046$ & $6.441 \pm 1.595$ & $0.226 \pm 0.034$ \\
 & Gaussian & $1.058 \pm 0.048$ & $0.049 \pm 0.035$ & $0.076 \pm 0.060$ \\
 & $\Delta$ (G $-$ D) & $+0.458 \pm 0.049$ & $-6.392 \pm 1.611$ & $-0.150 \pm 0.052$ \\
\bottomrule
\end{tabular}\endgroup
\caption{\textbf{Variance and co-expression under RNA residual noise.} Before-treatment cohort, training seed 42, three patient-disjoint outer folds. Values are means $\pm$ sample SD across folds. D denotes decoder means and G Gaussian residual noise, $\Delta=\mathrm{G}-\mathrm{D}$ is computed within fold before summarizing. Both synthetic-to-held-out ratios target one, higher edge F1 indicates greater agreement. Weights, temperatures, latent draws, and CNV are fixed.}
\label{tab:stochastic-decoder-primary}
\end{table}

\begin{table}[H]\centering
\begingroup\footnotesize
\setlength{\tabcolsep}{1.8pt}
\renewcommand{\arraystretch}{1.25}
\resizebox{\linewidth}{!}{\begin{tabular}{@{}llccccc@{}}
\toprule
Model & Condition & $W_1\downarrow$ & $\mathrm{MMD}^2\downarrow$ & \shortstack{Fisher-$z$\\Pearson $\uparrow$} & $r_\beta\uparrow$ & \shortstack{Median\\$|\Delta\beta|\downarrow$} \\
\midrule
TMO-Net & Decoder mean & $0.1620 \pm 0.0180$ & $0.01313 \pm 0.00243$ & $0.6925 \pm 0.0341$ & $0.5950 \pm 0.0577$ & $0.1463 \pm 0.0087$ \\
 & Gaussian & $0.1566 \pm 0.0145$ & $0.01310 \pm 0.00240$ & $0.6861 \pm 0.0349$ & $0.5922 \pm 0.0552$ & $0.1479 \pm 0.0134$ \\
 & $\Delta$ (G $-$ D) & $-0.0054 \pm 0.0035$ & $-0.00003 \pm 0.00004$ & $-0.0064 \pm 0.0018$ & $-0.0028 \pm 0.0047$ & $+0.0017 \pm 0.0060$ \\
\addlinespace[4pt]
ISOMERIC & Decoder mean & $0.1846 \pm 0.0178$ & $0.01370 \pm 0.00370$ & $0.6965 \pm 0.0188$ & $0.5463 \pm 0.1238$ & $0.1204 \pm 0.0070$ \\
 & Gaussian & $0.1545 \pm 0.0110$ & $0.01369 \pm 0.00378$ & $0.6593 \pm 0.0139$ & $0.5386 \pm 0.1232$ & $0.1229 \pm 0.0002$ \\
 & $\Delta$ (G $-$ D) & $-0.0301 \pm 0.0167$ & $-0.00002 \pm 0.00010$ & $-0.0373 \pm 0.0092$ & $-0.0077 \pm 0.0207$ & $+0.0025 \pm 0.0069$ \\
\addlinespace[4pt]
CustOmics & Decoder mean & $0.1864 \pm 0.0153$ & $0.02085 \pm 0.01004$ & $0.6704 \pm 0.0088$ & $0.5578 \pm 0.0388$ & $0.2229 \pm 0.0208$ \\
 & Gaussian & $0.1693 \pm 0.0080$ & $0.02059 \pm 0.00991$ & $0.6297 \pm 0.0114$ & $0.5421 \pm 0.0277$ & $0.2210 \pm 0.0210$ \\
 & $\Delta$ (G $-$ D) & $-0.0171 \pm 0.0087$ & $-0.00026 \pm 0.00015$ & $-0.0408 \pm 0.0052$ & $-0.0157 \pm 0.0184$ & $-0.0019 \pm 0.0023$ \\
\bottomrule
\end{tabular}}\endgroup
\caption{\textbf{RNA fidelity and cis RNA--CNV association under residual noise.} Before-treatment cohort, seed 42, means $\pm$ sample SD across three patient-disjoint outer folds. $\Delta$ summarizes paired Gaussian-minus-decoder-mean differences. $W_1$: mean gene-wise Wasserstein distance, MMD$^2$: squared maximum mean discrepancy in PCA space, Fisher-$z$ Pearson: RNA correlation concordance, $r_\beta$: slope Pearson correlation. Arrows indicate preferred directions against held-out data.}
\label{tab:stochastic-decoder-secondary}
\end{table}

\begin{table}[H]\centering
\begingroup\footnotesize
\setlength{\tabcolsep}{1.8pt}
\renewcommand{\arraystretch}{1.25}
\resizebox{\linewidth}{!}{\begin{tabular}{@{}llccccc@{}}
\toprule
Model & Condition & \shortstack{RNA\\NN ratio} & \shortstack{RNA\\AUC} & \shortstack{Paired\\NN ratio} & \shortstack{Paired\\AUC} & \shortstack{Paired NN\\agreement} \\
\midrule
TMO-Net & Decoder mean & $0.8855 \pm 0.0200$ & $0.8142 \pm 0.0321$ & $0.9552 \pm 0.0137$ & $0.6691 \pm 0.0419$ & $0.0998 \pm 0.0179$ \\
 & Gaussian & $0.8927 \pm 0.0172$ & $0.8136 \pm 0.0325$ & $0.9556 \pm 0.0139$ & $0.6687 \pm 0.0422$ & $0.0987 \pm 0.0145$ \\
 & $\Delta$ (G $-$ D) & $+0.0072 \pm 0.0028$ & $-0.0006 \pm 0.0007$ & $+0.0003 \pm 0.0003$ & $-0.0004 \pm 0.0005$ & $-0.0011 \pm 0.0035$ \\
\addlinespace[4pt]
ISOMERIC & Decoder mean & $0.9498 \pm 0.0117$ & $0.6486 \pm 0.0527$ & $0.9932 \pm 0.0142$ & $0.4422 \pm 0.0312$ & $0.0301 \pm 0.0098$ \\
 & Gaussian & $0.9627 \pm 0.0091$ & $0.6473 \pm 0.0529$ & $0.9946 \pm 0.0127$ & $0.4422 \pm 0.0305$ & $0.0332 \pm 0.0097$ \\
 & $\Delta$ (G $-$ D) & $+0.0129 \pm 0.0052$ & $-0.0013 \pm 0.0052$ & $+0.0014 \pm 0.0015$ & $+0.0000 \pm 0.0011$ & $+0.0030 \pm 0.0001$ \\
\addlinespace[4pt]
CustOmics & Decoder mean & $0.9389 \pm 0.0080$ & $0.6930 \pm 0.0481$ & $0.9858 \pm 0.0156$ & $0.5221 \pm 0.0537$ & $0.0181 \pm 0.0074$ \\
 & Gaussian & $0.9611 \pm 0.0067$ & $0.6865 \pm 0.0503$ & $0.9875 \pm 0.0142$ & $0.5229 \pm 0.0526$ & $0.0203 \pm 0.0074$ \\
 & $\Delta$ (G $-$ D) & $+0.0222 \pm 0.0023$ & $-0.0064 \pm 0.0026$ & $+0.0017 \pm 0.0014$ & $+0.0008 \pm 0.0015$ & $+0.0021 \pm 0.0037$ \\
\bottomrule
\end{tabular}}\endgroup
\caption{\textbf{Training proximity under RNA residual noise.} Before-treatment cohort, seed 42, means $\pm$ sample SD across three patient-disjoint outer folds. $\Delta$ summarizes paired Gaussian-minus-decoder-mean differences. Nearest-neighbor (NN) ratios compare distances from synthetic profiles to equally sized training and held-out sets. Ratios below one or membership AUC above $0.5$ indicate training preference. Paired NN agreement is the fraction whose closest training RNA and CNV profiles identify the same sample. These diagnostics do not establish privacy protection.}
\label{tab:stochastic-decoder-proximity}
\end{table}

\clearpage
\section{Controls for pairing sensitivity and independent empirical marginals}
\label{app:simple-controls}

These controls test whether agreement within each modality can coexist with
incorrect RNA--CNV pairing. We first disrupt real held-out pairs to assess the
sensitivity of the cis metrics, then evaluate a baseline that recombines
complete RNA and CNV profiles from different training patients.

\subsection{Pair-shuffled held-out control}
\label{app:pair-shuffled-control}

\paragraph{Protocol.}
Within each held-out fold, we keep RNA profiles fixed and randomly reassign
complete CNV profiles between different patients from the same tissue and
clinical treatment phase. Each CNV profile is used once per permutation.
This changes the pairing while preserving the individual RNA and CNV profiles.
Groups containing only one patient are excluded from both conditions.
The intact and shuffled comparisons use the same remaining samples and
estimable genes from the original gene panels.

We repeat the permutation 100 times in each of the three folds, separately
before and after treatment. For each cohort and fold, we recompute same-gene
RNA--CNV slopes using all eligible samples. Each slope describes how a gene's
RNA expression varies with its CNV across samples. Slopes from the original
real-training pairs provide the reference for both intact and shuffled
held-out slopes. The training pairs remain unchanged. No generator is trained.

\paragraph{Results and interpretation.}
Disrupting pairing reduces mean slope correlation from $0.742$ to $0.013$
before treatment and from $0.410$ to $0.005$ after treatment
(Table~\ref{tab:pair-shuffled-summary},
Figure~\ref{fig:pair-shuffled-correlation}). Absolute slope error increases and
sign agreement falls. Summaries average permutations within each fold, then
report mean $\pm$ sample standard deviation (SD) across the three folds.
Checks of RNA distributions, CNV G-scores, RNA correlations, and co-expression
edges confirm that these properties remain unchanged after shuffling.
Module and hub metrics were not recomputed.

The cis metrics therefore detect errors missed by these marginal and RNA-only
network measures. This is a descriptive intervention, not a formal patient-level
permutation test. Repeated samples and tissue or phase associations remain
relevant, so pooled shuffled slopes need not be exactly zero.

\begin{table}[htbp]
  \centering
  \small
  \setlength{\tabcolsep}{4pt}
  \renewcommand{\arraystretch}{1.12}
  \begin{tabular}{@{}llccc@{}}
    \toprule
    Cohort & Pairing & $r_\beta\uparrow$ &
      $\operatorname{median}|\Delta\beta|\downarrow$ & Sign agreement $\uparrow$ \\
    \midrule
    Before treatment & Intact & $0.742\pm0.046$ & $0.093\pm0.006$ & $0.958\pm0.008$ \\
     & Shuffled & $0.013\pm0.029$ & $0.436\pm0.021$ & $0.480\pm0.007$ \\
    \addlinespace
    After treatment & Intact & $0.410\pm0.061$ & $0.163\pm0.017$ & $0.922\pm0.022$ \\
     & Shuffled & $0.005\pm0.065$ & $0.472\pm0.037$ & $0.464\pm0.021$ \\
    \bottomrule
  \end{tabular}
  \caption{\textbf{Cis association scores for intact and shuffled held-out
  pairs against real-training slopes.} Values are means $\pm$ sample standard
  deviations across three outer folds. For shuffled pairs, the 100 permutation
  scores are first averaged within each fold. $r_\beta$ is the Pearson
  correlation of same-gene slopes. Sign agreement excludes exact-zero slopes.
  Intact and shuffled comparisons use the same eligible held-out samples and
  fixed gene set in each fold. Arrows indicate closer agreement with the
  reference, the standard deviations are descriptive, not confidence intervals.}
  \label{tab:pair-shuffled-summary}
\end{table}

\begin{figure}[p]
  \centering
  \includegraphics[width=0.70\linewidth,trim=20bp 250bp 620bp 62bp,clip]
    {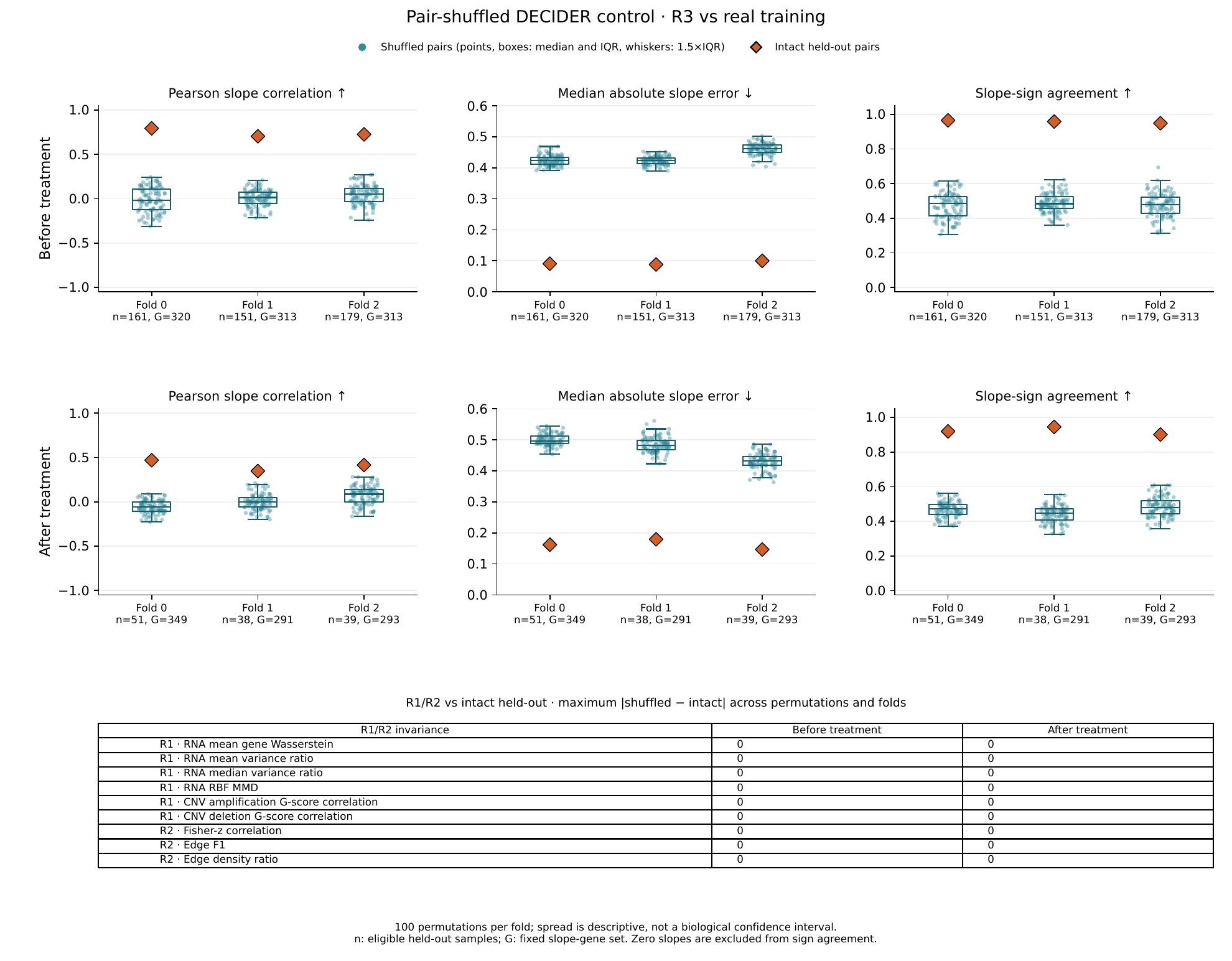}
  \caption{\textbf{Sensitivity of cis RNA--CNV slope correlation to pair
  shuffling.} The upper and lower panels show the before- and after-treatment
  cohorts. Orange diamonds show correlations between slopes from intact
  held-out pairs and slopes from the original real-training pairs. Teal points
  show the corresponding correlations after each of 100 CNV permutations in
  the held-out data. The training pairs remain unchanged in every comparison.
  Boxes show the median and
  interquartile range, whiskers extend to the most extreme values within
  1.5 interquartile ranges of the box. Labels give the number of eligible
  held-out samples ($n$) and the fixed number of estimable genes ($G$).
  Folds are displayed separately. The spread describes variation among
  permutations and is not a confidence interval.}
  \label{fig:pair-shuffled-correlation}
\end{figure}

\clearpage
\subsection{Independent empirical marginal baseline}
\label{app:independent-marginals-control}

\paragraph{Protocol.}
We construct each synthetic pair from a complete RNA profile and a complete
CNV profile drawn from the training data. Donors are initially sampled independently
within the same fold and tissue group, with replacement. We redraw both donors
if they belong to the same patient. No generator is trained.

Before- and after-treatment cohorts are sampled separately. Within the
after-treatment cohort, interval and relapse samples share the same donor
pool within each tissue group. Generated cohorts match the training sample
count and tissue composition, using the benchmark preprocessing and gene
panels. Individual profiles may be reused, so their frequencies can differ
from those in the training data.

We generate three replicate cohorts per fold and evaluate each with the
benchmark R1--R4 metrics. R1--R3 compare generated cohorts with intact held-out
data. R4 measures their proximity to training versus held-out samples.
Metric values are averaged over replicates within
each fold, then reported as mean $\pm$ sample SD across three folds
(Table~\ref{tab:independent-marginals-summary}). Unlike the adapted VAEs, this
baseline explicitly matches tissue composition, so the comparison does not
isolate model architecture.

\paragraph{Results and interpretation.}
Independent marginals has lower mean RNA Wasserstein distance and higher mean
RNA edge F1 than the three learned generators in both cohorts
(Figure~\ref{fig:independent-marginals-comparison}). These favorable RNA-only
scores coexist with near-zero cis slope correlation: $0.024\pm0.031$ before
treatment and $-0.022\pm0.083$ after treatment. The learned generators retain
higher correlations, supporting separate evaluation of paired dependence.

Paired membership AUC is high: $0.876\pm0.021$ before treatment and
$0.767\pm0.054$ after treatment, the latter similar to TMO-Net's
$0.771\pm0.059$. Individual profiles are copied from training donors by
construction. Favorable marginal scores do not establish novelty or privacy.
Balanced reference subsampling can omit a copied donor, so not every
nearest-training distance must be zero. These descriptive comparisons separate
marginal fidelity, paired association, and training proximity without claiming
significance from three fold summaries.

\begin{table}[htbp]
  \centering
  \small
  \setlength{\tabcolsep}{4pt}
  \renewcommand{\arraystretch}{1.12}
  \begin{tabular}{@{}lcccc@{}}
    \toprule
    Method & RNA $W_1\downarrow$ & Edge F1 $\uparrow$ &
      $r_\beta\uparrow$ & Paired AUC \\
    \midrule
    \multicolumn{5}{@{}l}{\textit{Before treatment}} \\
    ISOMERIC & $0.184\pm0.020$ & $0.412\pm0.018$ & $0.551\pm0.101$ & $0.446\pm0.035$ \\
    TMO-Net & $0.166\pm0.027$ & $0.474\pm0.010$ & $0.600\pm0.047$ & $0.657\pm0.043$ \\
    CustOmics & $0.185\pm0.013$ & $0.226\pm0.034$ & $0.567\pm0.049$ & $0.524\pm0.046$ \\
    Independent marginals & $0.115\pm0.007$ & $0.546\pm0.007$ & $0.024\pm0.031$ & $0.876\pm0.021$ \\
    \addlinespace
    \multicolumn{5}{@{}l}{\textit{After treatment}} \\
    ISOMERIC & $0.231\pm0.006$ & $0.246\pm0.015$ & $0.339\pm0.076$ & $0.563\pm0.070$ \\
    TMO-Net & $0.210\pm0.026$ & $0.254\pm0.016$ & $0.305\pm0.042$ & $0.771\pm0.059$ \\
    CustOmics & $0.269\pm0.041$ & $0.147\pm0.013$ & $0.223\pm0.083$ & $0.598\pm0.072$ \\
    Independent marginals & $0.190\pm0.007$ & $0.275\pm0.034$ & $-0.022\pm0.083$ & $0.767\pm0.054$ \\
    \bottomrule
  \end{tabular}
  \caption{\textbf{Independent empirical marginal baseline compared with the
  three learned generators.} RNA $W_1$ (R1), RNA co-expression edge F1 (R2), and
  same-gene slope correlation $r_\beta$ (R3) compare generated and held-out real
  cohorts. Paired membership AUC (R4) measures training-versus-held-out
  separation, $0.5$ denotes no separation and values above $0.5$ indicate
  training preference. Each entry is the mean $\pm$ sample standard deviation
  across three outer folds after averaging three replicates per fold.
  Replicates are sampling cohorts for Independent marginals and training
  initializations for the learned models. Arrows indicate closer held-out
  agreement. These four selected metrics do not summarize every property
  assessed by the complete R1--R4 evaluation.}
  \label{tab:independent-marginals-summary}
\end{table}

\begin{figure}[p]
  \centering
  \includegraphics[width=\linewidth]
    {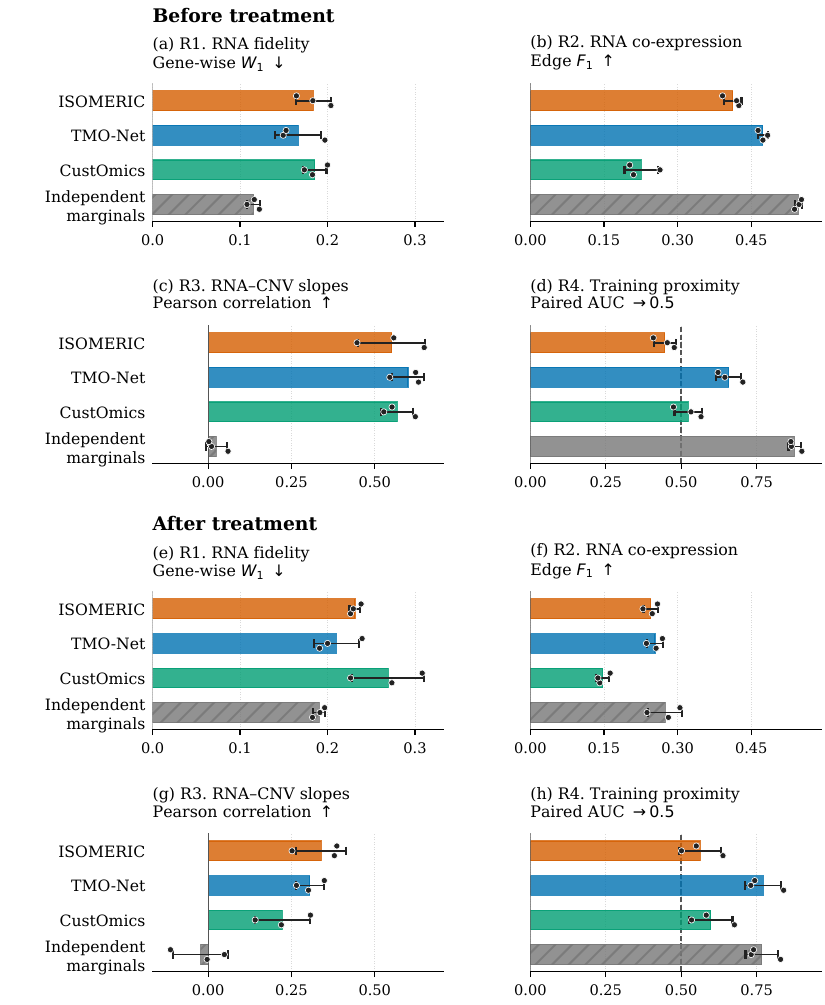}
  \caption{\textbf{Independent empirical marginals compared with the adapted
  VAEs.} Panels a--d show the before-treatment cohort and e--h show the
  after-treatment cohort. RNA Wasserstein distance (R1), RNA co-expression edge
  F1 (R2), and same-gene slope correlation (R3) compare generated cohorts with
  held-out data. Arrows indicate the direction of closer agreement. Paired
  membership AUC (R4) measures training-versus-held-out separation: $0.5$
  denotes no separation, and higher values indicate training preference.
  Bars show means and error bars show sample SDs across three outer folds.
  Points show individual fold values after averaging three generated cohorts
  for the baseline or three training initializations for each VAE.
  The baseline combines favorable RNA-only scores with near-zero slope
  correlations. Table~\ref{tab:independent-marginals-summary} reports the
  numerical values.}
  \label{fig:independent-marginals-comparison}
\end{figure}

\clearpage
\section{Sensitivity to the co-expression edge threshold}
\label{app:coexpression-threshold-sensitivity}

Does excess synthetic network density depend on the primary correlation
threshold of $0.70$? We repeat the before-treatment edge analysis at
$r\geq\tau$, with $\tau\in\{0.60,0.65,0.70,0.75,0.80\}$, retaining
Benjamini--Hochberg-adjusted $q<0.05$ across all gene pairs. We reuse the saved
Pearson correlation matrices for the three models, three outer folds, and
three training seeds. Each fold retains its 3,000-gene panel and equal-sized
real-training, held-out, and synthetic RNA subsets (161, 152, and 180 samples
per subset in folds 0, 1, and 2). Only edge selection changes.

Across the five thresholds, mean density ratios exceed one for all models,
and the observed mean edge-F1 ordering remains TMO-Net $>$ ISOMERIC $>$
CustOmics (Figure~\ref{fig:coexpression-threshold-sensitivity},
Table~\ref{tab:coexpression-threshold-sensitivity}). Both patterns also hold
within each fold after averaging seeds. This descriptive check concerns edge
metrics, module agreement and hub overlap were not recomputed. We retain
$0.70$ as the primary threshold.

\begin{figure}[H]
  \centering
  \includegraphics[width=\linewidth]
    {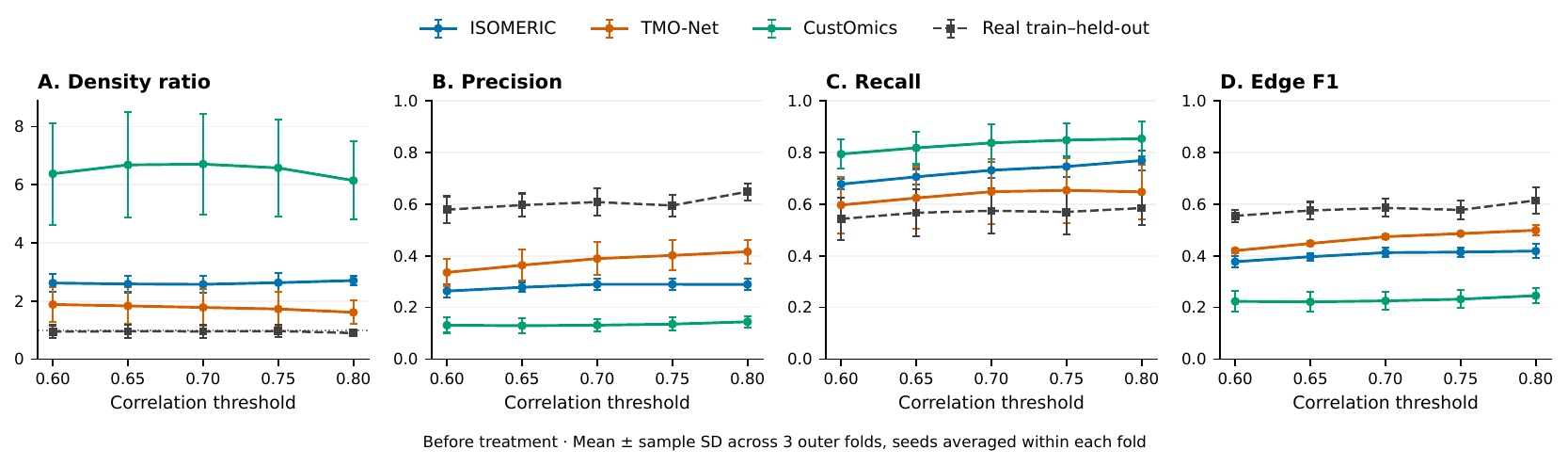}
  \caption{\textbf{Co-expression threshold sensitivity before treatment.}
  \textbf{(A)} Candidate/held-out network density ratio (one denotes equal
  density), \textbf{(B--D)} edge precision, recall, and F1 (higher indicates
  greater recovery of held-out edges). Colored curves compare synthetic RNA
  with held-out RNA, the gray dashed curve uses real-training RNA as the
  candidate. Points and bars show mean $\pm$ sample standard deviation across
  three outer folds, after averaging metrics over three training seeds per fold.}
  \label{fig:coexpression-threshold-sensitivity}
\end{figure}

\begin{table}[H]
  \centering
  \caption{\textbf{Edge metrics across correlation thresholds before treatment.}
  Values are mean $\pm$ sample SD across three outer folds, after averaging
  three training seeds within each fold. Each candidate is compared with
  held-out real RNA. The shared real-training reference appears once per
  threshold, $\tau=0.70$ is the primary setting.}
  \label{tab:coexpression-threshold-sensitivity}
  \footnotesize
  \setlength{\tabcolsep}{3pt}
  \begin{tabular}{@{}clcccc@{}}
    \toprule
    $\tau$ & Candidate & Density ratio & Precision & Recall & Edge F1 \\
    \midrule
    0.60 & ISOMERIC & $2.620 \pm 0.313$ & $0.264 \pm 0.025$ & $0.677 \pm 0.021$ & $0.377 \pm 0.022$ \\
     & TMO-Net & $1.882 \pm 0.607$ & $0.336 \pm 0.051$ & $0.596 \pm 0.109$ & $0.420 \pm 0.009$ \\
     & CustOmics & $6.380 \pm 1.749$ & $0.131 \pm 0.029$ & $0.793 \pm 0.056$ & $0.224 \pm 0.040$ \\
     & Real training & $0.952 \pm 0.216$ & $0.578 \pm 0.052$ & $0.542 \pm 0.081$ & $0.554 \pm 0.024$ \\
    \addlinespace[2pt]
    0.65 & ISOMERIC & $2.587 \pm 0.285$ & $0.278 \pm 0.019$ & $0.705 \pm 0.030$ & $0.396 \pm 0.014$ \\
     & TMO-Net & $1.833 \pm 0.634$ & $0.364 \pm 0.059$ & $0.623 \pm 0.120$ & $0.448 \pm 0.009$ \\
     & CustOmics & $6.685 \pm 1.823$ & $0.129 \pm 0.028$ & $0.817 \pm 0.063$ & $0.222 \pm 0.038$ \\
     & Real training & $0.961 \pm 0.218$ & $0.596 \pm 0.045$ & $0.566 \pm 0.092$ & $0.575 \pm 0.033$ \\
    \addlinespace[2pt]
    0.70 & ISOMERIC & $2.575 \pm 0.299$ & $0.290 \pm 0.021$ & $0.731 \pm 0.031$ & $0.412 \pm 0.018$ \\
     & TMO-Net & $1.780 \pm 0.633$ & $0.389 \pm 0.064$ & $0.648 \pm 0.126$ & $0.474 \pm 0.010$ \\
     & CustOmics & $6.710 \pm 1.740$ & $0.132 \pm 0.025$ & $0.836 \pm 0.072$ & $0.226 \pm 0.034$ \\
     & Real training & $0.956 \pm 0.215$ & $0.608 \pm 0.052$ & $0.574 \pm 0.088$ & $0.585 \pm 0.035$ \\
    \addlinespace[2pt]
    0.75 & ISOMERIC & $2.633 \pm 0.320$ & $0.289 \pm 0.021$ & $0.745 \pm 0.040$ & $0.414 \pm 0.019$ \\
     & TMO-Net & $1.727 \pm 0.584$ & $0.401 \pm 0.059$ & $0.653 \pm 0.126$ & $0.486 \pm 0.005$ \\
     & CustOmics & $6.581 \pm 1.659$ & $0.136 \pm 0.025$ & $0.847 \pm 0.065$ & $0.232 \pm 0.034$ \\
     & Real training & $0.966 \pm 0.204$ & $0.594 \pm 0.041$ & $0.569 \pm 0.088$ & $0.577 \pm 0.038$ \\
    \addlinespace[2pt]
    0.80 & ISOMERIC & $2.706 \pm 0.166$ & $0.289 \pm 0.021$ & $0.768 \pm 0.039$ & $0.418 \pm 0.027$ \\
     & TMO-Net & $1.613 \pm 0.409$ & $0.416 \pm 0.047$ & $0.647 \pm 0.105$ & $0.499 \pm 0.021$ \\
     & CustOmics & $6.148 \pm 1.346$ & $0.145 \pm 0.022$ & $0.853 \pm 0.067$ & $0.246 \pm 0.028$ \\
     & Real training & $0.900 \pm 0.062$ & $0.648 \pm 0.033$ & $0.584 \pm 0.066$ & $0.614 \pm 0.051$ \\
    \bottomrule
  \end{tabular}
\end{table}

\clearpage
\endgroup

\end{document}